\documentclass[11pt, logo, onecolumn, copyright, colorlinks=true, allcolors=blue]{nvidiatechreport}
\usepackage{url}
\usepackage{graphicx} 
\usepackage{subcaption,ragged2e}
\usepackage[round]{natbib}
\usepackage{placeins}

\usepackage[normalem]{ulem}
\usepackage[utf8]{inputenc} 
\usepackage[T1]{fontenc}    
\usepackage{tikz}
\usepackage{pgf-pie}
\usepackage{enumitem}
\usetikzlibrary{positioning}
\usetikzlibrary{calc}
\usepackage{array}
\usepackage{booktabs}
\usepackage{wrapfig}
\usepackage{caption} 
\usepackage{listings}
\usepackage{adjustbox}
\usepackage{footnote}
\usepackage{multirow}
\usepackage{amsmath}
\usepackage{amsfonts}
\usepackage{breakcites}
\usepackage{blindtext}
\usepackage{svg}
\usepackage[most]{tcolorbox}
\usepackage{tabularx}
\usepackage{graphicx} 
\usepackage{xcolor}
\usepackage{float}
\usepackage{needspace}
\usepackage{makecell}
\usepackage{colortbl}
\usepackage{stfloats}
\usepackage{siunitx}
\usepackage{graphicx}
\usepackage{booktabs}

\setlist[itemize]{topsep=1pt,itemsep=1pt}
\setlist[enumerate]{topsep=1pt,itemsep=1pt}
\usepackage{xparse}
\RenewDocumentCommand{\paragraph}{s o m}{%
  \par\medskip\noindent\textbf{#3}\quad\ignorespaces
}

\newcommand{\ourmodelfull}{NVIDIA Nemotron 3.5 Super\xspace}
\newcommand{\ourmodel}{Nemotron 3.5 Super\xspace}

\newcommand{\ourbasemodel}{Nemotron 3.5 Super 120B-A12B Base\xspace}
\newcommand{\lightbasemodel}{Lightning 3.5 30B-A3B Base\xspace}

\title{Nemotron 3.5 Super and Lightning: Open, Efficient Mixture-of-Experts Hybrid Mamba-Transformer Models for Agentic AI}
\author{\large NVIDIA}
\date{}

\begin{document}

\begin{abstract}
\large \textbf{Abstract.}
\normalsize




We introduce Nemotron 3.5 Super and Nemotron 3.5 Lightning, open and efficient Mixture-of-Experts Hybrid Mamba-Transformer models designed for reasoning, long-context understanding, and agentic workloads. Super is a 120B-parameter, 12B-active multimodal model, while Lightning is a smaller 30B-parameter, 3B-active text model; both designed for efficient deployment. Both models combine large-scale distillation, long-context training, and native speculative decoding, while Super further incorporates Indic continued pretraining and agentic software-engineering mid-training.

We develop a comprehensive post-training pipeline spanning supervised fine-tuning and reinforcement learning (RLVR and RLHF), complemented in Super by specialized teacher training and multi-teacher on-policy distillation (MOPD). This transforms Super from a text-only base model into a multimodal agent capable of reasoning over text, images, video, and interactive tool environments. We further explore test-time scaling through parallel agent execution and self-verification.

Across training and deployment, we jointly optimize capability and efficiency. Indic continued pretraining improves Indian-language benchmarks by up to 9.82 percentage points, while agentic mid-training strengthens software-engineering capabilities and robustness across agent harnesses. The post-trained Super model achieves 51.78\% on SWE-bench Pro, surpassing Nemotron 3 Ultra despite using approximately 4.6× fewer active parameters. Post-training MTP boosting improves speculative-decoding acceptance length by up to 21.8\% under probabilistic draft sampling, while NVFP4 quantization achieves 99.62\% median accuracy recovery relative to BF16. Inference optimizations enable up to 4,856 output tokens/s/GPU on GB200 while sustaining at least 50 tokens/s/user. Under the evaluated interactive agentic workload, NVFP4 achieves 2.43$\times$ the throughput of BF16, while Nemotron 3.5 Super delivers 2.34$\times$ and 1.90$\times$ the throughput of Qwen3.5 and Qwen3.8 Flash-Next, respectively. We release base, BF16 post-trained, and NVFP4-quantized model checkpoints, alongside standalone parallel speculative drafters (LiLiCorr, DFlash, and DSpark), with accompanying model cards, pretraining and post-training datasets, cookbooks, and training recipes.

\end{abstract}

\maketitle

\section{Introduction}
\label{sec:intro}

\begin{figure*}[t]
    \centering
    \includegraphics[width=\textwidth]{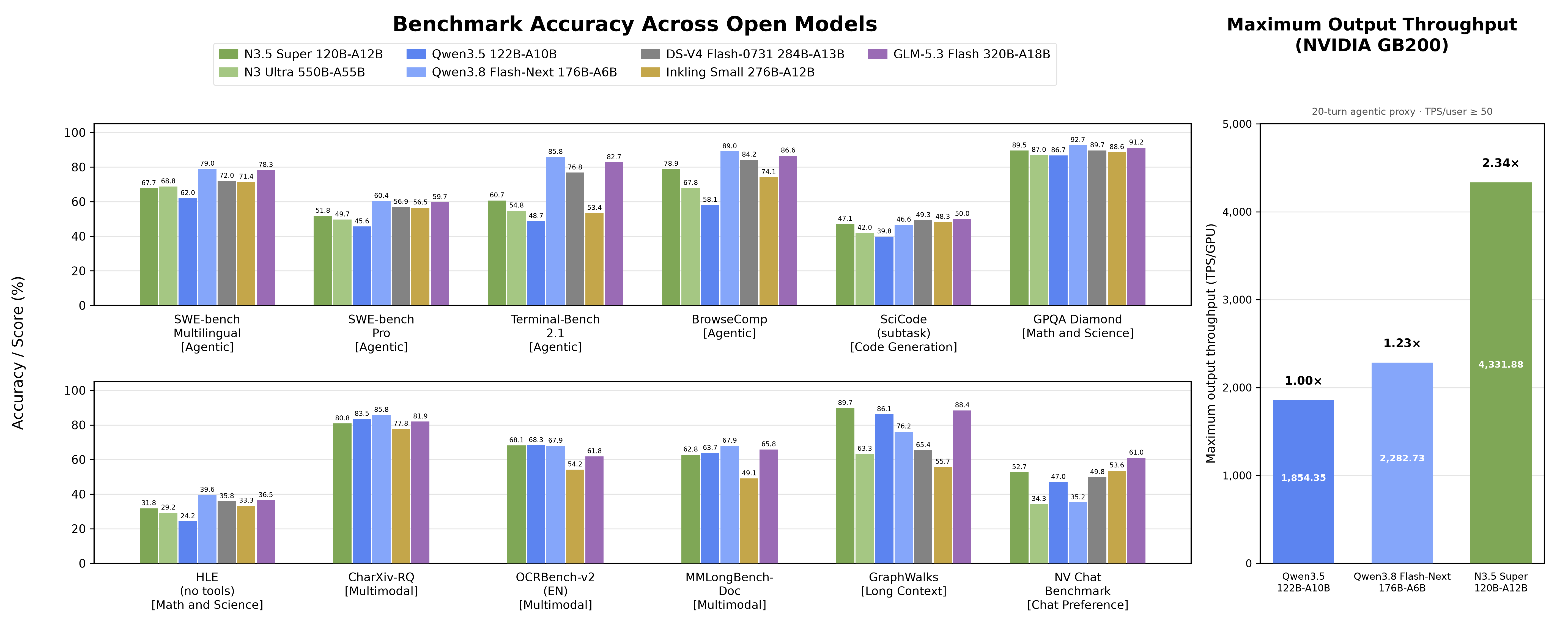}
    \caption{Benchmark performance of Nemotron 3.5 Super (120B-A12B) compared with six leading open models across agentic tasks, code generation, mathematical and scientific reasoning, multimodal understanding, long-context reasoning, and chat preference. Higher scores indicate better performance. Full evaluation results and methodology are provided in Section 3.8 and Table 10.}
    \label{fig:main}
\end{figure*}

We present Nemotron 3.5 Super and Nemotron 3.5 Lightning, efficient Mixture-of-Experts Hybrid Mamba-Transformer models designed for reasoning, multimodal understanding, and long-running agentic workloads. Nemotron 3.5 Super is a 120 billion total and 12 billion active parameter model initialized from Nemotron 3 Super~\citep{nvidia2026nemotron3superopen}, while Lightning 3.5 is a 30 billion total and 3 billion active parameter model initialized from Nano 3 Base~\citep{nvidia2025nemotron3nanoopen}. As LLM applications increasingly move toward long-horizon agents that write code, interact with tools, reason over large contexts, and operate across multimodal environments, both model capability and inference efficiency become increasingly important. Nemotron 3.5 targets this regime through efficient hybrid architectures, native speculative decoding with Multi-Token Prediction (MTP)~\citep{gloeckle2024better}, large-scale distillation, specialized continued pretraining and mid-training, agent-focused post-training, and NVFP4 quantization.

Figure~\ref{fig:main} compares Nemotron 3.5 Super against leading models across reasoning, coding, agentic, and multimodal benchmarks.

We begin by adapting the MTP architecture for efficient recursive speculative decoding. Nemotron 3.5 Super consolidates two non-shared MTP heads into a single recursively shared head, while Lightning initializes its MTP head from Nemotron 3 Super Base through architecture-aware projection fusion and pruning. Both models undergo a dedicated 100 billion token MTP warm-up with the backbone frozen. For Super, we additionally introduce hidden-state mixing to mitigate the train--inference conditioning mismatch during recursive drafting. We subsequently distill both models for 1 trillion tokens using the next-larger Nemotron form factor as teacher: Ultra 3 Base~\citep{nvidia_nemotron_3_ultra_2026} for Super and Super 3 Base~\citep{nvidia2026nemotron3superopen} for Lightning. Following distillation, Lightning undergoes long-context training, while Super proceeds through additional continued-pretraining and agentic mid-training stages.

We further improve Nemotron 3.5 Super through targeted continued-pretraining and mid-training stages. First, we perform 100 billion tokens of Indic continued pretraining, incorporating 10 billion tokens of synthetic and natural Indic data alongside 90 billion tokens from the preceding training mixture. To support this stage, we use NeMo Data Designer~\citep{nemo-data-designer,nemo-data-designer-tech-report} to generate approximately 2.83 trillion synthetic output tokens across 22 Indian languages and 34 language-script configurations. This stage improves all five evaluated Indic benchmarks, with gains of up to 9.82 percentage points. We then extend the model to a 1M-token context length using 35 billion tokens of continued pretraining, followed by 65.4 billion tokens of agentic software-engineering mid-training at 256K context. The mid-training mixture combines whole repositories, issue--pull-request pairs, commit histories, build and CI logs, and agent trajectories. Controlled evaluations using identical experimental SFT show improvements in long-horizon software engineering and robustness across agent harnesses, including DeepSWE pass@1 increasing from 6.4\% to 10.4\% and SWE-bench Verified under the Codex harness improving from 48.4\% to 61.8\%. To recover long-context capabilities affected by 256K mid-training, we perform an additional 9 billion token recovery stage, restoring RULER-1M~\citep{hsieh2024ruler} performance from 33.5 to 81.0.

We post-train Nemotron 3.5 Super through a comprehensive pipeline spanning Supervised Fine Tuning (SFT), Reinforcement Learning from Verifiable Rewards (RLVR), Reinforcement Learning from Human Feedback (RLHF), specialized teacher training, and Multi-teacher On-Policy Distillation (MOPD). Our two-stage SFT pipeline transforms the text-only base model into a vision-language model while jointly developing reasoning, chat, safety, long-context, and agentic capabilities. RL training spans diverse environments, including Olympiad-level competitive programming and finance-agent tasks requiring SEC filing search, document parsing, evidence retrieval, and tool use. We further train specialized teachers for video reasoning and tracking, vision tool use, and alignment, before consolidating their capabilities through MOPD. Nemotron 3.5 Lightning follows a lighter post-training pipeline consisting of single-stage text-only SFT and RLVR. The resulting Super model achieves 51.78\% on SWE-bench Pro, exceeding Nemotron 3 Ultra despite using approximately 4.6$\times$ fewer active parameters. Beyond model training, we explore test-time scaling through parallel agent execution and self-verification, improving Terminal-Bench 2.1 from 62.7\% to 71.9\% and DeepSWE from 24.6\% to 34.5\%.

In parallel, we optimize Nemotron 3.5 for efficient inference and deployment. Post-training MTP boosting improves average speculative-decoding acceptance length from 4.287 to 4.718 under greedy decoding and from 3.787 to 4.612 under probabilistic draft sampling on SPEED-Bench~\citep{speedbenchnvidia}. We develop NVFP4 post-training quantization and mixed-precision recipes across weights, caches, and the MTP drafter, complemented by quantization-aware distillation, achieving 99.62\% median accuracy recovery relative to BF16 across 36 benchmarks for the released quantized Super checkpoint. We additionally optimize hybrid-state caching and GPU kernels for efficient long-context and agentic serving, and develop standalone parallel speculative drafters for low-concurrency, memory-bound inference. On a single GB200 GPU, Nemotron 3.5 Super achieves up to 4,856 output tokens/s/GPU on a 20-turn agentic proxy workload while sustaining at least 50 tokens/s/user. Under the evaluated interactive serving conditions, NVFP4 delivers 2.43$\times$ the throughput of BF16, while Super achieves 2.34$\times$ and 1.90$\times$ the throughput of Qwen3.5 122B-A10B and Qwen3.8 Flash-Next, respectively. We release base, BF16 post-trained, and NVFP4-quantized model checkpoints, alongside standalone parallel speculative drafters (LiLiCorr for Super, and DFlash and DSpark for Lightning), with accompanying model cards, pretraining and post-training datasets, cookbooks, and training recipes.

\paragraph{Checkpoints}
\begin{itemize}
    \item \href{https://huggingface.co/nvidia/NVIDIA-Nemotron-3.5-Super-VL-120B-A12B-Base-BF16}{\texttt{Nemotron 3.5 Super VL 120B-A12B Base BF16} \includegraphics[height=0.9em]{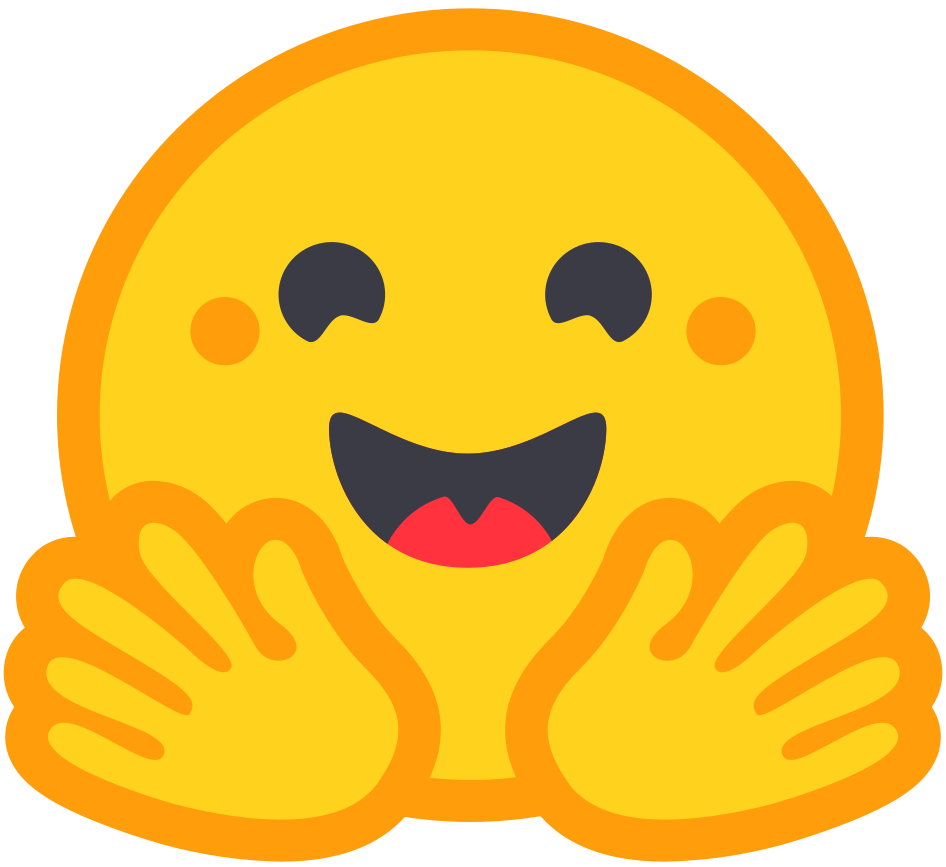}} : base model
    \item \href{https://huggingface.co/nvidia/NVIDIA-Nemotron-3.5-Super-VL-120B-A12B-BF16}{\texttt{Nemotron 3.5 Super VL 120B-A12B BF16} \includegraphics[height=0.9em]{assets/huggingface-color.png}} : post-trained model
    \item \href{https://huggingface.co/nvidia/NVIDIA-Nemotron-3.5-Super-VL-120B-A12B-NVFP4}{\texttt{Nemotron 3.5 Super VL 120B-A12B NVFP4} \includegraphics[height=0.9em]{assets/huggingface-color.png}} : post-trained and NVFP4-quantized model
    \item \href{https://huggingface.co/nvidia/NVIDIA-Nemotron-3.5-Super-VL-120B-A12B-NVFP4-LiLiCorr}{\texttt{Nemotron 3.5 Super VL 120B-A12B NVFP4 LiLiCorr} \includegraphics[height=0.9em]{assets/huggingface-color.png}} : LiliCorr

    \item \href{https://huggingface.co/nvidia/NVIDIA-Nemotron-3.5-Lightning-30B-A3B-Base-BF16}{\texttt{Nemotron 3.5 Lightning 30B-A3B Base BF16} \includegraphics[height=0.9em]{assets/huggingface-color.png}} : base model
    \item \href{https://huggingface.co/nvidia/NVIDIA-Nemotron-3.5-Lightning-30B-A3B-BF16}{\texttt{Nemotron 3.5 Lightning 30B-A3B BF16} \includegraphics[height=0.9em]{assets/huggingface-color.png}} : post-trained model
    \item \href{https://huggingface.co/nvidia/NVIDIA-Nemotron-3.5-Lightning-30B-A3B-NVFP4}{\texttt{Nemotron 3.5 Lightning 30B-A3B NVFP4} \includegraphics[height=0.9em]{assets/huggingface-color.png}} : post-trained and NVFP4-quantized model
    \item \href{https://huggingface.co/nvidia/NVIDIA-Nemotron-3.5-Lightning-30B-A3B-NVFP4-DFlash}{\texttt{Nemotron 3.5 Lightning 30B-A3B NVFP4 DFlash} \includegraphics[height=0.9em]{assets/huggingface-color.png}} : DFlash
    \item \href{https://huggingface.co/nvidia/NVIDIA-Nemotron-3.5-Lightning-30B-A3B-NVFP4-DSpark}{\texttt{Nemotron 3.5 Lightning 30B-A3B NVFP4 DSpark} \includegraphics[height=0.9em]{assets/huggingface-color.png}} : DSpark   

\end{itemize}


We organize the remainder of the report into four sections: Pretraining~(\S\ref{sec:pretraining}), Post-training~(\S\ref{sec:posttraining}), Quantization~(\S\ref{sec:quantization}), and Inference~(\S\ref{sec:inference}).
\section{Pretraining}
\label{sec:pretraining}

In this section, we describe the pretraining and mid-training recipe used to develop Nemotron 3.5 Super and Lightning 3.5. We first adapt and warm up the Multi-Token Prediction (MTP) head for recursive speculative decoding, then distill both models for 1 trillion tokens using the next-larger Nemotron 3 model as teacher. Following distillation, both models undergo long-context training. For Super, we additionally apply Indic continued pretraining, perform agentic software-engineering mid-training at 256K context, and finish with a short 1M-context recovery stage. We also report the resulting base-model capabilities and the effects of these stages on multilingual, long-context, and agentic performance.

\subsection{Multi-Token Prediction (MTP)}
\label{sec:mtp_init}

Nemotron 3.5 Super and Nemotron 3.5 Lightning feature native speculative-decoding support via a Multi-Token Prediction (MTP) head, adopting the recursively shared MTP formulation detailed in \citet{nvidia2026nemotron3superopen}. The MTP head functions as an integrated internal drafter that predicts multiple future tokens from the backbone's hidden representations. This enables multi-token candidate verification in a single forward pass over the backbone using rejection sampling~\citep{specdec} or exact token matching.

\paragraph{MTP Initialization.}
To establish the MTP drafting head, we employ model-specific initialization strategies tailored to the architectural origins of the Super and Lightning variants:

\begin{itemize}
    \item \textbf{Nemotron 3.5 Super:} We initialize the MTP weights directly from the \textit{Nemotron 3 Super Base} checkpoint (prior to long-context extension), which originally contained two distinct MTP heads. To consolidate multi-token drafting into a single, recursively shared head while preserving the underlying representation space, we initialize the MTP parameters exclusively from the first MTP head and discard the second.
    
    \item \textbf{Nemotron 3.5 Lightning:} We initialize Lightning's MTP head from that of \textit{Nemotron 3 Super Base}. We first fuse the latent MoE projections into the experts' up- and down-projection matrices, then prune the hidden size from 4096 to 2688, MoE intermediate size from 2688 to 1856, and number of routed experts from 512 to 128 to match Lightning.
\end{itemize}

\paragraph{MTP Warmup Training.}
Following initialization, we conduct a dedicated MTP warmup phase for \(100\,\text{B}\) tokens, freezing the backbone parameters and optimizing only the MTP head. 
For Nemotron 3.5 Lightning, we applied the standard 2-step recursive MTP training. However, this naive training induces a pronounced train--inference conditioning mismatch. During standard training, recursive step \(k\) conditions on the full shifted sequence of ground-truth hidden states generated at step \(k-1\). In contrast, during autoregressive drafting at inference time, the MTP head conditions on an increasingly heterogeneous mixture of backbone hidden states and synthetic states generated by preceding MTP steps. This discrepancy compounds with increasing draft depth, degrading token acceptance rates at later positions~\citep{nvidia2026nemotron3ultraopen}.

To bridge this distribution gap in \ourmodel, we adopt a hidden-state-mixing strategy similar to \citet{nvidia2026nemotron3ultraopen}. At training step \(k\), input hidden states are sampled from the sequence of states produced by earlier MTP steps \(1, \ldots, k-1\), rather than relying exclusively on step \(k-1\). Exposing the MTP head to heterogeneous hidden-state mixtures during warmup enhances robustness across deeper draft steps, achieving significant acceptance rate gains without the computationally intensive multi-step simulation required by test-time training (TTT)~\citep{li2025eagle3scalinginferenceacceleration}.

\subsection{Distillation}

We initialized Super 3.5 Base from Super 3 Base~\citep{nvidia2026nemotron3superopen} and Lightning 3.5 Base from Nano 3 Base~\citep{nvidia2025nemotron3nanoopen}, then distilled each model for 1 trillion tokens.

For Lightning 3.5 Base, we used top-128 logits from Super 3 Base on 1 trillion tokens of the Ultra 3 Phase 2 data~\citep{nvidia2026nemotron3ultraopen}. We used standard forward KL divergence loss along with an extra sink token to capture the residual probability mass outside top-128. To provide training signal for rare tokens, we combined the KL loss with cross-entropy loss, assigning them weights of 0.9 and 0.1, respectively. To limit early drift from the initialization, we began distillation with the minimum Learning Rate (LR) of Nano 3 Base ($10^{-5}$) and ramped it up slowly to $10^{-4}$, 10\% of the Nano 3 Base's peak LR, over 200 billion tokens. We then decayed the LR linearly back to $10^{-5}$ over the remaining 800 billion tokens. All other hyperparameters and settings match Nano 3 Base, except for the added MTP layers. We used an MTP loss scaling factor of 0.1.

For Super 3.5 Base, we used the same distillation recipe as Lightning 3.5 Base but with Ultra 3 Base as the teacher and Skew Reverse KL divergence~\citep{ko2024distillm} as the distillation loss function with $\alpha=0.1$. Additionally, we used a smaller MoE sequence load balancing loss coefficient of $10^{-5}$, as compared to $10^{-4}$ for Lightning 3.5. We chose learning rates the same way as Lightning 3.5 Base, but with Super 3 Base LRs as the anchor, using $4.5 \times 10^{-6}$ as the minimum and $4.5 \times 10^{-5}$ as the maximum.

\subsection{Indic Continued Pretraining}
\label{sec:indic-cpt}

We augment Nemotron 3.5 Super with an Indic continued-pretraining stage that combines synthetic text generated from English documents with natural Indic corpora to strengthen Indian-language capabilities. Our twelve-language synthetic-data pipeline uses Gemma-4-31B-it for translation. We describe the data and generation pipeline, then present Super's training mixture and results, followed by separate Lightning 3.5 mixture ablations.

\FloatBarrier
\paragraph{Indic data: motivation, coverage, and scale.}

Natural Indic text is scarce in many languages, and its coverage varies across languages, scripts, and subjects. We address these gaps by translating curated English documents into Indic languages. Nemotron-CC provides most of the English source documents. The sources cover general web content, education, mathematics, STEM, and question-answering. Keeping the source documents lets us check whether the generated text preserves their meaning and structure.

We generated data across 22 Indian languages: native-script and romanized text for twelve priority languages, and native-script text for ten additional lower-resource languages. The \href{https://huggingface.co/datasets/nvidia/Nemotron-Pretraining-Indic-v1}{Nemotron-Pretraining-Indic-v1} corpus contains a total of approximately 2.83 trillion tokens of generated Indic text across 34 language-script configurations, counted using a Gemma-4 tokenizer (Table~\ref{tab:indic-cpt-1}). The twelve priority languages contribute approximately 2.50 trillion tokens, including 224.6 billion romanized tokens; the additional ten languages contribute approximately 327 billion native-script tokens. This scale and coverage provide a large, reusable pool of training text across languages and subjects, including languages where natural text is limited. Model builders can draw on this pool to study language mixtures, script choices, and training budgets for their own models.

We created substantially more data than was used in Super's Indic CPT stage and will release the broader corpus publicly. Other teams can use the data directly, inspect how it was produced, and extend generation to their own needs.

\begin{table}[!htbp]
\centering
\footnotesize
\caption{Reported input and output token counts by language and script, in billions (B), using a Gemma-4 tokenizer. Input counts may overlap across configurations; they are not unique source volume or total pipeline processing. Subtotals sum displayed rounded values. The output-only corpus total remains 2,825.6B (approximately 2.83T), not the number of tokens used to train Super. The additional ten languages have native-script data only.}
\label{tab:indic-cpt-1}

\setlength{\tabcolsep}{3.5pt}
\renewcommand{\arraystretch}{1.12}
\begin{tabularx}{\linewidth}{@{}>{\hsize=.8\hsize\raggedright\arraybackslash}Xrrrr>{\hsize=1.2\hsize\raggedright\arraybackslash}Xrr@{}}
\toprule
\mbox{Priority} & \multicolumn{2}{c}{Native} & \multicolumn{2}{c}{Romanized} & \mbox{Additional} & \multicolumn{2}{c}{Native} \\
\mbox{language} & Input & Output & Input & Output & \mbox{language} & Input & Output \\
\midrule
Bengali & 114.0 & 138.9 & 12.2 & 19.0 & Assamese & 16.3 & 31.1 \\
Gujarati & 115.5 & 196.4 & 12.0 & 17.2 & Bodo & 8.6 & 18.7 \\
Hindi & 118.1 & 149.3 & 12.8 & 18.4 & Dogri & 23.4 & 39.9 \\
Kannada & 103.5 & 180.5 & 11.3 & 19.1 & Kashmiri & 11.4 & 33.1 \\
Malayalam & 108.9 & 183.5 & 11.4 & 19.3 & Konkani & 16.7 & 28.9 \\
Marathi & 114.4 & 152.8 & 11.6 & 17.6 & Maithili & 20.0 & 31.3 \\
Nepali & 118.0 & 171.3 & 12.4 & 19.7 & Manipuri & 6.6 & 50.1 \\
Odia & 109.3 & 320.1 & 12.2 & 19.0 & Sanskrit & 12.6 & 22.8 \\
Punjabi & 112.3 & 262.6 & 12.0 & 17.6 & Santali & 8.3 & 40.8 \\
Tamil & 110.5 & 157.5 & 10.9 & 20.0 & Sindhi & 14.6 & 30.5 \\
Telugu & 115.3 & 192.4 & 12.2 & 18.8 & - & - & - \\
Urdu & 117.8 & 168.5 & 12.8 & 18.9 & - & - & - \\
\midrule
Subtotal & 1,357.6 & 2,273.8 & 143.8 & 224.6 & Subtotal & 138.5 & 327.2 \\
\bottomrule
\end{tabularx}
\end{table}

\FloatBarrier
\paragraph{Synthetic-data pipeline and efficiency.}

\begin{figure}[!t]
\centering
\includegraphics[width=\linewidth]{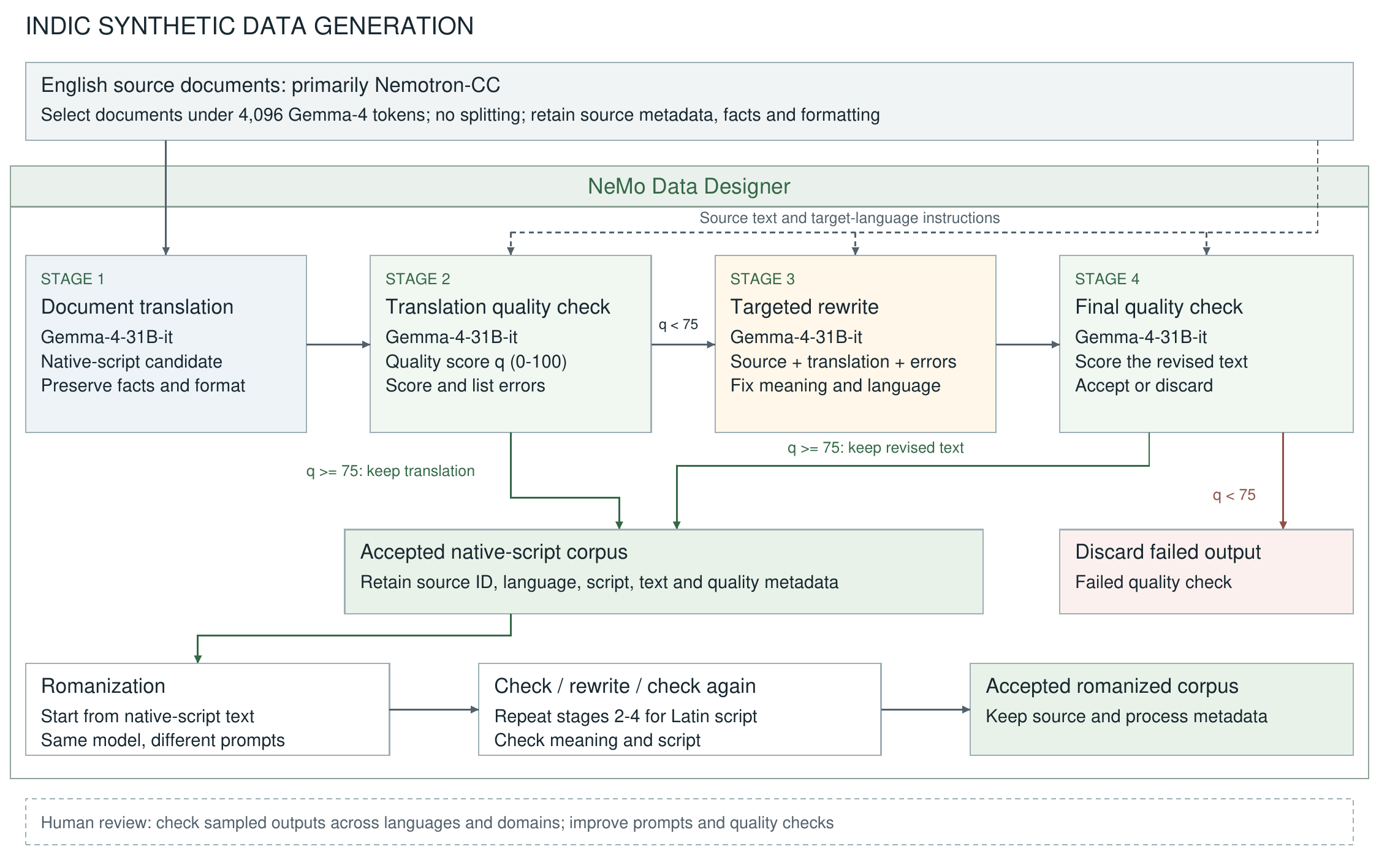}

\caption{Four-stage Indic synthetic-data pipeline implemented using NeMo Data Designer for the twelve priority languages. Selected English source documents, primarily from Nemotron-CC, contain fewer than 4,096 Gemma-4 tokens and are translated in full, without splitting. Translations are retained after passing the initial or post-rewrite quality check. Romanization starts from native-script text and uses the same model with script-specific prompts and quality checks. Sampled human review helps improve prompts and checks; it does not cover every document.}
\label{fig:indic-cpt-pipeline}
\end{figure}

We implement the four-stage pipeline for the twelve priority languages (Figure~\ref{fig:indic-cpt-pipeline}) using \href{https://docs.nvidia.com/nemo/datadesigner/}{NeMo Data Designer} \citep{nemo-data-designer,nemo-data-designer-tech-report}. Gemma-4-31B-it, selected based on translation-model comparisons, performs translation, quality assessment, and rewriting. We select source documents with fewer than 4,096 Gemma-4 tokens and translate each document in full, without splitting, using prompts designed to preserve facts, mathematical expressions, code, and formatting. The model then checks each translation against its source and assigns a quality score from 0 to 100. We keep translations scoring at least 75. Lower-scoring translations are rewritten using the source text and identified errors, then checked again and kept only if they pass.

Retained translations keep their source and quality metadata for traceability. Quality scores guide filtering, although using the same model for translation and judging may favor its own translations.

For the ten additional lower-resource languages, we use \href{https://huggingface.co/bodhan-ai/indic-translate}{Bodhan-MT} for translation and Gemma-4-31B-it for quality assessment, with language-specific acceptance thresholds. We use overlapping windows only for longer Santali documents and convert Manipuri translations from Bengali script to Meitei Mayek. The companion \textit{Scaling Synthetic Pre-training Data and Evaluation for Indic LLMs} report provides details.

For the twelve-language pipeline, we improved scheduling so translation and quality checks can run concurrently. Speculative decoding, FP8 weights, and FlashAttention reduced median translation cost from 0.835 to 0.335 GPU-hours per million English input tokens, a reduction of about 60\%. This is equivalent to processing 2.49 times as many tokens per GPU-hour. The comparison uses the same twelve-language benchmark, with approximately 46.8M input tokens per language. The gain applies to translation; quality checks and rewriting add further cost.

\paragraph{Data selected for Super continued pretraining.}

Super's Indic continued-pretraining run has a nominal 100B-token budget, with 10\% (10B tokens) allocated to selected synthetic text from our corpus and natural Indic text. The remaining 90\% (90B tokens) retains the preceding training mixture's relative source proportions. These are configured sampling allocations, not measured per-source token totals or counts of unique tokens in the corpus.

We chose the 10\% Indic share as a practical allocation within this budget, informed by the separate 200B-token Lightning mixture ablations (Table~\ref{tab:indic-cpt-3}). The 10\% mixture already delivered substantial gains; higher shares improved results further, but with smaller incremental gains.

\FloatBarrier
\paragraph{Super results and capability retention.}

We evaluate Indian-language knowledge, reasoning, and mathematics using MILU and human-verified Indic translations of MMLU-Pro, ARC-Challenge, ARC-Easy, and GSM8K.

Table~\ref{tab:indic-cpt-2} compares the distilled checkpoint immediately before Indic CPT with the checkpoint after this stage. Scores improve across all five Indic benchmarks, with gains ranging from 4.98 to 9.82 percentage points. These results describe the combined continued-pretraining stage; they do not separate the contributions of synthetic and natural Indic text.

\begin{table}[!htbp]
\centering
\footnotesize
\caption{Super performance on Indic benchmarks before and after Indic CPT. Scores are percentages; parentheses show the number of languages. Changes are calculated from unrounded scores.}
\label{tab:indic-cpt-2}

\setlength{\tabcolsep}{3.5pt}
\renewcommand{\arraystretch}{1.12}

\begin{tabularx}{\linewidth}{@{}>{\raggedright\arraybackslash}Xrrrrr@{}}
\toprule
\mbox{Checkpoint} & \shortstack[r]{MILU\\(10)} & \shortstack[r]{Indic MMLU-Pro\\(12)} & \shortstack[r]{Indic ARC-C\\(12)} & \shortstack[r]{Indic ARC-E\\(12)} & \shortstack[r]{Indic GSM8K\\strict (11)} \\
\midrule
Before Indic CPT (distilled) & 72.48 & 51.83 & 84.17 & 90.73 & 72.67 \\
After Indic CPT & 77.90 & 56.81 & 92.83 & 96.77 & 82.49 \\
Change (percentage points) & +5.42 & +4.98 & +8.66 & +6.04 & +9.82 \\
\bottomrule
\end{tabularx}
\end{table}

MILU covers ten supported languages; GSM8K excludes Odia because its evaluation exceeded the serving context limit. Translation also improves on twelve-language IGB/FLORES: English-to-Indic chrF++ rises from 35.43 to 47.47, and Indic-to-English from 58.75 to 64.30. The later midtraining checkpoint largely retains the Indic benchmark gains.

We also evaluated the Indic-CPT model across a broad set of English benchmarks and observed no regressions in the tested categories.

\begin{figure}[!htbp]
\centering
\includegraphics[width=\linewidth]{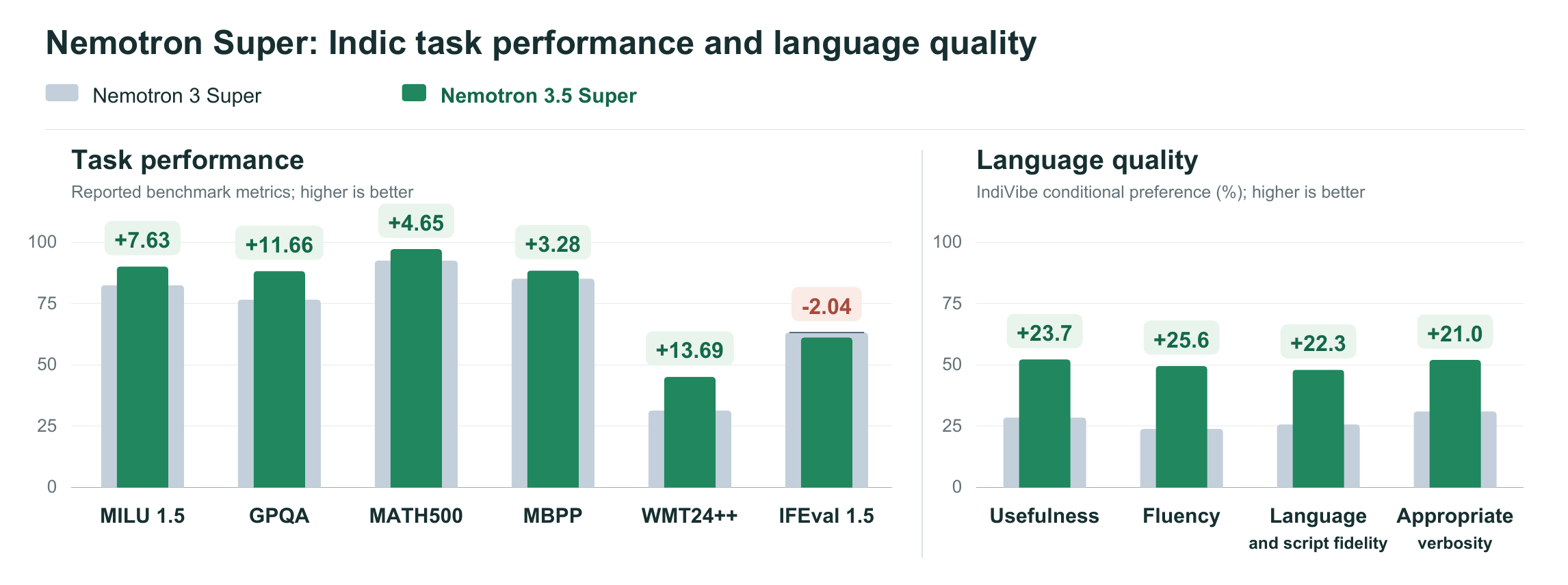}
\caption[Indic task performance and language quality across Super generations]{Indic task performance and language quality for Nemotron 3 Super (blue-gray) and Nemotron 3.5 Super (green). Left: six task benchmarks. Right: preliminary IndiVibe conditional mean preference scores across four language-quality dimensions, not direct head-to-head win rates. Labels show changes in percentage points (chrF points for WMT24++). Benchmark, scoring, and checkpoint details will appear in the companion Indic report.}
\label{fig:super35-indic-comparison}
\end{figure}

To complement this stage-level evaluation, Figure~\ref{fig:super35-indic-comparison} compares the evaluated Nemotron 3 and 3.5 Super models on a broader set of Indic tasks and language-quality measures. Nemotron 3.5 Super scores higher on five of the six task metrics shown and on all four IndiVibe dimensions. These results show the overall performance of the released Nemotron 3.5 Super model, with Indic CPT as part of its broader training pipeline.

\FloatBarrier
\paragraph{Lightning 3.5 mixture-ratio ablations.}

To study how the Indic share affects performance, we ran separate Lightning 3.5 research experiments with 10\%, 20\%, and 30\% Indic data under a fixed 200B-token training budget. These experiments are separate from the released Lightning 3.5 model, which did not include this Indic CPT stage. The four-benchmark mean increased with the Indic share, with the 30\% mixture scoring highest among the tested settings (Table~\ref{tab:indic-cpt-3}). These experiments inform mixture design but do not establish an optimal share for Super. Detailed language-level results, mixture configurations, and English-retention analysis will appear in the companion \textit{Scaling Synthetic Pre-training Data and Evaluation for Indic LLMs} report.

\begin{table}[!htbp]
\centering
\footnotesize
\caption{Lightning 3.5 research ablations. Each ablation uses a 200B-token budget. The reported mean covers MILU, MMLU-Pro, ARC-Challenge, and GSM8K. The base checkpoint is shown for reference; the fixed-budget comparison covers the three ablation runs.}
\label{tab:indic-cpt-3}

\setlength{\tabcolsep}{3.5pt}
\renewcommand{\arraystretch}{1.12}

\begin{tabularx}{\linewidth}{@{}>{\raggedright\arraybackslash}Xrr@{}}
\toprule
\mbox{Checkpoint} & \mbox{Indic share} & \mbox{Four-benchmark mean} \\
\midrule
Base reference & - & 45.5 \\
Mixture ablation & 10\% & 63.1 \\
Mixture ablation & 20\% & 64.6 \\
Mixture ablation & 30\% & 65.6 \\
\bottomrule
\end{tabularx}
\end{table}

Alongside the Nemotron 3.5 technical report, we will release the broader synthetic corpus, open-source the generation pipelines, and publish a companion technical report. That report will provide detailed translation-model comparisons, data-source ablations on Nemotron 3 Nano, and Lightning 3.5 mixture experiments. In collaboration with \href{https://ai4bharat.iitm.ac.in/}{AI4Bharat}, we will also release human-verified Indic LLM evaluation suite. Together, these resources will help model builders train, evaluate, and adapt models for Indian-language applications.

\FloatBarrier

\subsection{Long Context}

Similar to Nemotron 3 series, we added a long-context phase (LC-Phase) at the end of pretraining. In the LC-Phase, we performed continued pretraining (CPT) at 1M-token context length to equip the base model with long-context ability. For Lightning 3.5, we used a constant learning rate of $1*10^{-5}$. We used 8-way context parallelism, 8-way tensor parallelism, 8-way expert parallelism, and 2-way pipeline parallelism to train on GB200 GPUs. For Super 3.5, we used a constant learning rate of $4.5*10^{-6}$. We used 16-way context parallelism, 8-way tensor parallelism, and 64-way expert parallelism to train on GB300 GPUs.

We use the same data blend as Nemotron 3 Ultra \citep{nvidia2026nemotron3ultraopen}. The LC-Phase was trained for 21B tokens for Lightning and 35B tokens for Super. Note that for Super 3.5 after mid-training which was performed at a context length of 256K, we observed that the RULER-1M score dropped from 79.9 to 33.5. So we added a short long-context recovery phase after mid-training, for which we trained 9B tokens with the same recipe and successfully recovered the RULER score to 81.0.

\subsection{Mid-training}
\label{sec:midtraining}

After the long-context phase, we add a mid-training stage for agentic software engineering. Pretraining runs at short context and sees almost no agent interactions, and the natural data around software engineering, such as whole repositories, pull requests with their issues, commit histories, and build logs, is often too long for it. SFT in turn consists mostly of synthetic data, and very long SFT stages can reduce how much the model later gains from RL. Mid-training sits between the two. It trains the base model at long context on a diverse mixture of natural software-engineering data with a small share of agent traces, and leaves behavior shaping to post-training. The full recipe, including how each data source is built and filtered and the ablations behind each choice, is described in \citep{singhal2026midtraining}.

Table~\ref{tab:midtraining-data} shows the data blend we used for Super-3.5 mid-training (Lightning 3.5 did not have a mid-training stage). Most of it is natural data kept in its original format, without chat templates. The agent traces are the only SFT-style data, so we keep them to a small share of the blend and mask tool responses from the loss, training only on the model's own turns. Every source is decontaminated against SWE-bench and Terminal-Bench and has secrets removed.

\begin{table}[!htbp]
\centering
\footnotesize
\begin{tabular}{llr}
\toprule
Data source & Description & Share \\
\midrule
Issue--PR & GitHub issues with repository context and the merged patch & 41\% \\
Repositories & Whole repositories, with files that import each other placed together & 20\% \\
Agentic traces & Trajectories from many agent scaffolds, tool responses loss-masked & 10\% \\
Commit diffs & Multi-file commits and per-file edit histories & 3\% \\
Build and CI logs & Raw build and continuous-integration logs & 1\% \\
\midrule
Pretraining replay & Phase-2 pretraining blend & 20\% \\
Long-context data & Long-context phase data & 5\% \\
\bottomrule
\end{tabular}
\caption{Mid-training blend for \ourmodel{}, as shares of training tokens.}
\label{tab:midtraining-data}
\end{table}

We mid-trained the long-context checkpoint for 65.4B tokens at a sequence length of 256K. We reset the learning rate, warmed it up linearly to a maximum of $1 \times 10^{-5}$, about twice the final pretraining learning rate, and then held it constant for the rest of the run. Training at 256K lowered the RULER-1M score, so we followed mid-training with a short long-context recovery stage of about 9B tokens, described in the Long Context section above.

To measure the effect of mid-training, we ran identical SFT on the base checkpoint and on the mid-trained checkpoint and compared the two. This SFT is an experimental agentic blend of about 33B tokens (2K steps), not the one used in Section~\ref{sec:posttraining}, but we have found it to be a reliable signal of how a checkpoint will do after full post-training. Table~\ref{tab:midtraining-results} shows the results. The clearest gain on the standard benchmarks is on DeepSWE, our hardest long-horizon benchmark, where pass@1 rises from 6.4 to 10.4, and BrowseComp improves slightly. SWE-bench Verified, Multilingual, and Pro under OpenHands stay within run-to-run variation, which is about one point. The larger effect is on robustness across agent harnesses. Under OpenHands, the harness used for our standard evaluations, the base model already scores well and mid-training changes little. Under five other harnesses, the mid-trained model is better on every one, by 8.6 points on average on SWE-bench Verified and 6.8 points on Terminal-Bench 2.1, with the largest gain on Codex, where the base model is weakest.

\begin{table}[!htbp]
\centering
\footnotesize
\begin{tabular}{lcc}
\toprule
 & Base & Mid-trained \\
\midrule
SWE-bench Multilingual & \textbf{65.9} & 65.6 \\
SWE-bench Pro          & \textbf{52.1} & 51.7 \\
DeepSWE                & 6.4 & \textbf{10.4} \\
BrowseComp (subset)    & 64.2 & \textbf{65.0} \\
\midrule
\multicolumn{3}{l}{\emph{SWE-bench Verified, by harness}} \\
\quad OpenHands   & 71.3 & \textbf{71.9} \\
\quad Pi          & 57.4 & \textbf{67.8} \\
\quad OpenCode    & 54.4 & \textbf{65.8} \\
\quad Hermes      & 59.8 & \textbf{63.2} \\
\quad Codex       & 48.4 & \textbf{61.8} \\
\quad Claude Code & 53.8 & \textbf{58.0} \\
\midrule
\multicolumn{3}{l}{\emph{Terminal-Bench 2.1, by harness}} \\
\quad Pi          & 30.3 & \textbf{35.2} \\
\quad OpenCode    & 35.2 & \textbf{38.2} \\
\quad Hermes      & 44.2 & \textbf{53.9} \\
\quad Codex       & 27.0 & \textbf{39.7} \\
\quad Claude Code & 39.7 & \textbf{43.5} \\
\bottomrule
\end{tabular}
\caption{Effect of mid-training on \ourmodel{}, measured after the same experimental 33B-token (2K-step) SFT on the base and the mid-trained checkpoint (pass@1, \%). SWE-bench Multilingual and Pro use OpenHands, and DeepSWE is averaged over 4 runs. The OpenHands results come from a separate evaluation from the other harnesses.}
\label{tab:midtraining-results}
\end{table}

\subsection{Base Model Evaluations}
\label{subsec:base_model_evals}

We evaluate the base models across a broad suite of benchmarks covering
general knowledge, mathematical reasoning, code generation, commonsense
reasoning, reading comprehension, and multilingual capabilities.
Detailed evaluation settings and results are provided below.

\begin{table}[ht!]
\centering
\small
\setlength{\tabcolsep}{5pt}
\renewcommand{\arraystretch}{1.3}
\resizebox{0.92\textwidth}{!}{%
\begin{tabular}{lr|cc}
\toprule
\textbf{Task} & \textbf{Metric} &
\multicolumn{1}{c}{\textbf{Nemotron 3.5 Super}} &
\multicolumn{1}{c}{\textbf{Nemotron 3.5 Lightning}} \\
\midrule

\rowcolor{black!5}
\multicolumn{4}{l}{\textbf{General Knowledge}} \\
MMLU & \textit{exact match} & 84.14 & 78.63 \\
MMLU-Pro & \textit{exact match, TigerLab extract} & 76.25 & 67.50 \\
AGIEval-En & \textit{CoT exact match, custom extract} & 75.65 & 68.62 \\
SuperGPQA & \textit{CoT} & 47.65 & 33.21 \\
GPQA Diamond & \textit{5-shot, CoT exact match} & 66.16 & 33.33 \\
\midrule

\rowcolor{black!5}
\multicolumn{4}{l}{\textbf{Math}} \\
GSM8K & \textit{8-shot, CoT exact match} & 91.74 & 91.74 \\
Minerva MATH & \textit{exact match} & 87.66 & 83.10 \\
MATH-500 & \textit{4-shot, sampled pass@1, $n=32$} & 81.56 & 79.53 \\
\midrule

\rowcolor{black!5}
\multicolumn{4}{l}{\textbf{Code}} \\
HumanEval & \textit{sampled pass@1, $n=32$, EvalPlus sanitized} & 81.31 & 76.91 \\
MBPP-Sanitized & \textit{3-shot, sampled pass@1, $n=32$, EvalPlus sanitized} & 79.06 & 76.98 \\
LiveCodeBench CodeGen & \textit{pass@1} & 47.61 & 29.77 \\
BigCodeBench Complete & \textit{greedy} & 48.42 & 50.09 \\
BigCodeBench Instruct & \textit{greedy} & 49.39 & 51.75 \\
\midrule

\rowcolor{black!5}
\multicolumn{4}{l}{\textbf{Commonsense Understanding}} \\
CommonsenseQA & \textit{7-shot, acc} & 85.26 & 81.08 \\
ARC-Challenge & \textit{25-shot, acc\_norm} & 95.99 & 93.17 \\
HellaSwag & \textit{10-shot, acc\_norm} & 88.96 & 85.25 \\
OpenBookQA & \textit{0-shot, acc\_norm} & 48.80 & 47.80 \\
PIQA & \textit{0-shot, acc\_norm} & 85.09 & 85.09 \\
TruthfulQA MC2 & \textit{0-shot, acc} & 71.18 & 55.53 \\
WinoGrande & \textit{5-shot, acc} & 79.08 & 78.06 \\
SocialIQA & \textit{0-shot, acc} & 49.03 & 47.80 \\
\midrule

\rowcolor{black!5}
\multicolumn{4}{l}{\textbf{Reading Comprehension}} \\
RACE & \textit{0-shot, acc} & 91.39 & 87.85 \\
\midrule

\rowcolor{black!5}
\multicolumn{4}{l}{\textbf{Multilingual}} \\
Global MMLU & \textit{avg.\ across 8 languages} & 86.56 & 75.09 \\
MGSM & \textit{8-shot, native CoT avg.\ across 6 languages} & 87.53 & 82.27 \\
\bottomrule
\end{tabular}%
}

\caption[Base model evaluation results]{
Base model evaluation results for \textbf{\ourbasemodel} and
\textbf{\lightbasemodel}.
For sampled code and mathematical reasoning benchmarks, pass@1 is
estimated from 32 generations per prompt where indicated.
A dash indicates that a corresponding result is not available.
}
\label{tab:base-model-comparison}
\end{table}

All evaluation results reported for \ourbasemodel and \lightbasemodel were
collected via NeMo Evaluator SDK%
\footnote{\url{https://github.com/NVIDIA-NeMo/Evaluator}}
and NVIDIA's open-source container of LM Evaluation Harness%
\footnote{\url{https://github.com/EleutherAI/lm-evaluation-harness}},
unless otherwise stated.
For reproducibility, additional details on evaluation configurations are
available in the NeMo Evaluator SDK examples%
\footnote{\url{https://github.com/NVIDIA-NeMo/Evaluator/tree/main/examples}}.
The open-source LM Evaluation Harness container packaged through NVIDIA's
NeMo Evaluator SDK is available on NGC%
\footnote{\url{https://catalog.ngc.nvidia.com/orgs/nvidia/teams/eval-factory/containers/lm-evaluation-harness}}.

The evaluations use the following settings:

\begin{enumerate}
\setlength{\itemsep}{0.4em}

\item For general knowledge, we evaluate
MMLU~\citep{hendrycks2021measuringmassivemultitasklanguage},
MMLU-Pro, AGIEval-En, SuperGPQA, and GPQA Diamond.
We report exact-match metrics using benchmark-specific extraction
procedures where applicable. GPQA Diamond is evaluated using
5-shot chain-of-thought prompting.

\item For mathematical reasoning, we evaluate
GSM8K~\citep{cobbe2021trainingverifierssolvemath}
using 8-shot chain-of-thought exact match and Minerva MATH using
exact match. We additionally evaluate MATH-500 using 4-shot sampled
generation with 32 generations per prompt and report pass@1.

\item For code tasks, we evaluate
HumanEval~\citep{chen2021evaluatinglargelanguagemodels}
and MBPP~\citep{austin2021programsynthesislargelanguage}
using EvalPlus-sanitized variants~\citep{Liu_Is_Your_Code_2023}.
We report sampled pass@1 estimated from 32 generations per prompt.
MBPP-Sanitized uses 3-shot prompting, while HumanEval uses zero-shot
prompting. We additionally report results on LiveCodeBench CodeGen and
the complete and instruct variants of BigCodeBench.

\item For commonsense reasoning, we evaluate CommonsenseQA,
ARC-Challenge~\citep{Clark2018ThinkYH},
HellaSwag~\citep{zellers2019hellaswagmachinereallyfinish},
OpenBookQA~\citep{mihaylov2018suitarmorconductelectricity},
PIQA~\citep{bisk2019piqareasoningphysicalcommonsense},
TruthfulQA MC2,
WinoGrande~\citep{sakaguchi2019winograndeadversarialwinogradschema},
and SocialIQA using the accuracy or normalized-accuracy metrics
shown in Table~\ref{tab:base-model-comparison}.

\item For reading comprehension, we evaluate RACE using zero-shot
accuracy.

\item For multilingual capability, we evaluate Global MMLU across
German, Spanish, French, Italian, Japanese, Korean, Portuguese, and
Chinese, and report the average accuracy across these eight languages.
We evaluate MGSM~\citep{shi2022languagemodelsmultilingualchainofthought}
using 8-shot native chain-of-thought prompting in Spanish, German,
French, Chinese, Japanese, and Russian, and report the average
exact-match score across the six languages.

\end{enumerate}

Evaluation results for \ourbasemodel and \lightbasemodel are summarized in
Table~\ref{tab:base-model-comparison}.
\clearpage
\section{Post-Training}
\label{sec:posttraining}
\begin{figure*}[!t]
    \centering
    \includegraphics[width=\textwidth]{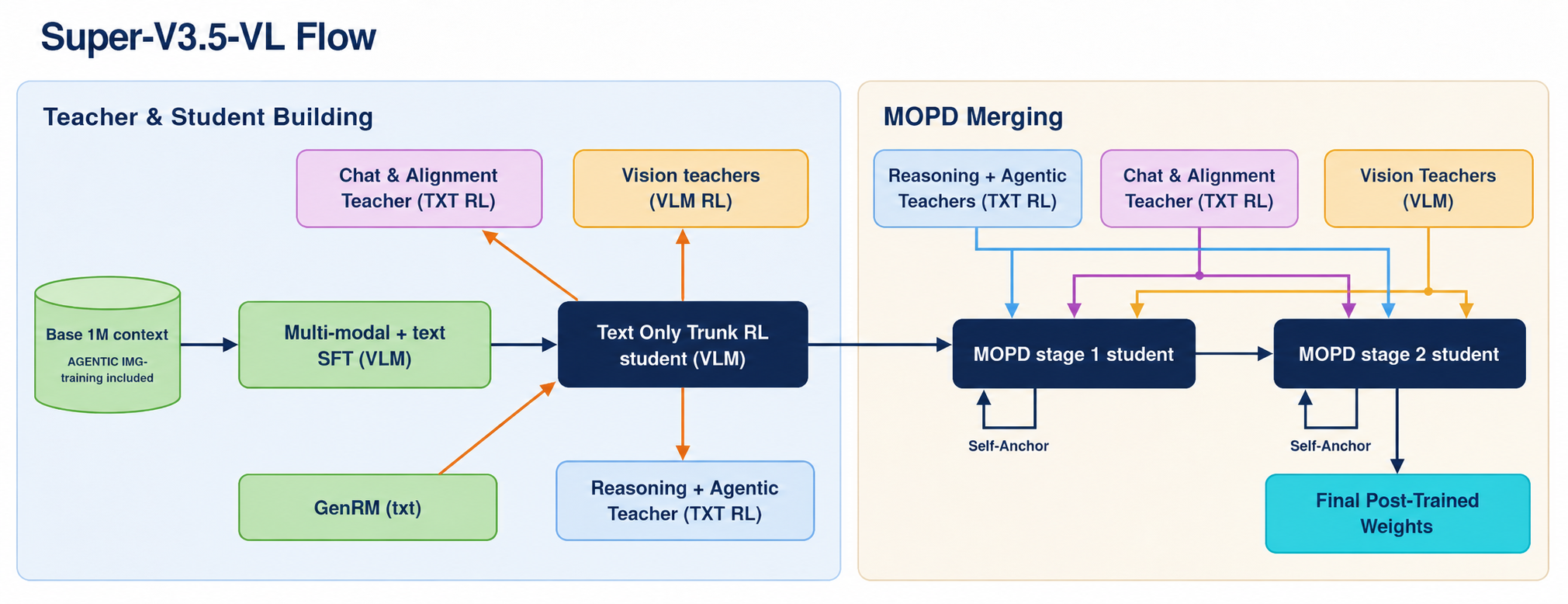}
    \caption{
        Overview of the Super-V3.5-VL post-training pipeline.
        Teacher models are developed from a shared trunk through
        supervised fine-tuning (SFT) and reinforcement learning (RL)
        stages, specializing in different capabilities.
        These teacher models are subsequently merged into a unified
        student model through MoPD (Multi-Teacher On-Policy Distillation).
    }
    \label{fig:super_v3_5_vl_flow}
\end{figure*}

Our post-training pipeline for Super is similar to that used in Nemotron 3 Ultra, consisting of SFT, RLVR, expert teacher model training, and finally MOPD~\citep{nvidia2026nemotron3ultraopen} as shown in~\ref{fig:super_v3_5_vl_flow}. We elaborate on each stage of the pipeline in the following sections. Lightning goes through a lighter pipeline, with text-only SFT, and RLVR. 

\subsection{Supervised Fine Tuning}
\label{subsec:SFT}

We train Nemotron 3.5 Super to be a VLM and Nemotron 3.5 Lightning to be a text only model. As the base model for Nemotron Super 3.5 is pre-trained on text-only data, we introduce large quantities of vision data in the SFT stage, along with agentic and reasoning text data. After warming up the vision projector \citep{nvidia2026nemotron3nanoomni}, we use a two-stage SFT process because learning vision and text capabilities simultaneously is challenging. In the first stage, we train on a general mix of vision, reasoning, chat, long-context, safety, and agentic data. We then do a small second stage of SFT, where we change the blend composition to boost some capabilities the first stage did not learn adequately. Nemotron 3.5 Lightning is trained with single SFT stage with the text only SFT data.

\begin{figure}[t]
    \centering
    \includegraphics[width=\linewidth]{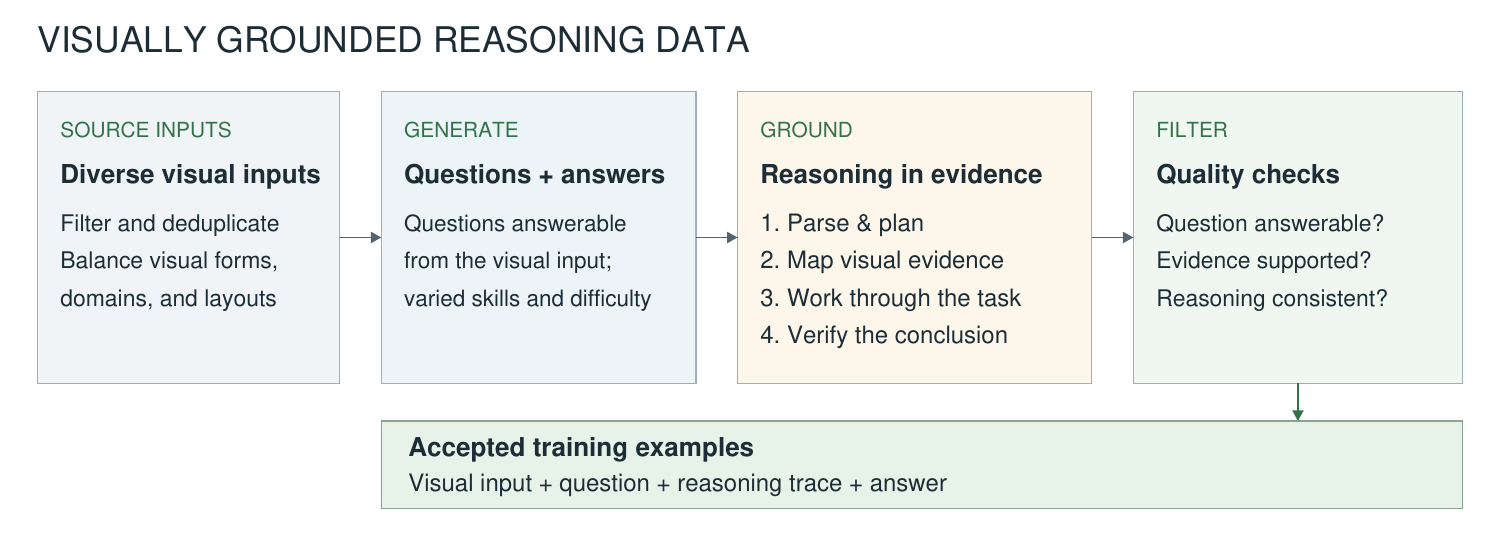}
    \caption{Synthetic vision-data generation through diverse input
    selection, question generation, grounded reasoning, and quality filtering.}
    \label{fig:vision-data-generation}
\end{figure}

\textbf{Vision Data.}
We leveraged vision data from the Nemotron 3 Nano Omni training corpus \citep{nvidia2026nemotron3nanoomni}, which covers general visual perception and VQA across charts, tables, OCR, multi-page documents, visual grounding, GUI understanding, and video understanding. It also covers visual reasoning over mathematics, science, and spatial and temporal relationships, with multilingual coverage concentrated in document and OCR tasks. We augmented this corpus with newly curated data using open-source models such as Qwen3.5 \citep{qwen2026qwen35} to strengthen chart understanding, grounded perception, multi-hop reasoning, GUI/web navigation, and multimodal coding. The data generation pipeline covers diverse visual structures, domains, question types, and reasoning tasks. We have observed that organizing the reasoning traces into an explicit sequence of concise steps, with each step grounded in relevant visual evidence, helps teach the model systematic planning and encourages iterative verification against the visual input. Additionally, we apply automated quality checks to filter ambiguous or visually unsupported questions, incomplete reasoning traces, hallucinated labels or values, incorrect visual or numerical interpretations, excessive repetition, and unproductive self-correction. Figure~\ref{fig:vision-data-generation} summarizes the pipeline. We provide a chart-understanding implementation of this approach as an open-source NeMo Data Designer generation pipeline recipe\footnote{\url{https://docs.nvidia.com/nemo/datadesigner/recipes/chart-qa/chart-question-answering}} and a companion development note\footnote{\url{https://docs.nvidia.com/nemo/datadesigner/dev-notes/chart-understanding}}.

\textbf{Agentic Vision Data.}
Visual tool use is a fundamental ability for agents. To teach the model to gain this agentic ability, we generate a comprehensive dataset spanning different areas. We use Kimi K2.6~\citep{kimiteam2025kimik2openagentic} and K3 \citep{kimiteam2026kimik3openfrontier} to generate agentic datasets with a focus on image tool operations such as crop, rotate, zoom, flip, etc for high-resolution images.
For video understanding, we use Kimi K2.6~\citep{kimiteam2025kimik2openagentic} to generate executable SpatialClaw \citep{cho2026spatialclaw} demonstrations from
existing video question--answer pairs, teaching the student to collect
visual evidence, use tools, and interpret observations before answering.
The source pools cover attribute perception, temporal understanding,
object reasoning, OCR, counting, and video synopsis.  
Each assistant action contains \texttt{Purpose}, \texttt{Reasoning},
\texttt{Next Goal}, and executable Python \texttt{Code}. Actions run in
a persistent kernel that returns execution feedback and visual observations.
The agent can inspect frames, query selected frames through a separate visual question-answering call, perform SAM3 segmentation, and reconstruct scene geometry before terminating with \texttt{ReturnAnswer}. 
We retain normally completed trajectories whose answers match the references and whose required tools execute successfully. Structured
answers use canonical matching; other free-form answers require
same-teacher, text-only equivalence judging with reported confidence
$\geq 0.90$. Subsequent filtering rejects malformed actions, execution failures, exceeded display budgets, and trajectories lacking a successful grounded action. We package the recorded conversations and
images for SFT, removing redundant formatting and messages.
The final corpus contains 14,169 trajectories and 81,366 assistant actions.
We show the performance with SpatialClaw harness in Table~\ref{tab:spatial_claw}.  

\begin{table*}[!t]
\centering
\small
\setlength{\tabcolsep}{5pt}
\renewcommand{\arraystretch}{1.3}
\resizebox{0.92\textwidth}{!}{%
\begin{minipage}{\textwidth}
\begin{minipage}[t]{0.485\textwidth}
\centering
\begin{tabular*}{\linewidth}{@{\extracolsep{\fill}}l|ccc@{}}
\toprule
\textbf{Task} & \textbf{Baseline} & \textbf{Ours} & \textbf{$\Delta$} \\
\midrule
\rowcolor{black!5}
\multicolumn{4}{l}{\strut\textbf{Single-image spatial reasoning}} \\
ERQA & 49.2 & 57.5 & +8.3 \\
Omni3D & 48.6 & 53.4 & +4.8 \\
OmniSpatial & 52.7 & 58.6 & +5.9 \\
SPBench & 60.1 & 63.8 & +3.7 \\
\midrule
\rowcolor{black!5}
\multicolumn{4}{l}{\strut\textbf{Multi-view spatial reasoning}} \\
MindCube & 55.7 & 62.2 & +6.5 \\
MMSI & 39.3 & 43.1 & +3.8 \\
SPAR-Bench & 53.9 & 64.2 & +10.3 \\
\midrule
\rowcolor{black!5}
\multicolumn{4}{l}{\strut\textbf{General spatial reasoning}} \\
BLINK & 69.1 & 71.3 & +2.2 \\
SpatialTree & 47.7 & 58.1 & +10.4 \\
ViewSpatial & 50.7 & 52.0 & +1.3 \\
\bottomrule
\end{tabular*}
\end{minipage}
\hfill
\begin{minipage}[t]{0.485\textwidth}
\centering
\begin{tabular*}{\linewidth}{@{\extracolsep{\fill}}l|ccc@{}}
\toprule
\textbf{Task} & \textbf{Baseline} & \textbf{Ours} & \textbf{$\Delta$} \\
\midrule
\rowcolor{black!5}
\multicolumn{4}{l}{\strut\textbf{Video spatial \& 4D reasoning}} \\
MMSI-Video & 32.3 & 34.5 & +2.2 \\
OSI-Bench & 30.3 & 30.6 & +0.3 \\
PAI-Bench & 57.6 & 63.8 & +6.2 \\
VSI-Bench-U & 40.7 & 48.1 & +7.4 \\
VSTI-Bench & 65.4 & 67.3 & +1.9 \\
DSI-Bench & 55.9 & 61.4 & +5.5 \\
\midrule
\rowcolor{black!5}
\multicolumn{4}{l}{\strut\textbf{General video understanding}} \\
CV-Bench & 68.0 & 69.6 & +1.6 \\
PercepComp & 39.1 & 39.4 & +0.3 \\
Video-MME & 73.5 & 74.4 & +0.9 \\
Video-MME-v2 & 20.9 & 22.2 & +1.3 \\
\midrule
\textbf{Average} & \textbf{50.5} & \textbf{54.8} & \textbf{+4.3} \\
\bottomrule
\end{tabular*}
\end{minipage}

\end{minipage}%
}
\par\vspace{3pt}
\raggedright\footnotesize
Baseline uses no agentic vision training. $\Delta$ denotes Ours minus Baseline.
\caption{Effect of agentic vision training on spatial reasoning and video understanding using the SpatialClaw harness.}
\label{tab:spatial_claw}
\end{table*}

\paragraph{Reasoning data.}
For reasoning supervised fine-tuning, we used the same underlying
reasoning problem sets as Nemotron Ultra
\cite{nvidia_nemotron_3_ultra_2026}. These problem sets span
mathematics, STEM, coding, and mathematical proof. For the mathematics
and STEM problem sets, we replaced the previous teacher with Kimi K3
\citep{kimiteam2026kimik3openfrontier} and regenerated both
chain-of-thought (CoT) and tool-integrated reasoning (TIR) response
trajectories. Mathematical-proof trajectories were generated using
DeepSeek V4 Pro \citep{deepseekai2026deepseekv4}, while
competitive-programming trajectories were generated using GLM-5.2
\citep{glm5team2026glm5vibecodingagentic}.

As part of data filtering, we removed reasoning responses shorter than 5,000 tokens to focus the training mixture on examples containing
sufficiently developed, multi-step reasoning. We also observed a difference in learning dynamics between data generated by the two primary reasoning teachers. For Nemotron Ultra, DeepSeek V4 Pro served
as the primary reasoning teacher, and downstream reasoning performance
largely plateaued after a single pass over the resulting training
trajectories. After switching the primary reasoning teacher to Kimi K3
for Nemotron 3.5 Super, performance continued to improve over multiple
passes through the regenerated trajectories. We hypothesize that the
more concise and information-dense reasoning produced by Kimi K3
requires additional exposure for a smaller student model to internalize.

\textbf{Chat and Safety Data.}
We largely follow the same recipe as Nemotron 3 Ultra, however we refresh our chat dataset with more recent Qwen, Kimi, and Gemma model responses. For safety data, we largely use the same blend as Nemotron 3 Ultra, plus some additional data focused on dual-risk CBRNE content and demographic biases. 

\textbf{Long-Context Data.} We used the 512K long-context data from Nemotron 3 Ultra~\citep{nvidia_nemotron_3_ultra_2026}. In addition, we synthetically generated graph-structured data to teach the model to trace nodes and edges through reasoning, rather than relying solely on copying and scanning. We also found that agent trajectories from long-horizon search and coding tasks can improve long-context generalization~\citep{su2026acc}. Finally, for Nemotron 3.5 Super, we synthetically generate long-context translation data from both the PG19 \citep{pg19} and FinePDFs-Edu \citep{kydlicek2025finepdfs} datasets where each instance is one long document translated in chunks from the source language to a target language using Deepseek V4 Flash \citep{deepseekai2026deepseekv4}. For detailed source and target languages as well as token counts, please see Tables~\ref{tab:pg19-language-tokens}-\ref{tab:finepdfs-edu-language-tokens}. To evaluate long-translation performance, we use the SEGALE evaluation pipeline~\citep{segale}. We found that long-translation performance was improved by also including synthetic transduction tasks~\citep{willette2026stayingtasktestingfoundations} that reinforce position tracking, retrieval, mutation, and output. Long translation results can be seen in Figure~\ref{fig:long-translation}. We also include long-context vision data generated with NeMo Data Designer \citep{nemo-data-designer} for multi-page and whole-document understanding over documents spanning up to hundreds of pages. These examples require locating and combining evidence from text, tables, charts, and diagrams across pages \footnote{\url{https://docs.nvidia.com/nemo/datadesigner/dev-notes/vlm-long-document-understanding}}.

\textbf{Training Config.}
For the SFT stage, we jointly fine-tune the language and vision components. Training uses packed sequences of up to 524,288 tokens with a global batch size of 32. The learning rate warms up over 100 packed-sample-equivalents—approximately 9,100 original records—to \(1 \times 10^{-5}\), then remains constant. We use Adam with \(\beta_1=0.9\), \(\beta_2=0.95\), weight decay 0.1, and gradient clipping at 1.0, while retaining two shared-weight MTP layers with an auxiliary-loss scaling factor of 0.3.

\subsection{Reinforcement Learning}
\label{subsec:RL}

Reinforcement learning improves the general capabilities of the SFT model and provides a strong foundation for developing specialized teacher models. At this stage, we jointly apply RLVR and RLHF, routing samples from each environment to the corresponding verifier or reward model for scoring. Compared with Nemotron 3 Ultra \citep{nvidia2026nemotron3ultraopen}, we refresh the RL training data with more recent collections and perform reward profiling before training. We use the same training algorithm as \citet{nvidia2026nemotron3ultraopen}, augmented with two optimizations designed to mitigate train–inference mismatch: router replay \citep{ma2025stabilizing} and using FP32 for LM head \citep{chen2025minimax}. Together, these optimizations reduce the KL error between the training and inference engines by more than an order of magnitude.

We also incorporate new data such as the competitive-programming RL environment of \citet{ficek2026posttraining}, comprising 2,847 training problems and 372 validation problems from past Olympiad-level competitions, each paired with a validated executable evaluator. Generated C++ solutions are compiled and executed against problem-specific tests, receiving a binary reward of 1 for full credit and 0 otherwise.

To improve tool use and cross-document reasoning on financial tasks, we built a finance-agent environment on top of NeMo Gym\footnote{\url{https://github.com/NVIDIA-NeMo/Gym/tree/main/resources_servers/finance_sec_search}}, targeting the capabilities measured by the Vals.ai Finance Agent benchmark\footnote{\url{https://www.vals.ai/benchmarks/finance_agent}}. The environment provides the same tools---SEC filing search, document parsing, evidence retrieval, and final source-supported submission---during trajectory generation, supervised fine-tuning (SFT), reinforcement learning (RL), and evaluation, avoiding format and tool mismatch across stages. We use two synthetic data generation (SDG) tracks. The document-grounded track samples windowed contexts from SEC 10-K, 10-Q, and 8-K filings of S\&P~500 companies, generates analyst-style questions with a teacher model, and verifies answerability and answer quality using multi-seed voting and GenSelect~\citep{toshniwal2025genselectgenerativeapproachbestofn}. The agentic track targets gaps found in a coverage audit against the Vals.ai category taxonomy by making the relationship among source documents---reporting period, form type, and company---an explicit sampling variable. Validated questions are re-solved by teacher models in the same environment to produce verified tool-use trajectories, and are routed to the SFT or RL pool according to their measured difficulty.

To support training across a broad and diverse set of environments, we use a global batch size of 8,192, with 16 rollouts generated per sample. The maximum generation length is initially set to 80K tokens and later increased to 128K.




\subsection{Agentic and reasoning teacher model} 

\subsubsection{Agentic Teacher}

We initialize the Agentic Teacher from the step-120 text-RLVR checkpoint and continue training with full end to end GRPO on agentic tasks. We do not use any single step environments during the training of the agentic teacher, rather opting to train the end to end agentic teacher on sandbox, fully executable agentic Terminal, SWE, and Search tasks.

We use 64 prompts and 16 generations per prompt, giving a total of 1024 batch size. We keep a max trajectory age of one for the training of the agentic teacher. Each task was given a maximum timeout of 2 hours per rollout. Tasks were profiled for difficulty for the step-120 text-RLVR checkpoint. 

\subsubsection{Reasoning teacher}
We initialize the reasoning teacher from the step-120 text-RLVR
checkpoint and continue training with the same asynchronous GRPO
framework. The main differences are the training mixture and the
generation budget. We replace the broad RLVR mixture with a
reasoning-focused curriculum of 25{,}600 examples, with each
256-prompt batch containing 90\% science, 5\% mathematics, and 5\%
competitive-coding problems. The maximum generation length is set to 131{,}072 throughout training.

We use 256 prompts per update and 16 responses per prompt,
giving 4{,}096 total batch size. We increase the maximum
trajectory age from one to two updates compared to the initial RLVR setup. The learning rate remains
$4\times10^{-6}$. Mathematics and STEM rewards use answer verification
with DeepSeek-V4-Flash\footnote{https://huggingface.co/deepseek-ai/DeepSeek-V4-Flash} as the LLM judge, while coding rewards use
execution-based tests. We select the checkpoint after 80 updates
as the reasoning teacher for the MOPD stage.

\subsection{Vision teacher models} 

\label{sec:posttraining-vision-teachers}

  \subsubsection{Video reasoning and tracking teacher}
  \label{sec:posttraining-video-teacher}
 We train a video teacher on a mixture of video-QA reasoning data which is carefully curated from video training subset of  Nemotron 3 Nano Omni \citep{nvidia2026nemotron3nanoomni} and SA-V object-tracking data
  \citep{ravi2024sam2}. Video-QA examples contain multiple-choice
  questions over 64 ordered frames, which are processed as a single video. SA-V
  examples contain box, point, and referring-object tracking tasks whose frames
  are processed as ordered multi-image inputs. This mixture targets complementary
  high-level video-reasoning (\textit{e.g.}~temporal, attribute, action and object) and fine-grained spatiotemporal-tracking
  capabilities.

  Both sources use deterministic, verifiable task rewards rather than an LLM
  judge. Video-QA uses a binary reward based on multiple-choice answer
  correctness. For SA-V, the model generates structured tracks over the requested
  frames, and the verifier assigns a continuous reward based on localization
  quality, object-visibility accuracy, and the validity and completeness of the
  predicted track. Missing, invalid, or extraneous track entries are penalized.

  We additionally apply two auxiliary shaping signals. First, a response without
  exactly one properly matched \texttt{<think>...</think>} block is considered
  malformed: its task reward is set to zero and its training advantage is
  overridden to $-5$. This prevents an otherwise correct answer with malformed
  reasoning structure from being reinforced. Second, we apply a soft length
  preference only to correct responses. Reasoning up to 1{,}024 tokens retains
  the full task reward; between 1{,}024 and 4{,}096 tokens, the reward multiplier
  decreases linearly from $1.0$ to $0.95$; and beyond 4{,}096 tokens it remains
  at $0.95$. This profile band does not truncate generation and introduces at
  most a 5\% preference for concise, correct reasoning.

  We initialize from the step-120 text-RLVR checkpoint and train using
  asynchronous GRPO \citep{deepseek-math} in NeMo-RL \citep{nemo-rl}. Each update
  uses 128 prompts and 16 responses per prompt, giving a global batch of 2{,}048
  trajectories. We replay the MoE expert-routing decisions recorded during
  rollout generation when evaluating the policy, reducing train--generation
  log-probability discrepancies. The MoE router and vision modules remain frozen.
  Training uses a constant learning rate of $3\times10^{-6}$ on 128 NVIDIA GB200
  GPUs.

  We select step 60 as the final teacher. It achieves 74.90 on Video-MME under
  the 256-frame reasoning protocol \citep{fu2025video}, 19.10 on Video-MME-v2
  under the corresponding 256-frame reasoning protocol \citep{fu2026video},
  52.59 on GOT-10k \citep{huang2021got10k}, and 60.00 on Molmo2-VideoTrack
  \citep{molmo2openweightsdata}. This checkpoint is subsequently used as the
  video-specialized teacher during MOPD.

\subsubsection{Vision tool use teacher}
To promote effective tool use in complex vision tasks, we train a vision tool-calling teacher on a curated mixture of internal data. The target tools include image cropping, zooming, rotation, flipping, difference detection, side-by-side comparison of images, overlays, object counting, and color analysis, with all tool calls represented as structured XML. The training set consists of synthetically generated multiple-choice and numerical questions that require one or more of the target tools to solve. We further filter the data by difficulty, retaining samples that are nearly impossible to solve correctly without using the available tools. We use outcome correctness as the primary reward, as well as an auxiliary tool-use reward to penalize duplicate and malformed tool calls.
We initialize from the text-RLVR checkpoint and train using asynchronous GRPO \citep{deepseek-math} in NeMo-RL \citep{nemo-rl}. The teacher checkpoint selection is guided by the target vision tool use benchmark accuracy on V* \cite{vstar}, HRBench8K \cite{hrbench}, and BabyVision \cite{babyvision}..

  \subsubsection{Vision reasoning RLVR}
  To improve reasoning over visual inputs, we train on a curated mixture of chart understanding, document and table reasoning, OCR, geometry, and visual reasoning tasks. The first stage draws from 16,090 profiled examples. The second stage uses a  mixture of 13,848 examples, comprising 8,366 chart-reasoning questions from Chart datasets, together with 5,482 examples emphasizing document, table, and OCR understanding. We audit reference-answer formats and
  compatibility with the reward functions, and remove exact question--answer duplicates from the second-stage mixture. We use outcome correctness as the training reward, evaluated using task-appropriate mathematical, multiple-choice, and string-matching verifiers.

  We initialize from an existing multimodal RLVR checkpoint and train using asynchronous GRPO \citep{deepseek-math} in NeMo-RL \citep{nemo-rl}, with 256 prompts per update and 16 sampled responses per prompt. The second stage initializes from
  the first-stage checkpoint at step 114 with a fresh optimizer. We use the checkpoint obtained after 20 additional updates on the second-stage mixture. Both stages use 128 GPUs distributed across 32 compute nodes.

\subsection{Alignment teacher}
We train an alignment teacher that has superior chat capabilities and is well calibrated with human preferences. We train this teacher primarily using RLHF. First, we train a new variant of the Nemotron 3 Ultra GenRM that is trained with a reasoning length penalty to be more token efficient. Combined with MTP inference, it enables substantially faster training. We then conduct policy training in a similar manner to Nemotron 3 Ultra. However, rather than doing self-comparison in a cyclical format, we instead compare all policy responses to a fixed response from a strong open-source model, Gemma 4 31B-it. We also use the mean of per rubric scores to evaluate responses instead of an overall score. Our training blend is similar to Nemotron 3 Ultra but includes newly collected preference data, updated model identity and capabilities data, and has more emphasis on addressing language mixing. We mix this RLHF training with the abstention training strategy from Nemotron 3 Ultra and use a small amount of reasoning length penalty. We evaluate downstream policy improvements on our internal chat benchmark.

\subsection{MOPD}
\label{subsec:MOPD}

\ourmodel{} included a multi-teacher on-policy distillation (MOPD) stage after RL which
distills the capabilities of the specialized teachers of the preceding
subsections into a single student model similar to the development of Nemotron 3 Ultra~\citep{nvidia2026nemotron3ultraopen}; the final MTP boosting phase
(Section~\ref{subsec:mtp_boosting}) then runs on the MOPD output. 
Nemotron 3.5 Lightning, however did not have an MOPD stage. In this section we detail the MOPD process for \ourmodel{}.
The student is the RL checkpoint of Section~\ref{subsec:RL} after 120 steps, the same
checkpoint the teachers were branched from. Given a prompt from some domain,
the student generates a response; for each token $x$ with context $c$ (the
prompt and preceding tokens), the teacher that owns the domain is consulted
for its probability $p_{T}(x \mid c)$, and the student is updated to raise the
probability of tokens the teacher rates higher than it does and to lower it
for tokens the teacher rates lower. Concretely, let $\pi_{S}$ be the student
with parameters $\theta$, $\pi_{T}$ the domain teacher, and $c_t$ the context
of token $y_t$ in the student's response $y$. The student minimizes the
reverse KL $D_{\mathrm{KL}}(\pi_{S}(\cdot|c_t)\,\|\,\pi_{T}(\cdot|c_t))$ at
every position of its own responses, using the token-local
update~\citep{lu2025onpolicydistillation,yang2026nemotron,xiao2026mimo} and
treating the sampled contexts as fixed during each update,
\begin{align}
\Delta\theta\;&\propto\;\mathbb{E}_{y\sim\bar{\pi}_{S}(\cdot|q)}\Big[\sum_{t}\widehat{A}_t\,\nabla_\theta\log\pi_{S}(y_t|c_t)\Big]\Big|_{\theta=\bar{\theta}},
\label{eq:super_mopd_update}\\
\widehat{A}_t\;&=\;\mathrm{sg}\!\left[\log\pi_{T}(y_t|c_t)-\log\bar{\pi}_{S}(y_t|c_t)\right],
\label{eq:super_mopd_adv}
\end{align}
where $\bar{\pi}_{S}$ is the frozen snapshot of the student taken before the
update (parameters $\bar{\theta}$), $\mathrm{sg}[\cdot]$ denotes
stop-gradient, and the expectation is over fresh responses sampled from
$\bar{\pi}_{S}$ with the update evaluated at $\theta=\bar{\theta}$. Every
generated token receives a dense advantage from a single teacher forward pass,
with no verifier and no sampling from the teacher; the direct contribution is
zero where student and teacher agree, so the signal concentrates where the
student departs from the specialist. In practice training is asynchronous and
responses may come from the previous snapshot $\bar{\pi}_{S-1}$, i.e.\ one
step stale; following the Ultra recipe, tokens are then importance-weighted by
$\bar{\pi}_{S}/\bar{\pi}_{S-1}$ and those with weights outside $[0.2, 5]$ are
masked~\citep{ring1t}. These token-level weights correct action probabilities
at the sampled contexts; context-distribution mismatch and masking introduce
bias. Only student-generated tokens enter the loss.

MOPD for \ourmodel{} is run in two stages on 128 GB200 nodes. We found
that benchmarks improve at different rates under distillation, and that some
peak within a few steps and then decline with further training. To mitigate
this, stage 1 is weighted toward the domains that improve quickly and tend to
regress later, and stage 2 toward the domains that improve more slowly.
Prompts come from the environments used to train each teacher and are routed
to that teacher; environments no specialist owns are routed to a
\emph{self-anchor}, a frozen copy of the student's starting checkpoint, which
provides no signal while the student still matches it and regularizes
departures from it.

\paragraph{Stage 1.}
The student is the RL checkpoint. It is distilled from the agentic, reasoning,
alignment (RLHF) and vision tool-use teachers with prompt shares of 35\%,
25\%, 20\% and 15\%, plus 5\% routed to the RL checkpoint as self-anchor. Batches
contain 512 prompts with one rollout each, generation is capped at 48K tokens
in a 128K context, and the learning rate is $2\times10^{-6}$. We take the
checkpoint after 30 steps as the stage-1 model.

\paragraph{Stage 2.}
Stage 2 targets chat quality (e.g.\ LMArena-style evaluations) and the vision
domains, which improved more slowly in stage 1 than the agentic and reasoning
benchmarks. The stage-1 model is the student and
self-anchor. The agentic and reasoning teachers are retired and their
environments folded into the self-anchor stream, so the student is regularized
toward the stage-1 model on them rather than distilled again. Four specialists
are distilled: the alignment teacher,
the vision tool-use teacher, a vision-reasoning teacher, and the video
reasoning and tracking teacher of Section~\ref{sec:posttraining-video-teacher},
with shares of 30\%, 20\%, 20\% and 20\%, plus 10\% self-anchor. Batches grow
to 1{,}024 prompts, generation may use the full 128K context, the learning
rate is $3\times10^{-6}$, and the stage runs for 40 steps, one pass over its
prompts.

In stage 2 the distillation target is smoothed with the proximal-teacher
construction of trust-region policy distillation~\citep{xie2026topd} (see also
\citet{xing2026tropd}): the teacher in Equation~\ref{eq:super_mopd_adv} is
replaced by the frozen mixture
$\alpha\,\pi_{T}+(1-\alpha)\,\bar{\pi}_{S}$ with $\alpha=0.2$. This
bounds the advantage below by $\log(1-\alpha)\approx-0.22$, so tokens the
teacher considers very unlikely receive a bounded negative signal rather than
an unbounded one, which stabilizes training once the student has already
absorbed one round of teachers. Stage 1 uses $\alpha=1$.

Teachers run as independent, non-colocated inference fleets (tensor parallel
8, expert parallel 8) sized to each domain's share and sequence length, and
score trajectories as they arrive, so rollout generation, teacher scoring and
student optimization form a pipeline in NeMo-RL~\citep{nemo-rl}. The MTP heads
are trained jointly with the language model throughout
(Section~\ref{subsec:mtp_boosting}).

\subsection{MTP Boosting}

\label{subsec:mtp_boosting}
We apply a dedicated MTP-only training recipe at two points: first as a short warm start from the \ourmodel{} SFT checkpoint, and again after all backbone post-training stages are complete. We refer to the latter phase as \emph{MTP boosting}. For Nemotron~3.5 Lightning, we skip the warm start and apply only MTP boosting. In both runs, the backbone remains frozen and only the MTP head parameters are updated.

The recipe has two complementary components. First, we continue applying the hidden-state-mixing strategy described in Section~\ref{sec:mtp_init}. This further exposes the MTP head to inference-like conditions, reducing the train--inference mismatch and improving acceptance at longer draft lengths. Second, whereas the preceding MTP training stages use cross-entropy loss, these runs use a knowledge-distillation loss with the frozen backbone as the teacher. This teacher-distribution supervision improves alignment between the MTP head and the backbone and is particularly beneficial under non-greedy sampling.

In order to measure MTP draft quality we report evaluation results on the SPEED-Bench~\citep{2026speedbench} \textit{qualitative} split using per-sample acceptance length (AL). 

\paragraph{Data.}
We generate on-policy rollouts using the corresponding backbone checkpoint. Seed prompts are drawn from \textit{Nemotron-Post-Training-Dataset-v2}\footnote{\url{https://huggingface.co/datasets/nvidia/Nemotron-Post-Training-Dataset-v2}} and \textit{Nemotron-RL-Super-Training-Blends}\footnote{\url{https://huggingface.co/datasets/nvidia/Nemotron-RL-Super-Training-Blends}} to cover general-purpose and agentic inputs, respectively. We sample rollouts at \(temp=1\), cap sequences at 8K tokens, and accumulate the loss only over assistant-response tokens.

\paragraph{Training Procedure.}
We first train the MTP head for $4k$ steps from the \ourmodel SFT checkpoint. This intermediate run plateaus after substantially fewer training iterations than the final-checkpoint run. When transferred to the final post-trained checkpoint, the resulting MTP head retains significant part of its acceptance-length improvement, as shown in Table~\ref{tab:sft_mtp_boosting_al}. We therefore use the SFT-trained MTP head to initialize the final MTP boosting phase and train it for an additional $25k$ steps against the final \ourmodel backbone. For Nemotron 3.5 Lightining, we ran MTP boosting for $48k$ steps against the final backbone. Whereas Nemotron 3 Ultra uses a temperature-scaled forward-KL objective, we now use the LK loss~\citep{samarin2026lk}.


\begin{table}[ht!]
\centering
\setlength{\tabcolsep}{6pt}
\begin{tabular}{lcc}
\toprule
Model & $temp=0$ & $temp=1$ \\
\midrule
\ourmodel (SFT checkpoint) & 4.589 & 4.048 \\
+ 4k MTP-only training iterations & 5.080 & 4.730 \\
\midrule
\ourmodel (Final)                   & 4.287 & 3.787 \\
+ MTP head from the SFT-stage run           & 4.528 & 4.436 \\
+ 25k MTP Boosting iterations   & 4.718  & 4.612 \\
\bottomrule
\end{tabular}
\caption{Effect of MTP Boosting on average acceptance length (AL), at draft length 7.
\emph{Top}: MTP training applied directly to an SFT checkpoint. \emph{Bottom}: \ourmodel evaluated with the native MTP, the resulting MTP head from the SFT-stage run, and after $25k$ of boosting iterations. $temp=0$ is greedy decoding; $temp=1$ uses probabilistic
draft sampling. For each training phase we generated data using the corresponding backbone.}
\label{tab:sft_mtp_boosting_al}
\end{table}

\begin{figure}[t]
\centering
\includegraphics[width=0.5\linewidth]{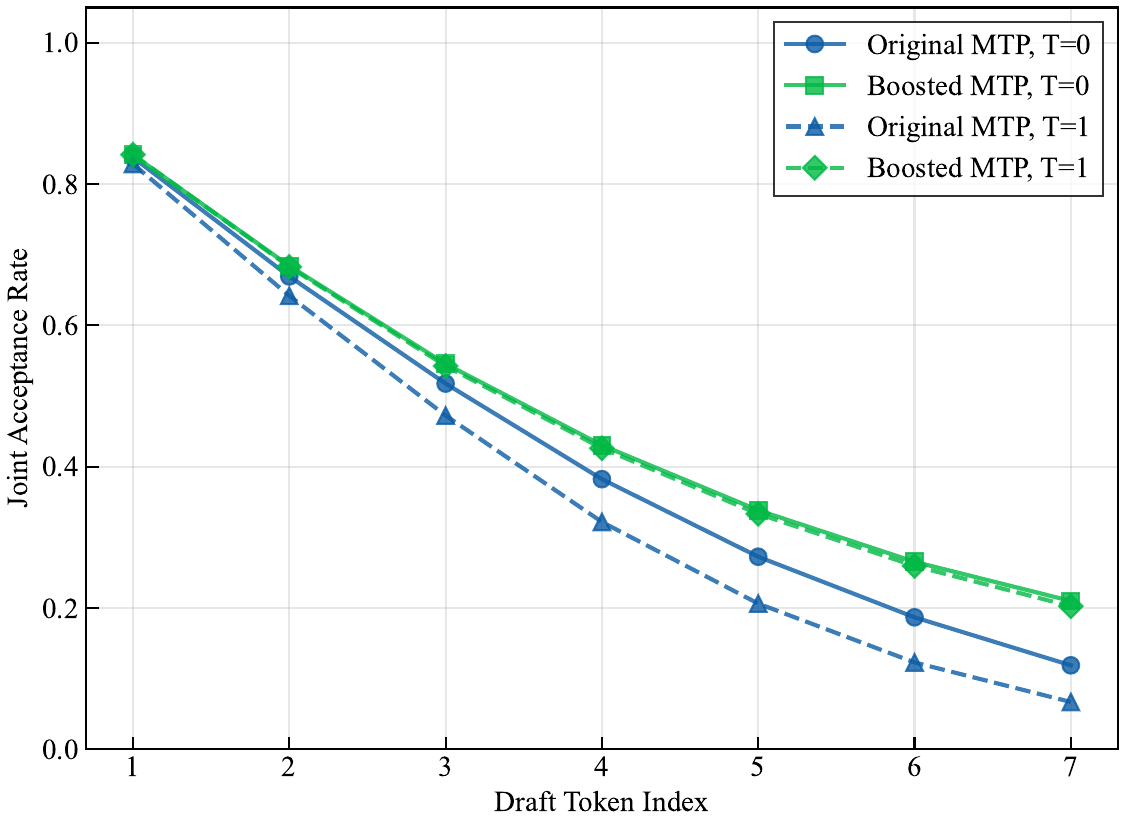}
\caption{Joint MTP acceptance rate at each draft position on the
SPEED-Bench \textit{qualitative} split (draft length 7).} 
\label{fig:unconditional_ar}
\end{figure}

\paragraph{\ourmodel{} Results.}
The Boosted-MTP head improves over the original MTP at every draft position
(Figure~\ref{fig:unconditional_ar}), with the largest gains at deeper
positions, where the train--inference mismatch is most severe: at
inference, these positions condition on the head's own hidden states
rather than on the backbone's. These per-position gains translate into higher AL in every SPEED-Bench
category (Table~\ref{tab:mtp_speedbench}), raising the average AL at $T=1$ from 3.78 to 4.61 accepted tokens.

\begin{table}[t]
\centering
\footnotesize
\setlength{\tabcolsep}{4pt}
\resizebox{1\linewidth}{!}{
\begin{tabular}{l|cc|cc|ccc}
\toprule
\multirow[b]{2}{*}{Category}
& \multicolumn{2}{c|}{\ourmodel}
& \multirow[b]{2}{*}{\shortstack{Gemma-4\\31B}}
& \multirow[b]{2}{*}{\shortstack{Qwen3.5\\122B-A10B}}
& \multicolumn{3}{c}{Qwen3.8-Flash-Next} \\
\cmidrule(lr){2-3} \cmidrule(lr){6-8}
& Native & Boosted MTP & & & low & medium & xhigh \\
\midrule
Coding        & 4.966 (4.343) & 5.563 (5.441) & \textbf{5.792} (\textbf{5.459}) & 5.630 (5.221) & 5.506 (5.310) & 5.355 (5.124) & 3.975 (3.714) \\
Humanities    & 3.703 (3.245) & 3.975 (3.817) & 4.488 (\textbf{4.039}) & \textbf{4.584} (3.956) & 4.122 (3.763) & 4.073 (3.628) & 3.873 (3.005) \\
Math          & 4.533 (4.093) & 4.841 (4.825) & \textbf{5.676} (\textbf{5.321}) & 5.149 (4.826) & 5.207 (4.982) & 5.169 (4.815) & 3.942 (3.672) \\
Multilingual  & 4.945 (4.266) & 5.548 (5.358) & \textbf{5.703} (\textbf{5.435}) & 4.995 (4.613) & 5.056 (4.911) & 4.942 (4.678) & 4.218 (3.926) \\
QA            & 4.083 (3.624) & \textbf{4.526} (\textbf{4.378}) & 4.366 (4.092) & 4.513 (4.111) & 3.988 (3.913) & 4.096 (3.748) & 3.626 (3.322) \\
RAG           & 5.054 (4.424) & \textbf{5.538} (\textbf{5.496}) & 5.532 (5.338) & 5.233 (4.986) & 5.453 (5.146) & 5.469 (5.185) & 4.424 (3.981) \\
Reasoning     & 4.314 (3.919) & 4.791 (\textbf{4.751}) & \textbf{5.086} (4.715) & 4.747 (4.427) & 4.545 (4.358) & 4.463 (4.213) & 3.797 (3.619) \\
Roleplay      & 3.444 (3.046) & 3.781 (\textbf{3.703}) & 3.475 (3.348) & \textbf{4.614} (3.331) & 2.657 (2.523) & 2.605 (2.457) & 2.881 (2.584) \\
STEM          & 3.761 (3.406) & 4.102 (4.045) & \textbf{4.858} (\textbf{4.464}) & 4.640 (4.210) & 4.556 (4.135) & 4.414 (4.038) & 3.614 (3.182) \\
Summarization & 4.716 (4.110) & \textbf{5.235} (\textbf{5.041}) & 4.536 (4.438) & 4.893 (4.416) & 4.248 (4.135) & 4.288 (4.113) & 3.627 (3.517) \\
Writing       & 3.646 (3.183) & \textbf{4.001} (\textbf{3.873}) & 3.697 (3.566) & 3.787 (3.549) & 3.340 (3.162) & 3.255 (3.045) & 3.424 (2.968) \\
\midrule
Average       & 4.288 (3.787) & 4.718 (\textbf{4.612}) & \textbf{4.837} (4.565) & 4.799 (4.331) & 4.425 (4.213) & 4.375 (4.095) & 3.764 (3.408) \\
\bottomrule
\end{tabular}}
\captionof{table}{MTP average acceptance lengths on the SPEED-Bench \textit{qualitative} split using draft length 7. Main values use greedy decoding with $temp=0$; parenthetical values use $temp=1$ with probabilistic draft sampling. Reasoning is turned on for all models. For \ourmodel, \textit{Native} is the pre-boosting MTP; for Qwen3.8-Flash-Next the sub-columns give the reasoning effort. \textbf{Bold} marks the best value per row, computed separately for $temp=0$ and $temp=1$.}
\label{tab:mtp_speedbench}
\end{table}

\subsection{Post-trained Model Evaluations}
\label{subsec:final_model_evals}
\begin{table*}[ht!]
  \centering
  \scriptsize
  \setlength{\tabcolsep}{3pt}
  \caption{Evaluation suite for \ourmodel. We compare against six open models. All SWE evaluations use OpenCode as a default harness. Terminal Bench 2.1 uses Terminus 2 and a 3 hours task timeout.}
  \label{tab:posttrained:main}
  \begin{tabular}{l|rrrrrrr}
    \toprule
    \textbf{Benchmark}
    & \makecell{\textbf{N3.5-Super}\\\textbf{120B-A12B}} 
    & \makecell{\textbf{N3-Ultra}\\\textbf{550B-A55B}} 
    & \makecell{\textbf{Qwen3.5}\\\textbf{122B-A10B}} 
    & \makecell{\textbf{Qwen3.8}\\\textbf{Flash-Next}} 
    & \makecell{\textbf{DS-V4}\\\textbf{Flash-0731}} 
    & \makecell{\textbf{Inkling}\\\textbf{Small}} 
    & \makecell{\textbf{GLM-5.3}\\\textbf{Flash}} \\
    \midrule
    \multicolumn{8}{l}{\textbf{Agentic}} \\
    \cmidrule(lr){1-8}
    SWE-bench Verified      & 71.10     & 73.93     & 71.87     & 72.60     & 77.80     & 75.27         & 79.10 \\
    SWE-bench Multilingual  & 67.72  & 68.78     & 62.00     & 79.00     & 72.00     & 71.44         & 78.30 \\
    SWE-bench Pro           & 51.78 & 49.66     & 45.60     & 60.37     & 56.91     & 56.54         & 59.70 \\
    Tau2 Retail             & 69.24     & 73.68     & 55.37     & 71.60     & 73.79     & 71.82         & 78.18 \\
    Tau2 Airline            & 76.50     & 79.75     & 63.50     & 80.75     & 83.50     & 62.75         & 81.75 \\
    Tau2 Telecom            & 73.19     & 83.00     & 87.17     & 90.35     & 98.25     & 90.12         & 98.46 \\
    Tau3 Banking            & 18.88     & 18.35     & 14.23     & 44.54     & 38.56     & \textemdash{} & 34.85 \\
    Terminal-Bench 2.1      & 60.67 & 54.78     & 48.70     & 85.80     & 76.80     & 53.40         & 82.70 \\
    GDPVal-AA-v2 ELO        & 1,211.20  & 1,148.30  & 860.73    & 1,561.69  & 1,382.30  & 1,313.27      & 1,563.39 \\
    GDPVal-AA-v2 norm ELO   & 0.356   & 0.324     & 0.180     & 0.531     & 0.441     & 0.407         & 0.532 \\
    BrowseComp              & 78.91     & 67.85 & 58.10 & 89.02     & 84.20     & 74.10         & 86.57 \\
    BrowseComp (OpenCode)            & 72.35     & 61.06 & 45.42 & 76.07     & 64.77     & 55.37         & 81.91 \\
    \addlinespace[0.5em]
    \multicolumn{8}{l}{\textbf{Code Generation}} \\
    \cmidrule(lr){1-8}
    SciCode (subtask)           & 47.13 & 41.96 & 39.83 & 46.62 & 49.30 & 48.28 & 49.96 \\
    SciCode (problem)           & 15.10 & 13.46 & 10.77 & 14.42 & 18.27 & 18.65 & 17.50 \\
    LiveCodeBench v6 & 88.22 & 89.21 & 76.65 & 94.13 & 94.64 & 88.25 & 80.62 \\
    
    \addlinespace[0.5em]    
    \multicolumn{8}{l}{\textbf{Math and Science}} \\
    \cmidrule(lr){1-8}
    GPQAD                    & 89.52 & 86.99 & 86.74     & 92.68         & 89.65 & 88.57 & 91.22 \\
    CritPt                              & 10.86 & 3.14  & 1.14      & 15.70 & 12.57 & 9.14  & 9.70 \\
    HLE (no tools, text)                     & 31.79 & 29.24 & 24.24     & 39.57         & 35.82 & 33.27 & 36.47 \\
    HLE (with tools, text)                    & 46.01 & 37.40 & 47.50     & \textemdash{} & 45.10 & 47.80 & 55.30 \\
    Apex-Shortlist                     & 88.36 & 75.00 & 54.65     & 91.76         & 88.70 & 71.54 & 77.13 \\
    AA-Omniscience              &  &  & & & & & \\

    \quad Accuracy             & 26.20 & 25.30 & 26.25     & 27.87         & 44.95 & 35.38 & 34.22 \\
    \quad Non-Hallucination           & 69.96 & 77.91 & 15.5      & 58.69         & 4.09  & 40.39 & 69.75 \\

    \addlinespace[0.5em]
    \multicolumn{8}{l}{\textbf{Multimodal}} \\
    \cmidrule(lr){1-8}
    BabyVision          & 29.60 & \textemdash{} & 35.65 & 46.82 & \textemdash{} & 24.74         & 39.09 \\
    CharXiv-RQ          & 80.82 & \textemdash{} & 83.47 & 85.77 & \textemdash{} & 77.77         & 81.93 \\
    MathVision          & 78.16 & \textemdash{} & 83.29 & 87.54 & \textemdash{} & 78.77         & 82.52 \\
    OCR-Reasoning       & 65.08 & \textemdash{} & 64.27 & 67.10 & \textemdash{} & 54.69         & 64.61 \\
    OCRBench-v2 (EN)    & 68.12 & \textemdash{} & 68.27 & 67.91 & \textemdash{} & 54.21         & 61.81 \\
    OCRBench-v2 (ZH)    & 59.15 & \textemdash{} & 67.26 & 68.28 & \textemdash{} & 52.38         & 56.81 \\
    Video-MME-v2        & 32.85 & \textemdash{} & 31.20 & 45.05 & \textemdash{} & \textemdash{} & 14.79 \\
    V*                  & 85.60 & \textemdash{} & 91.10 & 93.02 & \textemdash{} & 70.51         & 88.13 \\
    HLE (full)          & 29.20 & \textemdash{} & 24.44 & 38.76 & \textemdash{} & 31.60         & 35.08 \\
    MMLongBench-Doc     & 62.82 & \textemdash{} & 63.74 & 67.93 & \textemdash{} & 49.06         & 65.83 \\
    OSWorld             & 52.14 & \textemdash{} & 51.47 & 55.76 & \textemdash{} & \textemdash{} & 68.86 \\
    
    \addlinespace[0.5em]
    \multicolumn{8}{l}{\textbf{Long Context}} \\
    \cmidrule(lr){1-8}
    AA-LCR      & 68.63 & 75.19 & 75.13 & 78.38 & 79.69 & 77.31 & 79.50 \\
    GraphWalks ($\leq$1M)  & 89.71 & 63.28 & 86.12 & 76.15 & 65.38 & 55.74 & 88.41 \\
    
    \addlinespace[0.5em]
    \multicolumn{8}{l}{\textbf{Instruction Following and Chat Preference}}\\
    \cmidrule(lr){1-8}
    IFBench         & 70.86 & 82.94 & 67.56 & 75.63 & 76.19 & 79.31 & 57.67 \\
    NV Chat Benchmark      & 52.67 & 34.27 & 46.98 & 35.16 & 49.78 & 53.60 & 61.00 \\

    \addlinespace[0.5em]
    \multicolumn{8}{l}{\textbf{Multilingual}} \\
    \cmidrule(lr){1-8}
    MMLU-ProX       & 76.91 & 83.10 & 83.19 & 82.98 & 85.53 & 83.97 & 82.99 \\
    WMT24pp COMET   & 84.80 & 82.06 & 86.34 & 86.20 & 85.50 & 84.60 & 86.13 \\
    \bottomrule
  \end{tabular}
\end{table*}

\subsubsection{Evaluation Setup}
We evaluate \ourmodel on a comprehensive suite of benchmarks spanning agentic capabilities, reasoning and knowledge, multimodal, conversational ability and instruction following, long-context understanding, and multilingual performance.
All evaluation results for \ourmodel and baselines were collected via Nemo Gym\footnote{\url{https://github.com/NVIDIA-NeMo/Gym}}. For reproducibility purposes, more details on the evaluation settings can be found in the Nemo Gym recipes folder and the corresponding reproducibility tutorial\footnote{\url{https://github.com/NVIDIA-NeMo/Gym/blob/main/nemotron_recipes/super-3.5/reproducibility.md}}.

The following benchmarks are not onboarded yet in our open source tools and for these we used either their official open source implementation or otherwise an internal scaffolding that we plan to open source in the future: BabyVision, CharXiv-RQ, MathVision, MMLongBench-Doc, OCR-Reasoning, OCRBench-v2, V*, Video-MME-v2. NV Chat Benchmark is an internal chat benchmark developed at NVIDIA.

All models were evaluated under the same evaluation settings including agentic resources (CPU compute and timeouts, tools, search APIs), input data, prompt templates, number of repeats, and extraction/metric implementation. Inference parameters such as temperature and top\_p, and maximum number of tokens were directly extracted from the recommendations of the corresponding model cards. 

\paragraph{Agentic Capabilities.}
We evaluate agentic capabilities using SWE-bench Verified~\citep{jimenez2023swe}, SWE-bench Multilingual~\citep{swesmith}, and SWE-bench Pro~\citep{deng2025swebenchpro} for software engineering; Terminal-Bench 2.1~\citep{merrill2026terminal} for terminal-based task execution; the Retail, Airline, and Telecom domains of $\tau^2$-Bench~\citep{barres2025tau} and the Banking domain of $\tau^3$-Bench~\citep{shi2026tau} for conversational tool use; GDPVal~\citep{patwardhan2025gdpval} for economically valuable real-world workplace tasks; and BrowseComp~\citep{wei2025browsecomp} for web browsing and information seeking.
All SWE-bench variants run in the OpenCode harness\footnote{\url{https://github.com/anomalyco/opencode}} inside a sandbox with web search and web fetch disabled, and we report the resolve rate averaged over three runs. Terminal-Bench 2.1 uses the Terminus 2 harness and is averaged over eight runs. All $\tau$-Bench domains are averaged over eight trials. The $\tau^2$-Bench domains use Qwen3-235B-A22B-Instruct-2507 as the user simulator; in the Banking domain the agent retrieves from the bank's knowledge base with BM25 and grep tools, and GPT-5.4-mini acts as the user simulator. 

Our open-source implementation of GDPVal loosely follows the AA v2.1 methodology using the same sandboxing environment with the respect to python and system packages, and a Stirrup agent. The reference committee includes a subset of the models from the AA leaderboard for which the reference files were collected independently (DeepSeek V4 Flash 0731, DeepSeek V4 Pro, Inkling Small, GLM-5.1, Kimi K2.6, Nemotron 3 Ultra, Qwen3.6-35B-A3B, Qwen3.5-122B-A10B, Qwen3.5-397B-A17B, GPT-OSS-120b, Gemma 4 26B-A4B, Qwen3-30B-A3B-Thinking-2507). The judge committee is the same as the one described in AA methodology (GPT-5.6 Sol, medium reasoning, Gemini 3.8 Flash, high reasoning, and Claude Opus 5, high effort). 

BrowseComp lacks a standardized harness, and scores vary substantially with search provider, browsing tools, context management, and step budget. We therefore evaluate all models on a single open-source harness released in NeMo Gym. The harness provides \texttt{search} and \texttt{browse} tools backed by Exa\footnote{\url{https://exa.ai}} (deep mode) and stores retrieved pages in a per-sample workspace accessible through a read-only shell tool. Context is managed with a discard-all policy: when the conversation exceeds a token threshold, prior turns are dropped and the model maintains progress across resets via a memory tool. We additionally report BrowseComp on an OpenCode-based harness (with minor tool-configuration changes) under identical serving settings for all models, and observe that \ourmodel's performance transfers to this second harness as well.

\paragraph{Code Generation.}
We evaluate code generation using LiveCodeBench v6~\citep{jain2024livecodebench} for competitive programming and SciCode~\citep{tian2024scicoderesearchcodingbenchmark} for scientific research coding. For LiveCodeBench, we use the 454 problems released between August 2024 and May 2025, prompt the model to solve them without tools, and report pass@1 averaged over three samples. SciCode contains 65 research problems written by scientists and decomposed into 288 subproblems; we include the scientist-provided background in the prompt and average over eight repeats. We report both subtask accuracy, the fraction of subproblems whose generated code passes its tests, and problem accuracy, the fraction of problems for which all subproblems pass.

\paragraph{Math and Science.}
We evaluate mathematical and scientific reasoning using GPQA-Diamond~\citep{rein2023gpqa} for graduate-level science questions; Humanity's Last Exam (HLE)~\citep{phan2025humanitysexam} for expert-level questions at the frontier of human knowledge; MathArena Apex-Shortlist~\citep{dekoninck2026matharena} for competition mathematics; CritPt~\citep{zhu2025critpt} for research-level physics reasoning; and AA-Omniscience~\citep{jackson2025aaomniscience} for factual knowledge and hallucination resistance.
GPQA-Diamond is averaged over eight repeats and Apex-Shortlist over 16. Apex-Shortlist answers are checked by symbolic equivalence, and the model has no tool access.
For HLE we report the 2,158 text-only questions here, both without tools and with tools. Without tools, answers are graded by GPT-5.6 Luna using the official HLE judge prompt.
In the with-tools setting, \ourmodel can run Python code and search the web; this result comes from a separate internal harness.
CritPt is scored with the official Artificial Analysis grading API, using five samples for each of its 70 problems.
For AA-Omniscience, we use the 600-question public set, average over ten repeats, and grade with GPT-5.6 Luna. We report accuracy and the non-hallucination rate. 

\paragraph{Multimodal.}
We evaluate multimodal understanding across visual reasoning, charts, documents and OCR, video, and computer use. We use BabyVision~\citep{babyvision} for basic visual reasoning; CharXiv reasoning questions~\citep{wang2024charxiv} for scientific chart understanding; MathVision~\citep{wang2024mathvision} for mathematical reasoning in visual contexts; OCR-Reasoning~\citep{huang2025ocrreasoning} and OCRBench v2~\citep{fu2024ocrbenchv2}, the latter with separate English and Chinese scores, for text-rich image understanding; V*~\citep{vstar} for fine-grained visual search in high-resolution images; Video-MME-v2~\citep{fu2026video} for video understanding; MMLongBench-Doc~\citep{ma2024mmlongbenchdoc} for long multi-page document understanding; the full HLE set~\citep{phan2025humanitysexam}, including image-based questions; and OSWorld-Verified~\citep{xie2024osworld,xie2025osworldverified} for computer-use agents.
Unless noted otherwise, results are averaged over three repeats. Free-form answers are extracted or graded by an LLM judge: GPT-4o-mini for BabyVision, MathVision, and OCR-Reasoning, Claude Opus 4.7 for CharXiv, GPT-4o for MMLongBench-Doc, and GPT-3.5-Turbo for answer extraction on V*. OCRBench v2 and Video-MME-v2 use rule-based scoring. V* is evaluated as static multiple choice without a visual search tool. Video-MME-v2 is evaluated without subtitles. MMLongBench-Doc provides up to 100 rendered pages per document, one image per page. HLE (full) covers all 2,500 questions with a single sample, graded by GPT-5.6 Luna.
For OSWorld-Verified, we run the 361 tasks that do not require Google Drive access on cloud-hosted desktop virtual machines. The agent observes screenshots only, acts through \texttt{pyautogui} commands, and has up to 100 steps per task. We report the average task score, which includes partial credit, over three runs.

\paragraph{Long Context.}
We evaluate long-context understanding with AA-LCR~\citep{artificialanalysis2025lcr} and GraphWalks~\citep{openai2025graphwalks}. AA-LCR (v1.1) contains 100 questions that require reasoning across sets of real-world documents totaling about 100K tokens. We average over 16 repeats and grade answers with GPT-5.6 Luna following the official protocol. GraphWalks gives the model a graph as an edge list in the prompt and asks for the result of a breadth-first search or for the parents of a node. We include both task types across all context-length buckets and keep the 950 of 1,150 prompts that fit within 1M tokens; all models are scored on the same 950 prompts. The score is the F1 between the predicted and reference node sets, from a single sample.

\paragraph{Instruction Following and Chat Preference.}
We assess instruction following with IFBench~\citep{pyatkin2025generalizing}, which measures how well models generalize to 58 new, out-of-domain verifiable output constraints. We report the fraction of constraints satisfied per prompt under strict verification, averaged over prompts and eight samples.
NV Chat Benchmark measures chat preference as a win rate in pairwise comparisons. It covers 3,713 prompts, and each prompt has an opponent response drawn from human-preference battles. Gemini 3.1 Pro judges each pair in both presentation orders, with style control based on response length.
In addition, Table~\ref{tab:arena} reports Text Arena~\citep{chiang2024chatbot,li2024style} Elo ratings derived from human pairwise preference votes, for all prompts and for English prompts only, with and without style control.

\begin{table*}[t]
  \centering
  \scriptsize
  \setlength{\tabcolsep}{3pt}
  \caption{Text Arena Elo scores. We report number of votes and CI from Arena. Note, N3.5-Super 120B-A12B NVFP4 are preliminary results, and Arena does not report public scores for Qwen3.8 Flash-Next.}
  \label{tab:arena}
  \begin{tabular}{l|rrrrrrr}
    \toprule
    \textbf{Text Arena}
    & \makecell{\textbf{N3.5-Super}\\\textbf{120B-A12B}\\\textbf{NVFP4}}
    & \makecell{\textbf{N3-Ultra}\\\textbf{550B-A55B}\\\textbf{NVFP4}}
    & \makecell{\textbf{Qwen3.5}\\\textbf{122B-A10B}}
    & \makecell{\textbf{DS-V4}\\\textbf{Flash}}
    & \makecell{\textbf{Inkling}\\\textbf{Small}}
    & \makecell{\textbf{GLM-5.3}\\\textbf{Flash}}
    & \makecell{\textbf{Gemma-4}\\\textbf{26B-A4B}} \\
    \midrule
    \multicolumn{8}{l}{\textbf{Overall}} \\
    \cmidrule(lr){1-8}
    With Style Control
      & \makecell{1420 $\pm$ 12\\(2,328 votes)}
      & \makecell{1426 $\pm$ 7\\(11,570 votes)}
      & \makecell{1416 $\pm$ 4\\(29,737 votes)}
      & \makecell{1436 $\pm$ 4\\(51,988 votes)}
      & \makecell{1404 $\pm$ 5\\(31,831 votes)}
      & \makecell{1474 $\pm$ 5\\(32,013 votes)}
      & \makecell{1438 $\pm$ 7\\(6,095 votes)} \\
    \addlinespace[0.2em]
    With Factuality
      & \makecell{1441 $\pm$ 20\\(2,328 votes)}
      & \makecell{1386 $\pm$ 12\\(11,570 votes)}
      & \makecell{1431 $\pm$ 9\\(29,737 votes)}
      & \makecell{1433 $\pm$ 6\\(51,988 votes)}
      & \makecell{1429 $\pm$ 7\\(31,831 votes)}
      & \makecell{1438 $\pm$ 7\\(32,013 votes)}
      & \makecell{1450 $\pm$ 20\\(6,095 votes)} \\
    \addlinespace[0.2em]
    No Adjustment
      & \makecell{1413 $\pm$ 12\\(2,328 votes)}
      & \makecell{1445 $\pm$ 7\\(11,570 votes)}
      & \makecell{1417 $\pm$ 4\\(29,737 votes)}
      & \makecell{1432 $\pm$ 4\\(51,988 votes)}
      & \makecell{1412 $\pm$ 5\\(31,831 votes)}
      & \makecell{1471 $\pm$ 5\\(32,013 votes)}
      & \makecell{1434 $\pm$ 7\\(6,095 votes)} \\

    \addlinespace[0.5em]
    \multicolumn{8}{l}{\textbf{English}} \\
    \cmidrule(lr){1-8}
    With Style Control
      & \makecell{1446 $\pm$ 20\\(884 votes)}
      & \makecell{1440 $\pm$ 9\\(5,381 votes)}
      & \makecell{1430 $\pm$ 6\\(13,158 votes)}
      & \makecell{1445 $\pm$ 5\\(23,133 votes)}
      & \makecell{1417 $\pm$ 6\\(13,047 votes)}
      & \makecell{1479 $\pm$ 7\\(11,852 votes)}
      & \makecell{1450 $\pm$ 12\\(2,249 votes)} \\
    \addlinespace[0.2em]
    With Factuality
      & \makecell{1444 $\pm$ 28\\(884 votes)}
      & \makecell{1397 $\pm$ 19\\(5,381 votes)}
      & \makecell{1431 $\pm$ 14\\(13,158 votes)}
      & \makecell{1437 $\pm$ 9\\(23,133 votes)}
      & \makecell{1435 $\pm$ 11\\(13,047 votes)}
      & \makecell{1442 $\pm$ 11\\(11,852 votes)}
      & \makecell{1454 $\pm$ 35\\(2,249 votes)} \\
    \addlinespace[0.2em]
    No Adjustment
      & \makecell{1436 $\pm$ 19\\(884 votes)}
      & \makecell{1454 $\pm$ 9\\(5,381 votes)}
      & \makecell{1428 $\pm$ 6\\(13,158 votes)}
      & \makecell{1441 $\pm$ 5\\(23,133 votes)}
      & \makecell{1421 $\pm$ 6\\(13,047 votes)}
      & \makecell{1474 $\pm$ 6\\(11,852 votes)}
      & \makecell{1444 $\pm$ 12\\(2,249 votes)} \\
    \bottomrule
  \end{tabular}
\end{table*}

\paragraph{Multilingual.}
We evaluate multilingual capabilities using MMLU-ProX~\citep{xuan2025mmluprox} for multilingual knowledge and reasoning and WMT24++~\citep{deutsch2025wmt24pp} for machine translation. For MMLU-ProX, we report accuracy averaged over ten languages: English, German, Spanish, French, Italian, Japanese, Chinese, Hindi, Portuguese, and Korean. For WMT24++, we translate from English into the nine non-English languages of this set and report reference-based XCOMET-XXL~\citep{guerreiro2024xcomet} scores, macro-averaged over language pairs.

\subsection{Reasoning Efficiency and Control}
\label{subsec:reasoning_effort}

Nemotron 3.5 Super is trained for three reasoning modes: reasoning-off, regular and medium-effort. The regular and medium-effort reasoning modes have the option to be used in conjunction with inference-time budget control. These combinations of controls provide flexibilities that cover the entire spectrum of accuracy-efficiency trade-off to meet customers’ needs in various application scenarios.

The medium-effort reasoning mode is introduced during the SFT stage, with data generated by Kimi K3's low-effort mode, and later optimized during the RL stage.
Following the same recipe as Nemotron 3 Ultra, the reasoning traces of a small fraction of SFT data are truncated to random reasoning budgets while the responses remain the same and with </think> tokens masked from training loss, to prepare the model for inference-time budget control.
Approximately 4\% of the RL training prompts are in the medium-effort mode and they cover math, STEM and coding, and length-based adjustments are applied on the RL rewards for them.
This is the same recipe that has trained Nemotron 3 Ultra and Super and the effects of effort-level optimization generalize to a variety of tasks beyond math, STEM and coding.
Nemotron 3.5 Super's medium-effort mode uses around 2.2X less tokens than the regular mode, averaged over the benchmarks in Artificial Analysis Intelligence Index, at the cost of approximately 9\% drop in accuracy.

\begin{figure*}[t]
    \centering
    \includegraphics[width=0.9\textwidth]{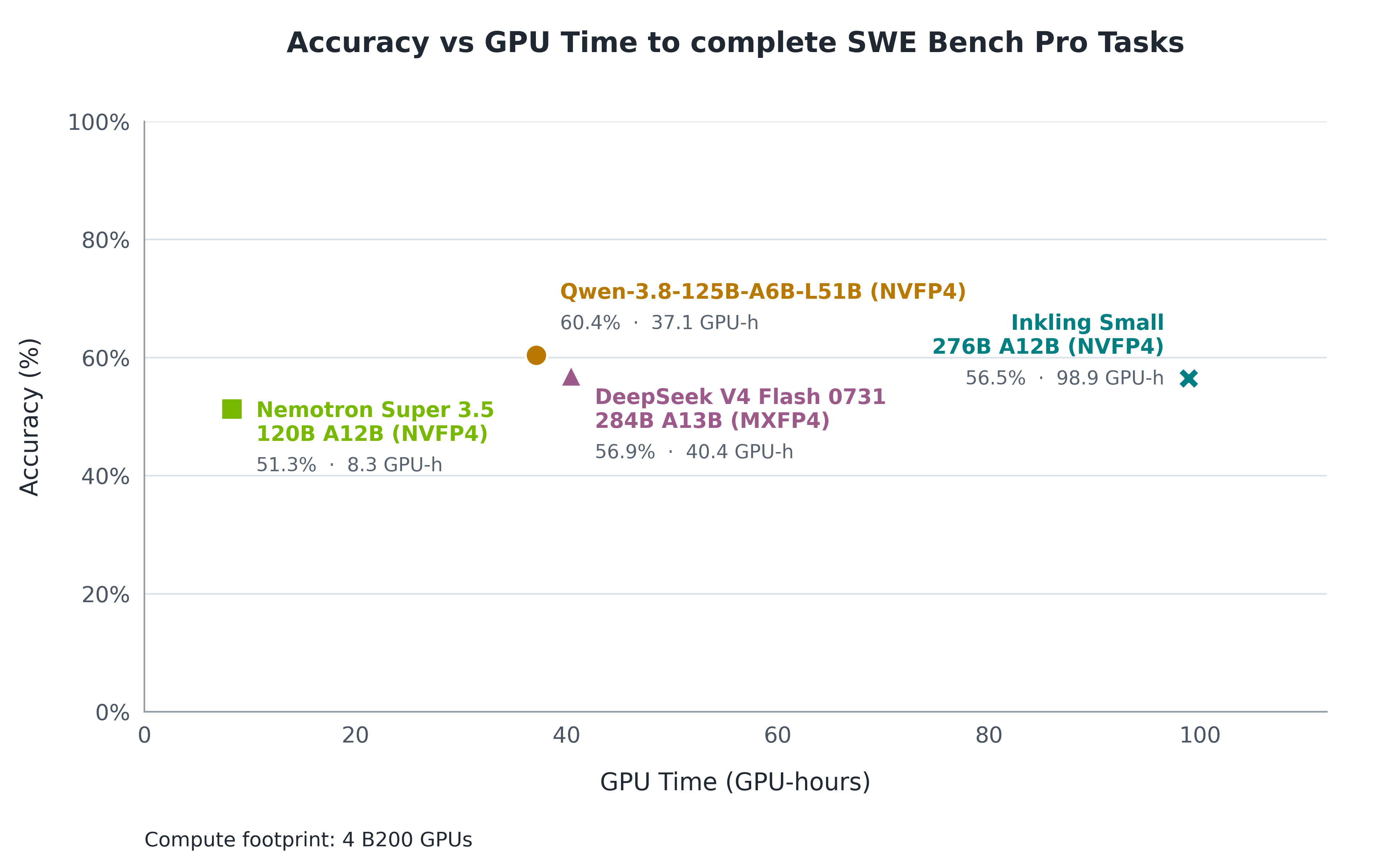}
    \caption{Accuracy--compute trade-off on SWE-bench Pro.
    We evaluate each model, record the resulting agent trajectories,
    and use the captured traces to benchmark task throughput
    (tasks per hour). The vertical axis reports Pass@1 accuracy
    averaged over three evaluation runs. The horizontal axis reports
    the aggregate GPU time, in GPU-hours, required to complete
    2,193 tasks (three repeats of the public set). The compute footprint is four NVIDIA B200 GPUs.
    All models use NVFP4 precision except DeepSeek V4 Flash 0731,
    which uses MXFP4.}
    \label{fig:accuracy-vs-gpu-time}
\end{figure*}

Figure~\ref{fig:accuracy-vs-gpu-time} compares Nemotron 3.5 Super against similar-sized models on SWE-bench Pro and demonstrates substantial efficiency advantage in GPU-hours.

\subsection{Agentic Coding Across Harnesses}
\label{sec:posttraining-agentic-harnesses}

\begin{figure}[!htbp]
    \centering
    \includegraphics[width=\linewidth]{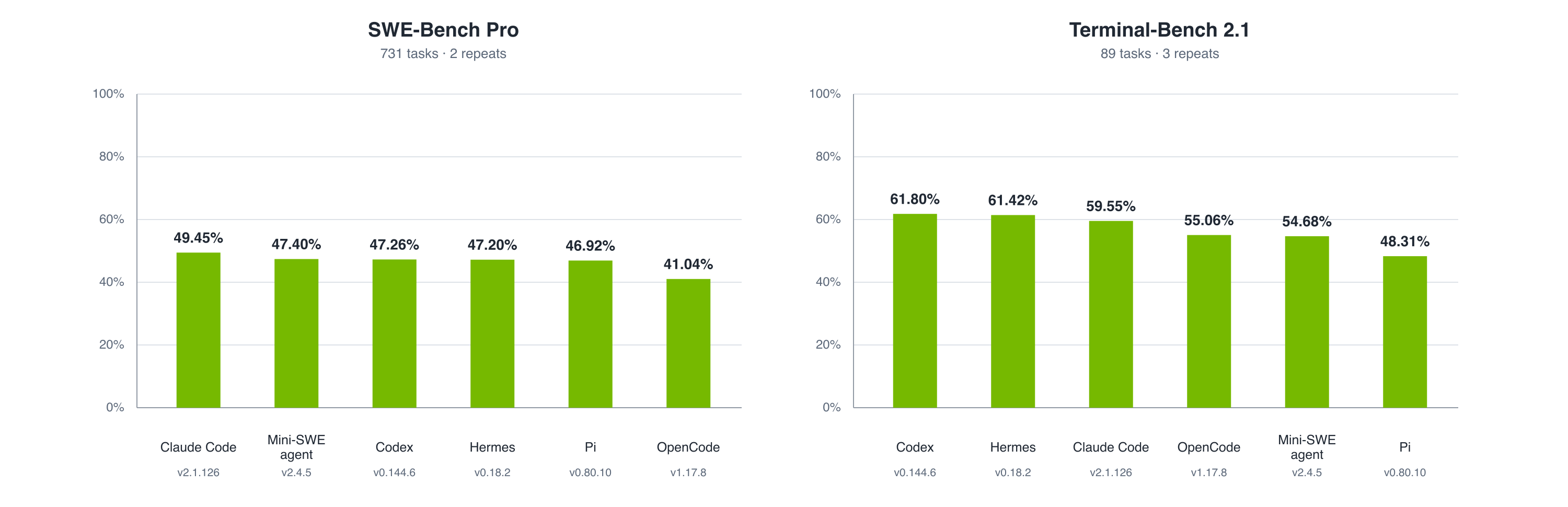}
    \caption{
        Nemotron 3.5 Super performance across six agent harnesses
        on SWE-bench Pro and Terminal-Bench 2.1, using unmodified
        harness system prompts and standard configurations.
        For SWE-bench Pro, Git history is removed and a user
        instruction prohibits cheating or seeking online resources.
    }
    \label{fig:agentic-coding-harnesses}
\end{figure}

It is important for the model to not "overfit" into one single harness. We evaluate Nemotron 3.5 Super across six agent harnesses on
SWE-bench Pro and Terminal-Bench 2.1.
Figure~\ref{fig:agentic-coding-harnesses} compares the observed
scores under each harness's standard configuration.
Claude Code achieves the highest observed score on SWE-bench Pro
(49.45\%), while Codex achieves the highest observed score on
Terminal-Bench 2.1 (61.80\%). The ordering of harnesses differs
between benchmarks, highlighting the importance of reporting
the harness and its configuration alongside model results.

\subsection{Test-Time Scaling through Parallel Agents and Self-Verification}

We further improve the final model's agentic performance through
parallel agents followed by self-verification. For each task, $N$
independent instances of the same post-trained agent attempt the task,
producing a pool of candidate trajectories. These agents can execute
in parallel without inter-agent communication. The same checkpoint
then serves as a verifier, selecting one candidate using
LLM-as-a-Verifier (LAV)
\cite{kwok2026llmasaverifiergeneralpurposeverificationframework}.
This requires neither additional parameter updates nor a separately
trained verifier.

To illustrate the selection process, consider Terminal-Bench 2.1,
where each task has five candidate trajectories. The verifier receives
the task description and a pair of trajectories, and evaluates them
against three criteria: specification adherence, output match, and
error signal detection. These criteria assess whether the agent meets
the task's explicit requirements, whether its observed terminal output
supports successful completion, and whether unresolved errors remain.
Each evaluation focuses on one criterion and assigns a separate grade
distribution to each trajectory. Following LAV, we use a 20-level
grading scale represented by single-token labels A--T, ordered from
best to worst. We compute each trajectory's score as the
probability-weighted average over these grades, normalized to $[0,1]$,
rather than selecting a single discrete grade.

For each compared pair and criterion, we evaluate both presentation
orders (once with trajectory $i$ first and $j$ second, and once with
their positions reversed; $K=2$). We align the resulting scores with
the original candidate identities and average across both orders and
all three criteria to mitigate presentation-order bias and evaluation
noise. Each compared pair therefore receives six evaluations of the
same two trajectories, without generating additional agent rollouts.
The resulting pair-specific scores $R_i$ and $R_j$ are converted into
a preference probability using the Bradley--Terry model
$p_{ij}=1/(1+\exp(-(R_i-R_j)))$ \cite{bradley1952rank}.

These pairwise evaluations are organized using LAV's Probabilistic
Pivot Tournament (PPT). For the five Terminal-Bench candidates, the
tournament first compares neighboring candidates in a random ring,
for example $(1,2)$, $(2,3)$, $(3,4)$, $(4,5)$, and $(5,1)$.
The two candidates with the highest mean preference across their
ring comparisons become pivots. Each of the remaining three
candidates is then compared against both pivots, and the two pivots
are compared with each other; cached evaluations are reused for
identical directed comparisons. Finally, the preferences from both
stages are aggregated, and the candidate with the highest mean
preference across its comparisons is selected. Normalizing by the
number of comparisons accounts for the pivots participating in more
comparisons, and any of the five candidates can ultimately win.

We apply the same procedure across all six benchmarks, using five
candidate trajectories for Terminal-Bench 2.1 and DeepSWE, and three
for SWE-bench Verified, Pro, Multilingual, and BrowseComp. All
benchmarks use three criteria and $K=2$. Terminal-Bench and the three
SWE-bench variants use two pivots, while DeepSWE and BrowseComp use
one pivot. For BrowseComp, all reported metrics use the 400-problem
subset used for our internal evaluation tracking, given the cost of
evaluating the full benchmark.

\begin{figure}[h]
    \centering
    \includegraphics[width=0.9\linewidth]{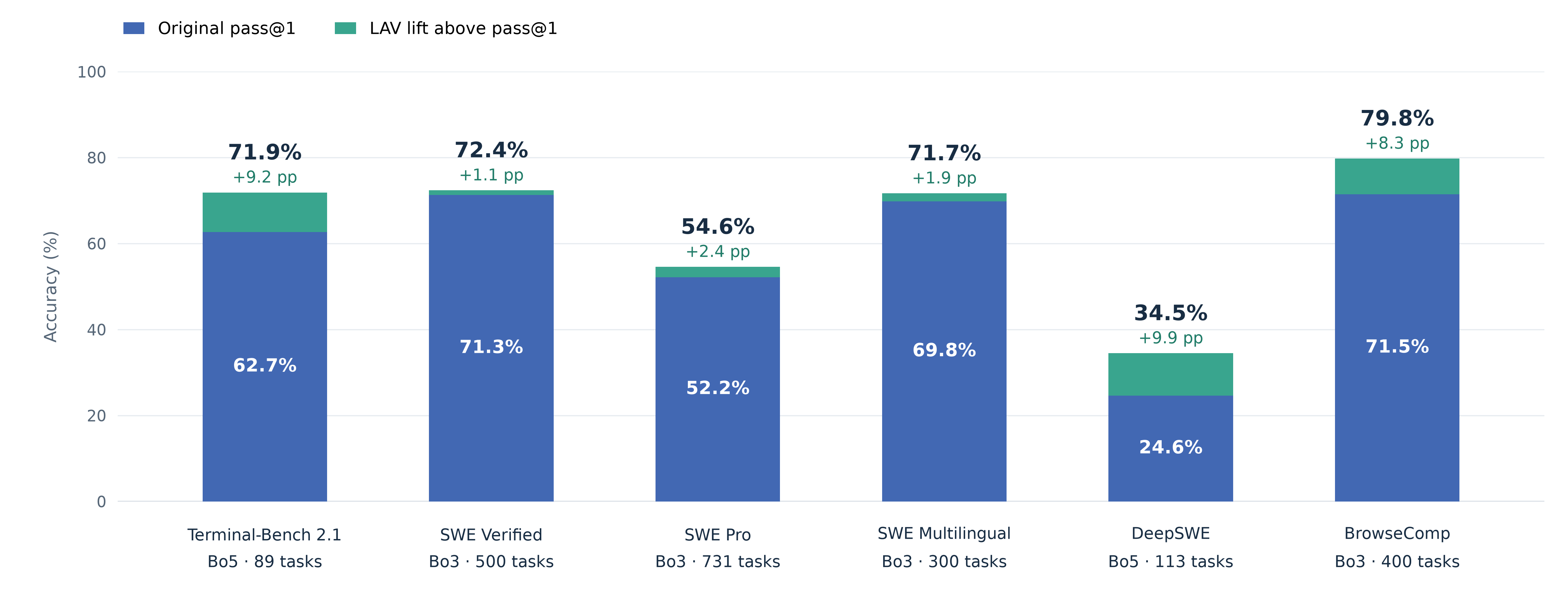}
    \caption{
        Test-time scaling with the final post-trained checkpoint.
        Blue bars show empirical pass@1, averaged over the candidate
        rollouts; green segments show the improvement from LAV
        selection, with total bar height indicating selected accuracy.
        Candidate-pool sizes are shown below each benchmark.
        BrowseComp is evaluated on a 400-problem subset used for
        internal evaluation tracking.
    }
    \label{fig:test-time-scaling}
\end{figure}

As shown in Fig.~\ref{fig:test-time-scaling}, selection improves
observed accuracy across all six benchmarks. Terminal-Bench accuracy
increases from 62.7\% to 71.9\%, DeepSWE from 24.6\% to 34.5\%, and
BrowseComp from 71.5\% to 79.8\%---absolute gains of 9.2, 9.9, and
8.3 percentage points, respectively. SWE-bench Verified, Pro, and
Multilingual improve by 1.1, 2.4, and 1.9 percentage points.
On the BrowseComp subset, selection recovers approximately 70\%
of the gap between pass@1 and oracle best-of-three performance.
These results demonstrate an inference-time path to higher task
accuracy that complements post-training without changing model
parameters.

\subsection{Benchmark Decontamination}
\label{subsec:decontamination}

Post-training data drawn from many synthetic and curated sources can absorb benchmark
content, not only as verbatim copies but also as paraphrases, translations,
augmentations, and other derived samples that string matching
misses~\citep{yang2023rethinking}. Such contamination inflates benchmark scores and,
because the same benchmarks guide training monitoring and checkpoint selection, it can
also result in a worse final model. We therefore performed checks of post-training data
against 30 benchmarks (see the complete list in Appendix~\ref{app:decon-benchmarks})
and removed samples related to benchmark questions. This data-side detection
complements model-side detection with CoDeC~\citep{zawalski2026codec},
which probes a trained model for traces of a benchmark. The pipeline will be released as part of NeMo Curator.

\paragraph{Decontamination pipeline.}
The pipeline combines two simple similarity signals, which select candidate matches
with high recall, with an LLM judge, which verifies each candidate precisely. Every
training sample passes through four stages:
\begin{enumerate}
  \item \textbf{Normalization.} The sample is reduced to its problem statement, the part
        into which a benchmark question could leak. Text templates that recur across
        its dataset are stripped to avoid spurious matches.
  \item \textbf{Candidate retrieval.} The sample is compared with every benchmark question
        using two signals. \emph{Verbatim overlap} measures how many $n$-grams of the question's text
        appear in the sample. It detects verbatim copies, even
        inside longer texts, but misses any rephrasing. \emph{Semantic overlap} measures
        the embedding similarity between parts of the two texts. It effectively captures paraphrases,
        but can be biased by style similarity. Both signals are used
        to shortlist the most likely contamination candidates.
  \item \textbf{Verification.} Each shortlisted pair of a sample and a benchmark question
        is verified by an LLM judge. The judge classifies the
        relationship either as contamination (direct copy, paraphrase, answer
        leakage, or method leakage) or as benign (style overlap or no contamination), and decides
        its severity. Because the shortlist is small, even a strong judge can be used in this step.
  \item \textbf{Removal.} A sample is removed if the judge finds it severely contaminated
        with respect to any benchmark question.
\end{enumerate}

\paragraph{Pipeline quality and results.}
We evaluated the quality of the pipeline on synthetic contamination. For this purpose, we generated
over 10,000 variants of Terminal-Bench~2.1 and GPQA Diamond questions,
produced by 13 augmentations of varying strength. Those augmentations ranged from character noise and
cropping to LLM-generated paraphrases, translations, and abstractions. The generated variants were
mixed with a similar number of same-domain distractors.
The judge correctly identified 87.5\% and 79.1\% of the contaminated
samples, respectively, while flagging only a single distractor as contaminated.
More details on the evaluation setup can be found in Appendix~\ref{app:decon-eval}.

From the post-training data, the pipeline removed 20,671 samples (around 0.09\%; 14,793 SFT and 5,878 RL samples),
matching 1,808 benchmark questions (6.0\%) in 22 of the 30 benchmarks. Of the
candidate pairs the judge worked on, around 3\% were flagged as severe contamination. See Appendix~\ref{app:decon-cases} for examples.

\clearpage

\FloatBarrier
\section{Quantization}
\label{sec:quantization}

%
\providecommand{\TODO}[1]{\textcolor{red}{[\textbf{TODO:} #1]}}
\providecommand{\owner}[1]{\textit{\small Owner: #1}\par\vspace{2pt}}


We apply post-training quantization (PTQ) and quantization-aware distillation (QAD) with \mbox{ModelOpt}\footnote{\url{https://github.com/NVIDIA/Model-Optimizer}} on Lightning 3.5 and Super 3.5 VL to create a new efficiency-accuracy frontier relative to the unquantized, post-trained BF16 checkpoint.

Recipe development starts by defining bookends: conservative and aggressive. Some models might also define a \textit{very aggressive} variant. These bookends are used to project runtime efficiency gains at various SLAs in the throughput-interactivity Pareto. Unless otherwise noted, we assume 51200, 2048, and 1024 as the pre-cached ISL, new ISL, and OSL, respectively. We measure speedups between a quantized candidate and BF16 using two SLAs:

\begin{itemize}
    \item Throughput (tok/s/gpu) at an SLA of 50 tok/s/usr
    \item Max throughput (e.g., usually highest concurrency)
\end{itemize}

The deployment use case also influences recipe selection. For example, Lightning~3.5 is
designed to excel in local, agentic use cases as well as datacenter deployments. We therefore
quantize Lightning~3.5 to minimize weight fetch and raise activation precision to maintain accuracy. Lightning~3.5 deploys to W4A16 regardless of GPU SKU, while
Super~3.5~VL's NVFP4 checkpoint deploys W4A4 on Blackwell, and W4A16 on non-Blackwell platforms.

Quantizing the KV and Mamba SSM caches reduces HBM footprint and improves decode throughput by minimizing data transfer. The KV cache grows linearly with sequence length, whereas the Mamba SSM state is constant in size. New to Super 3.5 VL is ablations on quantized NVFP4 KV cache. We show in \S\ref{sec:quant:kv} that quantization of KV cache to NVFP4 in combination with quantized SSM meets acceptance criteria.

Finally, we aim for the highest efficiency and lowest HBM footprint while satisfying the following accuracy criteria:

\begin{itemize}
    \item Median accuracy recovery of 99\% or higher, defined as the median of the ratios of NVFP4 to BF16 eval scores across several repeats.
    \item Verbosity increase of no more than 5\% relative to BF16.
\end{itemize}

Evaluation results for quantized Lightning 3.5 and Super 3.5 VL are shown in \S\ref{sec:quant:lightning-results} and \S\ref{sec:quant:results}, respectively.
\subsection{Recipe Design}
\label{sec:quant:recipes}

We explored three predefined precision recipes (Table~\ref{tab:quant:tiers}).
The released checkpoint uses the \emph{aggressive} recipe, with NVFP4 MoE
and shared-expert GEMMs and FP8 attention projections and Mamba linear
layers. AutoQuant was explored as an alternative sensitivity-
driven mixed-precision approach.

\begin{table}[t]
  \centering
  \footnotesize
  \setlength{\tabcolsep}{4.5pt}
  \caption{Per-module precision across predefined recipe tiers. The
  Aggressive column is the released recipe. NVFP4 denotes e2m1-based
  4-bit quantization; FP8 denotes e4m3; W4A16 denotes NVFP4 weights with
  16-bit activations; {+}SR denotes stochastic rounding.
  Reference columns denote precision policies, not complete benchmark
  serving configurations; the very-aggressive column was explored but
  not released. Whole-model BPE includes quantization scales and
  BF16-stored router weights: Aggressive is measured from the released
  checkpoint; the other columns are policy-based estimates.}
  \label{tab:quant:tiers}
  \begin{tabular}{@{}lccccc@{}}
    \toprule
     & \multicolumn{2}{c}{\emph{Reference}}
     & \multicolumn{3}{c}{\emph{Predefined recipes}} \\
    \cmidrule(lr){2-3}\cmidrule(lr){4-6}
    Layer / axis & BF16 & FP8 & Conservative & \textbf{Aggressive} & Very aggressive \\
    \midrule
    \multicolumn{6}{@{}l}{\emph{Vision path}} \\
    C-RADIO encoder     & BF16 & BF16 & BF16 & BF16 & BF16 \\
    VL connector        & BF16 & BF16 & BF16 & BF16 & BF16 \\
    \addlinespace[2pt]
    \multicolumn{6}{@{}l}{\emph{Latent MoE}} \\
    MoE + shared GEMM   & BF16 & FP8 & NVFP4\textsuperscript{a} & NVFP4 & NVFP4 \\
    Latent down/up proj & BF16 & FP8 & BF16 & BF16 & NVFP4 \\
    Router / gate (compute)\textsuperscript{b} & FP32 & FP32 & FP32 & FP32 & FP32 \\
    \addlinespace[2pt]
    \multicolumn{6}{@{}l}{\emph{Attention}} \\
    QKV / out proj      & BF16 & FP8 & BF16 & FP8 & NVFP4 \\
    KV cache            & BF16 & FP8 & FP8 & FP8 & FP4 \\
    BMM1 ($QK^{\top}$)  & BF16 & FP8 & FP8 & FP8 & FP4 \\
    BMM2 ($PV$)         & BF16 & FP8 & FP8 & FP8 & FP8 \\
    \addlinespace[2pt]
    \multicolumn{6}{@{}l}{\emph{Mamba}} \\
    Mamba linear        & BF16 & FP8 & BF16 & FP8 & NVFP4 \\
    Mamba SSM cache     & FP32 & FP32 & FP32 & FP16{+}SR & FP16{+}SR \\
    Mamba convolution   & BF16 & BF16 & BF16 & BF16 & BF16 \\
    \addlinespace[2pt]
    \multicolumn{6}{@{}l}{\emph{Embed / head}} \\
    Embedding           & BF16 & BF16 & BF16 & BF16 & BF16 \\
    \texttt{lm\_head}   & BF16 & FP8 & BF16 & FP8 & W4A16 \\
    \midrule
    Avg.\ BPE           & 16.00\textsuperscript{c} & 8.10\textsuperscript{c} & 5.15\textsuperscript{c} & \textbf{4.82} & 4.64\textsuperscript{c} \\
    \bottomrule
  \end{tabular}

  \vspace{2pt}
  \footnotesize
  \textsuperscript{a}~FP8 considered for the shared expert during exploration;
  the Conservative BPE estimate assumes NVFP4 for all shared experts.
  \textsuperscript{b}~FP32 denotes router computation, not checkpoint storage;
  BPE uses BF16-stored router weights in every column.
  \textsuperscript{c}~Estimated by applying each column's weight policy to
  corresponding backbone and MTP modules; MTP-specific fusion projection
  and normalization parameters remain BF16. The released MTP also retains
  BF16 attention weights; its storage is counted directly in the measured BPE.
  The released runtime uses an FP32 convolution cache and FP16+SR SSM cache.
\end{table}

We compute whole-model BPE as $8S/N$, where $S$ is stored tensor payload
in bytes, including quantization scales, and $N$ is the original parameter
count, including vision, MTP, and unquantized modules. For the released
checkpoint, $S=74.987$ GB and $N=124.453$ billion, giving 4.82 BPE;
runtime caches and activations are excluded.

PTQ establishes the initial accuracy of candidate recipes, and QAD
is used to further recover the quantization gap. The released checkpoint
uses the predefined \emph{aggressive} PTQ recipe followed by SFT-based QAD
at \textbf{200 iterations}. Its median accuracy recovery is
\textbf{99.62\%} across 36 benchmarks relative to the BF16-weight
reference (also served with FP8 KV cache). For each
benchmark, we first average completed runs separately for the released checkpoint
and BF16, compute their score ratio, and then take the median across
benchmarks. Across 36 benchmarks with token measurements, mean
per-benchmark verbosity (Ratio of totals) increases by \textbf{0.44\%}.

The released checkpoint was evaluated with an FP8 KV cache. Our KV cache ablations showed comparable accuracy with FP8 and NVFP4. It also includes the quantized, boosted MTP. As discussed, the checkpoint uses the predefined \emph{aggressive} recipe with QAD recovery.

\subsection{Post-Training Quantization}
\label{sec:quant:ptq}
\paragraph{Post-Training Quantization.}
We apply post-training quantization (PTQ) using NVIDIA Model Optimizer to quantize Nemotron 3.5 Super to NVFP4 for efficient inference on NVIDIA GPUs. The model is a 120B-parameter, 12B-active mixture-of-experts vision-language model comprising a C-RADIO vision encoder, a hybrid Mamba/attention language backbone with 512 routed experts per MoE layer, and a multi-token prediction (MTP) head. 

\paragraph{Calibration Data.}
Activation scales are learned from a calibration set of \textbf{image--text pairs}, rather than text alone. Images are passed through the vision encoder so that the language model observes the true input distribution, including the projected visual tokens. We use the SFT data employed during post-training as the calibration corpus and calibrate the model using 1,024 samples.

\paragraph{Calibration Parallelism.} 
We explored several parallelism strategies for efficient calibration:  Pipeline parallelism, Data Parallel (DP) × Expert Parallel (EP), and FSDP2. Pipeline parallelism left only one GPU active at a time, so we chose DP $\times$ EP. Each of the four ranks processes 256 different samples, covering all 1,024 calibration samples. The 512 routed experts are split evenly across the ranks, and the MTP experts are split the same way. We keep the vision encoder and projector, dense attention layers, and Mamba layers replicated, since the experts make up most of the model’s parameters.

\paragraph{Weight-Scale Selection.}
We compared three weight-scale selection schemes at a fixed NVFP4 quantization tier:

\begin{itemize}
    \item \textbf{Max}: The scale is determined from the observed absolute maximum.
    \item \textbf{MSE}: Each block scale is selected to minimize quantization error by sweeping over FP8 scale values.
    \item \textbf{MSE 4-over-6}: MSE-based scale selection restricted to a narrower search space.
\end{itemize}
From the evaluation results (Table~\ref{tab:calib:accuracy}) we observe that the MSE algorithm is superior in terms of accuracy recovery compared to the Max, and MSE 4-over-6. Hence we use the MSE calibration for the PTQ and AutoQuant recipes.

\begin{table}[t]
  \centering
  \small
  \caption{Accuracy for the aggressive tier calibrated using three different
  algorithms: MSE, Max, and MSE 4-over-6. Deltas are relative to the BF16
  baseline.}
  \label{tab:calib:accuracy}
  \resizebox{0.85\textwidth}{!}{%
  \begin{tabular}{lcccc}
    \toprule
    Benchmark & BF16 & MSE & Max & MSE 4-over-6 \\
    \midrule
    MMLU-ProX            & 76.92 & 73.92 (\textcolor{red}{$-3.00$})  & 72.77 (\textcolor{red}{$-4.15$})  & 74.77 (\textcolor{red}{$-2.15$})  \\
    GPQA-D w/o tools     & 89.52 & 90.03 (\textcolor{blue}{$+0.51$}) & 89.90 (\textcolor{blue}{$+0.38$}) & 89.52 ($\pm 0.00$)               \\
    APEX Shortlist       & 88.56 & 86.97 (\textcolor{red}{$-1.60$})  & 87.37 (\textcolor{red}{$-1.20$})  & 88.70 (\textcolor{blue}{$+0.13$}) \\
    SciCode Subtask      & 47.52 & 48.37 (\textcolor{blue}{$+0.86$}) & 46.81 (\textcolor{red}{$-0.70$})  & 46.79 (\textcolor{red}{$-0.72$})  \\
    SciCode Problem      & 15.77 & 16.54 (\textcolor{blue}{$+0.77$}) & 15.77 ($\pm 0.00$)               & 16.35 (\textcolor{blue}{$+0.58$}) \\
    AA-LCR               & 69.00 & 68.06 (\textcolor{red}{$-0.94$})  & 66.50 (\textcolor{red}{$-2.50$})  & 67.00 (\textcolor{red}{$-2.00$})  \\
    AA-Omni Acc.         & 26.21 & 25.25 (\textcolor{red}{$-0.96$})  & 25.52 (\textcolor{red}{$-0.69$})  & 25.25 (\textcolor{red}{$-0.96$})  \\
    AA-Omni Non-Halluci. & 70.09 & 71.19 (\textcolor{blue}{$+1.10$}) & 71.98 (\textcolor{blue}{$+1.89$}) & 71.50 (\textcolor{blue}{$+1.41$}) \\
    HLE w/o tools        & 31.84 & 31.70 (\textcolor{red}{$-0.14$})  & 30.21 (\textcolor{red}{$-1.62$})  & 31.14 (\textcolor{red}{$-0.70$})  \\
    Tau3 Banking         & 19.46 & 20.10 (\textcolor{blue}{$+0.64$}) & 18.94 (\textcolor{red}{$-0.52$})  & 18.94 (\textcolor{red}{$-0.52$})  \\
    Baby Vision          & 29.81 & 29.47 (\textcolor{red}{$-0.34$})  & 29.81 ($\pm 0.00$)               & 28.69 (\textcolor{red}{$-1.12$})  \\
    V* w/o CI            & 85.69 & 86.21 (\textcolor{blue}{$+0.52$}) & 86.91 (\textcolor{blue}{$+1.22$}) & 85.51 (\textcolor{red}{$-0.17$})  \\
    \midrule
    Score Recovery       & --    & \textbf{100.06\%} & \textbf{98.58\%} & \textbf{98.15\%} \\
    \bottomrule
  \end{tabular}%
  }
\end{table}
\subsection{Sensitivity-Driven Mixed Precision}
\label{sec:quant:autoquant}

We use ModelOpt AutoQuantize (AutoQuant) to replace hand-authored layer rules with a sensitivity-constrained mixed-precision search. For each quantizable module $i$ and candidate format $q\in\{\mathrm{NVFP4},\mathrm{FP8},\mathrm{BF16}\}$, AutoQuant estimates a sensitivity $S_i(q)$ from the quantization-induced output error weighted by the squared task-loss gradient. It then solves
\begin{equation}
  \min_{\{q_i\}} \sum_i S_i(q_i)
  \quad \text{s.t.} \quad
  \sum_i n_i b(q_i) \leq \bar{b}\sum_i n_i,
\end{equation}
where $n_i$ is the number of weights, $b(q_i)$ includes the modeled storage overhead of format $q_i$, and $\bar{b}$ is the target effective bitwidth. Operators that must share a deployment format---such as Q/K/V projections and the routed experts within an MoE layer---form a single search decision. The vision encoder, cache quantizers, and fixed MTP recipe are held outside this search.

The sensitivity profile is highly nonuniform: the \texttt{lm\_head} accounts for nearly half of the NVFP4 sensitivity, attention and Mamba projections are the next most sensitive families, and MoE experts are comparatively robust. The candidate assignments follow this hierarchy: they concentrate NVFP4 on the parameter-dominant expert weights, reserve FP8 or BF16 for sensitive attention and Mamba projections, and retain higher precision for the \texttt{lm\_head}.

\subsection{Quantization-Aware Distillation (QAD)}
\label{sec:quant:qad}

\paragraph{QAD Student (PTQ Initialization)} We initialize each QAD student from a post-training quantized (PTQ) checkpoint.
We derive the quantization configuration from an existing \textit{quantization-reference checkpoint}, which comes from ModelOpt AutoQuant or fixed-recipe PTQ (Sections~\ref{sec:quant:ptq} and \ref{sec:quant:autoquant}). 
The only intentional change is KV cache format; quantization-reference checkpoints use FP8 KV cache, whereas our PTQ student uses NVFP4 KV quantization. 
As such, all results in this subsection refer to deployment with NVFP4 KV cache.
Calibration uses 512 packed multimodal samples, with variable-length records packed up to 524,288 tokens and the packing buffer contains 5,000 source samples.
The PTQ workflow for QAD student initialization used eight nodes with four NVIDIA GB200 GPUs per node (i.e., 32 GPUs total). The complete workflow
took approximately 50 minutes per candidate (i.e., about 27 GB200 GPU-hours per PTQ candidate).


\paragraph{Data Blend}
We study two data blend settings: the SFT data blend employed during post-training (100\% SFT), and a mixture of 75\% SFT data with 25\% teacher-generated RL rollouts. 
For the RL rollouts we sample 16 trajectories per source record with vLLM.
Generation 
executes the original RL task workflows and retains environment and tool interactions.
We mask out prompt, conversation-history, tool-output, and prefilled response-prefix tokens. 
The rollout mixture provided only modest benefits in our ablations.

\paragraph{QAD Experimental Setup}
Each QAD experiment starts from its calibrated PTQ student checkpoint  with a fresh optimizer. 
We optimize only the weights of language-model GEMMs with enabled quantizers. All other parameters remain frozen, including token embeddings, unquantized language-model parameters, normalization parameters, the vision encoder and projector, and the MTP module. Router-bias updates are explicitly disabled. The quantization configuration and calibrated quantizer statistics are restored from the PTQ checkpoint rather than recalibrated during QAD.
We use KL divergence as our objective. We disable the main hard-target cross-entropy loss, MTP cross-entropy loss, and MoE load-balancing auxiliary loss and don't use intermediate-feature distillation loss. Optimization uses distributed AdamW with $\beta_1=0.9$, $\beta_2=0.95$, $\epsilon=10^{-8}$, weight decay $0.1$, and gradient clipping at a global norm of $1.0$. The learning rate increases linearly from zero to $5\times10^{-6}$ over 9,100 original source records and remains constant thereafter.
We run QAD on 128 nodes with four NVIDIA GB200 GPUs per node (512 GPUs total). QAD for 100 iterations took about 2 hours (i.e., 1024 GB200 GPU-hours) for each of our quantization candidates.

\paragraph{QAD Results}
We evaluated three quantization recipes (for which we generated PTQ and QAD checkpoints as described above): a) a \textit{4.8 BPE NVFP4 recipe} (based on the quantization configuration selected by the corresponding AutoQuant run), b) a \textit{4.9 BPE NVFP4 recipe} (similarly, based on a 4.8 BPE target AutoQuant), c) a \textit{fixed 4.82 BPE NVFP4 recipe}.

Below, we show the results (accuracy and verbosity) for the \textit{4.8 BPE recipe}; specifically, the PTQ checkpoint and the QAD checkpoints for \textit{200}, \textit{400}, \textit{600} iterations, for both the \textit{100\% SFT} blend and the \textit{75/25\% SFT/RL} blend cases. This initial evaluation sweep was based on an earlier (i.e., before final) version of the BF16 counterpart. Tables~\ref{tab:row90-4p8-accuracy} and~\ref{tab:row90-4p8-verbosity} compare the quantized models against that BF16 baseline. PTQ preserves a median 97.68\% of the BF16 accuracy while producing 0.22\% fewer completion tokens on average. QAD consistently closes much of the remaining accuracy gap, with the evaluated configurations achieving between 98.36\% and 99.18\% median score recovery. We obtain the best overall accuracy--verbosity tradeoff after 200 QAD iterations using 100\% SFT data (99.18\% median accuracy recovery - 1.50 percentage points above PTQ - with only a 1.11\% increase in aggregate
completion-token count relative to BF16). Additional QAD iterations were not uniformly beneficial. For the 100\% SFT experiments, the 100 iterations checkpoint provides the strongest overall accuracy--verbosity tradeoff. The 75/25\% SFT/RL experiments perform best at the intermediate iterations checkpoint. Overall, our results favor shorter QAD training with 100\% SFT data and show that optimal training duration depends on the selected data mixture.

We adjusted the QAD experiments based on the above observations for the final post-trained BF16 checkpoint. In particular,
we reduced checkpointing interval to 100 iterations, used the
100\% SFT blend only, and evaluated QAD checkpoints at 200, 300, and 400 iterations. We repeated this process for each of the three recipes considered previously: 4.8, 4.82, and 4.9 BPE.
Tables~\ref{tab:row105-qad-accuracy} and~\ref{tab:row105-qad-verbosity} report the accuracy and verbosity results used for final candidate downselection (wherein selected candidate checkpoints undergo a more thorough evaluation with a broader set of benchmarks). The effect of additional QAD iterations is recipe-dependent. For the 4.8-BPE recipe, moving from 200 to 300 iterations improves both accuracy recovery and verbosity, reducing the aggregate token increase from 2.30\% to 0.75\%. Similarly, for the 4.9-BPE recipe, moving from 200 to 300 iterations provides substantially better recovery and reduces verbosity change from 1.79\% to parity with BF16. The 4.82-BPE recipe behaves differently: its 200-iteration checkpoint provides the strongest accuracy recovery (almost matches median BF16 accuracy and verbosity), whereas the 300-iteration one is less accurate and slightly less verbose. Overall, the 4.82-BPE recipe at 200 iterations provides the strongest accuracy--verbosity tradeoff and constitutes the strongest candidate. We also selected the best candidate from each of the other two recipes for a final stage full benchmark evaluation sweep.

Note that although the 400-iteration QAD checkpoints were also evaluated, they are omitted from Tables~\ref{tab:row105-qad-accuracy}
and~\ref{tab:row105-qad-verbosity}, and do not alter the candidate downselection. Our results in fact indicate that the principal QAD benefits largely saturate by 300 iterations. 

\begin{table*}[t]
\centering
\caption{Accuracy results for the BF16 baseline, the 4.8-BPE MSE PTQ
checkpoint, and QAD checkpoints trained with either 100\% SFT or a
75/25 SFT/RL mixture. Higher is better.}
\label{tab:row90-4p8-accuracy}
\begingroup
\scriptsize
\renewcommand{\arraystretch}{1.12}
\begin{tabular*}{\textwidth}{@{\extracolsep{\fill}}lrrrrrrrr@{}}
\toprule
& \textbf{BF16}
& \textbf{PTQ}
& \multicolumn{6}{c}{\textbf{QAD}} \\
\cmidrule(lr){4-9}
& & &
\multicolumn{2}{c}{\textbf{Iter.\ 200}} &
\multicolumn{2}{c}{\textbf{Iter.\ 400}} &
\multicolumn{2}{c}{\textbf{Iter.\ 600}} \\
\cmidrule(lr){4-5}
\cmidrule(lr){6-7}
\cmidrule(lr){8-9}
\textbf{Benchmark / checkpoint}
& & &
\textbf{100/0} & \textbf{75/25} &
\textbf{100/0} & \textbf{75/25} &
\textbf{100/0} & \textbf{75/25} \\
\midrule
MMLU-ProX
  & 74.49 & 71.39 & 70.81 & 71.50 & 70.49 & 71.04 & 70.79 & 71.65 \\
GPQA-D w/o tools
  & 90.28 & 90.09 & 89.58 & 90.09 & 91.10 & 89.77 & 90.21 & 89.39 \\
APEX Shortlist
  & 89.36 & 86.04 & 89.49 & 88.30 & 88.30 & 89.10 & 88.03 & 88.96 \\
SciCode subtask
  & 48.25 & 48.43 & 48.75 & 47.84 & 48.31 & 47.30 & 48.58 & 47.38 \\
SciCode problem
  & 17.88 & 16.73 & 17.31 & 16.54 & 16.54 & 16.15 & 16.92 & 16.35 \\
AA-LCR
  & 70.00 & 68.38 & 69.00 & 67.88 & 68.81 & 68.38 & 68.00 & 68.50 \\
AA-Omni accuracy
  & 26.40 & 25.95 & 26.38 & 25.97 & 25.97 & 26.63 & 25.90 & 26.37 \\
AA-Omni non-halluc.
  & 67.93 & 69.14 & 67.51 & 66.89 & 67.74 & 67.33 & 68.20 & 67.18 \\
HLE w/o tools
  & 32.95 & 31.65 & 31.42 & 31.70 & 31.93 & 32.02 & 32.67 & 31.37 \\
Tau3 Banking
  & 21.65 & 20.10 & 19.72 & 19.46 & 20.49 & 22.04 & 20.23 & 20.23 \\
BabyVision
  & 31.01 & 28.78 & 30.67 & 28.69 & 28.61 & 29.47 & 30.93 & 30.58 \\
CharXiv-RQ
  & 81.03 & 81.10 & 82.07 & 81.33 & 81.93 & 80.73 & 81.73 & 81.10 \\
V$^\star$ w/o CI
  & 85.51 & 86.39 & 84.82 & 84.47 & 85.51 & 86.74 & 85.86 & 87.09 \\
\midrule
\textbf{Score recovery median}
  & \textbf{100.00\%}
  & \textbf{97.68\%}
  & \textbf{99.18\%}
  & \textbf{98.36\%}
  & \textbf{98.36\%}
  & \textbf{99.12\%}
  & \textbf{99.16\%}
  & \textbf{98.61\%} \\
\bottomrule
\end{tabular*}
\endgroup
\end{table*}

\begin{table*}[t]
\centering
\caption{Average completion-token counts for the BF16 baseline, the
4.8-BPE MSE PTQ checkpoint, and the QAD checkpoints. Lower values
indicate less verbose generations. The final row reports the change in total completion tokens (summed over all benchmarks) relative to the BF16 baseline.}
\label{tab:row90-4p8-verbosity}
\begingroup
\scriptsize
\renewcommand{\arraystretch}{1.12}
\begin{tabular*}{\textwidth}{@{\extracolsep{\fill}}lrrrrrrrr@{}}
\toprule
& \textbf{BF16}
& \textbf{PTQ}
& \multicolumn{6}{c}{\textbf{QAD}} \\
\cmidrule(lr){4-9}
& & &
\multicolumn{2}{c}{\textbf{Iter.\ 200}} &
\multicolumn{2}{c}{\textbf{Iter.\ 400}} &
\multicolumn{2}{c}{\textbf{Iter.\ 600}} \\
\cmidrule(lr){4-5}
\cmidrule(lr){6-7}
\cmidrule(lr){8-9}
\textbf{Benchmark / checkpoint}
& & &
\textbf{100/0} & \textbf{75/25} &
\textbf{100/0} & \textbf{75/25} &
\textbf{100/0} & \textbf{75/25} \\
\midrule
MMLU-ProX
  & 3,381 & 3,395 & 3,286 & 3,480
  & 3,389 & 3,477 & 3,411 & 3,492 \\
GPQA-D w/o tools
  & 11,991 & 12,157 & 12,548 & 13,262
  & 12,975 & 12,972 & 12,477 & 12,350 \\
APEX Shortlist
  & 61,817 & 64,980 & 64,208 & 65,010
  & 64,711 & 63,650 & 66,093 & 64,762 \\
SciCode
  & 6,594 & 6,504 & 6,513 & 6,872
  & 7,040 & 6,959 & 6,912 & 6,001 \\
AA-LCR
  & 4,698 & 4,206 & 4,362 & 5,100
  & 4,556 & 4,200 & 4,612 & 4,258 \\
AA-Omni
  & 5,825 & 4,812 & 5,533 & 6,494
  & 6,122 & 5,705 & 6,221 & 6,633 \\
HLE w/o tools
  & 37,198 & 36,863 & 37,301 & 40,480
  & 39,436 & 38,240 & 38,486 & 38,300 \\
Tau3 Banking
  & 13,712 & 13,583 & 13,798 & 13,851
  & 15,488 & 13,390 & 14,526 & 14,042 \\
BabyVision
  & 11,456 & 10,580 & 11,905 & 11,981
  & 11,114 & 11,561 & 11,619 & 11,551 \\
CharXiv-RQ
  & 3,496 & 2,930 & 2,726 & 3,534
  & 2,681 & 3,773 & 3,238 & 3,728 \\
V$^\star$ w/o CI
  & 1,267 & 1,071 & 1,048 & 1,446
  & 1,093 & 1,374 & 1,373 & 1,048 \\
\midrule
\textbf{Verbosity change}
& \textbf{0.00\%}
& \textbf{-0.22\%}
& \textbf{+1.11\%} & \textbf{+6.24\%}
& \textbf{+4.44\%} & \textbf{+2.39\%}
& \textbf{+4.67\%} & \textbf{+2.93\%} \\

\bottomrule
\end{tabular*}
\endgroup
\end{table*}

\begin{table*}[t]
\centering
\caption{Accuracy of BF16 baseline and the PTQ and QAD
checkpoints obtained from the 4.8, 4.82, and 4.9 BPE recipes. Q200 and
Q300 denote QAD checkpoints after 200 and 300 iterations, respectively.
Higher is better.}
\label{tab:row105-qad-accuracy}
\begingroup
\footnotesize
\setlength{\tabcolsep}{1.8pt}
\renewcommand{\arraystretch}{1.12}

\begin{tabular*}{\textwidth}
{@{}l@{\hspace{1pt}}r@{\extracolsep{\fill}}rrrrrrrrr@{}}
\toprule
& \textbf{BF16}
& \multicolumn{3}{c}{\textbf{4.8 BPE}}
& \multicolumn{3}{c}{\textbf{4.82 BPE}}
& \multicolumn{3}{c}{\textbf{4.9 BPE}} \\
\cmidrule(lr){3-5}
\cmidrule(lr){6-8}
\cmidrule(lr){9-11}
\textbf{Benchmark / metric}
& \textbf{Base}
& \textbf{PTQ} & \textbf{Q200} & \textbf{Q300}
& \textbf{PTQ} & \textbf{Q200} & \textbf{Q300}
& \textbf{PTQ} & \textbf{Q200} & \textbf{Q300} \\
\midrule
MMLU-ProX
& 76.92
& 73.82 & 74.03 & 73.75
& 73.66 & 74.49 & 75.07
& 75.11 & 73.26 & 74.28 \\

GPQA-D w/o tools
& 89.52
& 90.34 & 90.03 & 90.21
& 89.84 & 89.71 & 90.03
& 89.90 & 89.84 & 89.84 \\

APEX Shortlist
& 88.56
& 89.36 & 86.84 & 88.96
& 88.16 & 87.10 & 90.03
& 88.43 & 88.70 & 88.96 \\

SciCode subtask
& 47.52
& 48.22 & 46.88 & 47.31
& 47.33 & 49.08 & 48.58
& 48.17 & 46.69 & 47.44 \\

SciCode problem
& 15.77
& 16.92 & 14.42 & 16.15
& 15.38 & 16.73 & 16.92
& 16.54 & 15.00 & 15.00 \\

AA-LCR
& 69.00
& 67.50 & 68.44 & 67.69
& 67.63 & 69.38 & 68.06
& 68.19 & 69.19 & 68.81 \\

AA-Omni accuracy
& 26.21
& 26.08 & 25.73 & 25.63
& 25.65 & 25.53 & 26.22
& 25.68 & 25.83 & 26.27 \\

AA-Omni non-halluc.
& 70.09
& 71.57 & 70.95 & 70.34
& 71.01 & 71.81 & 69.88
& 71.24 & 70.89 & 70.77 \\

HLE w/o tools
& 31.84
& 30.54 & 31.79 & 31.84
& 31.51 & 30.91 & 30.77
& 31.00 & 31.14 & 30.68 \\

Tau3 Banking
& 19.46
& 19.46 & 18.43 & 17.78
& 19.07 & 18.56 & 17.40
& 20.10 & 18.94 & 18.30 \\

BabyVision
& 29.81
& 29.90 & 30.07 & 29.30
& 28.78 & 29.47 & 28.87
& 29.04 & 28.69 & 29.38 \\

CharXiv-RQ
& 80.93
& 80.73 & 80.23 & 80.40
& 80.27 & 81.03 & 80.83
& 80.73 & 80.53 & 80.40 \\

V$^\star$ w/o CI
& 85.69
& 84.99 & 86.56 & 86.04
& 85.69 & 86.91 & 84.29
& 84.99 & 85.34 & 86.74 \\

\midrule
\textbf{Score recov. median}
& \textbf{100.00\%}
& \textbf{100.00\%}
& \textbf{99.14\%}
& \textbf{99.57\%}
& \textbf{98.98\%}
& \textbf{100.12\%}
& \textbf{99.71\%}
& \textbf{99.75\%}
& \textbf{98.56\%}
& \textbf{99.73\%} \\
\bottomrule
\end{tabular*}

\endgroup
\end{table*}

\begin{table*}[t]
\centering
\caption{Average completion-token counts for the PTQ and QAD checkpoints obtained from the 4.8, 4.82, and 4.9 BPE recipes. Lower token counts indicate less verbose
generations. The final row reports the change in total completion tokens (summed over all benchmarks) relative to the BF16 baseline.}
\label{tab:row105-qad-verbosity}
\begingroup
\footnotesize
\setlength{\tabcolsep}{1.8pt}
\renewcommand{\arraystretch}{1.12}

\begin{tabular*}{\textwidth}
{@{}l@{\hspace{1pt}}r@{\extracolsep{\fill}}rrrrrrrrr@{}}
\toprule
& \textbf{BF16}
& \multicolumn{3}{c}{\textbf{4.8 BPE}}
& \multicolumn{3}{c}{\textbf{4.82 BPE}}
& \multicolumn{3}{c}{\textbf{4.9 BPE}} \\
\cmidrule(lr){3-5}
\cmidrule(lr){6-8}
\cmidrule(lr){9-11}
\textbf{Benchmark / metric}
& \textbf{Base}
& \textbf{PTQ} & \textbf{Q200} & \textbf{Q300}
& \textbf{PTQ} & \textbf{Q200} & \textbf{Q300}
& \textbf{PTQ} & \textbf{Q200} & \textbf{Q300} \\
\midrule
MMLU-ProX
& 3,487
& 3,498 & 3,516 & 3,508
& 3,522 & 3,441 & 3,520
& 3,509 & 3,510 & 3,537 \\

GPQA-D w/o tools
& 14,424
& 14,553 & 15,284 & 14,802
& 14,774 & 14,709 & 14,388
& 14,943 & 14,867 & 14,765 \\

APEX Shortlist
& 68,401
& 66,810 & 67,748 & 67,934
& 66,830 & 66,687 & 66,721
& 66,866 & 67,717 & 64,926 \\

SciCode
& 6,459
& 6,960 & 7,404 & 6,904
& 6,535 & 6,782 & 6,822
& 6,972 & 7,078 & 7,205 \\

AA-LCR
& 4,444
& 4,242 & 4,367 & 4,100
& 4,280 & 4,971 & 4,153
& 4,158 & 4,319 & 4,664 \\

AA-Omni
& 4,485
& 4,382 & 4,789 & 4,623
& 4,282 & 4,458 & 4,537
& 4,310 & 4,504 & 4,584 \\

HLE w/o tools
& 39,253
& 39,042 & 41,512 & 40,341
& 40,494 & 40,383 & 39,610
& 39,669 & 40,701 & 40,392 \\

Tau3 Banking
& 12,073
& 11,856 & 11,808 & 12,555
& 12,037 & 11,701 & 12,317
& 11,871 & 12,851 & 12,258 \\

BabyVision
& 11,493
& 10,743 & 11,790 & 11,508
& 11,115 & 11,357 & 10,734
& 11,348 & 11,883 & 11,796 \\

CharXiv-RQ
& 3,961
& 4,235 & 4,073 & 3,558
& 3,796 & 3,955 & 3,613
& 3,829 & 4,109 & 4,582 \\

V$^\star$ w/o CI
& 876
& 1,170 & 960 & 794
& 798 & 805 & 897
& 826 & 843 & 747 \\

\midrule
\textbf{Verbosity change}
& \textbf{0.00\%}
& \textbf{-1.10\%} & \textbf{+2.30\%} & \textbf{+0.75\%}
& \textbf{-0.53\%} & \textbf{-0.06\%} & \textbf{-1.21\%}
& \textbf{-0.62\%} & \textbf{+1.79\%} & \textbf{+0.06\%} \\
\bottomrule
\end{tabular*}

\endgroup
\end{table*}




\subsection{KV Cache Quantization Evaluations}
\label{sec:quant:kv}

As part of our studies, we evaluate both FP8 and NVFP4 quantization cases; in each case both the key and value
caches across attention layers are quantized in FP8 or NVFP4, respectively. NVFP4 stores 4-bit E2M1 values with one
FP8 E4M3 scale per 16-element block, for an effective storage cost of
4.5 bits per cached element, excluding tensor-level scale overhead.
Relative to FP8's 8 bits per element, this reduces the ideal KV cache payload by $1-4.5/8=43.75\%$. This is a storage calculation, not a measured reduction in runtime memory traffic or latency; KV cache BPE is separate from the whole-model checkpoint BPE in Section~\ref{sec:quant:recipes}.

\paragraph{Evaluation details.}
Our reference uses \textbf{BF16 weights with an FP8 KV cache}. The two configurations in Table~\ref{tab:quant:kv-comparison} use our
\emph{aggressive} NVFP4 recipe with QAD at 200 iterations, evaluated
with either (uniform) NVFP4 or FP8 KV cache. For each benchmark, we
average completed runs within each configuration and the
reference, then take the median of the 35 configuration-to-reference
score ratios. Median accuracy recovery is \textbf{99.36\%} for the
NVFP4 KV cache configuration and \textbf{99.52\%} for the released FP8 KV cache configuration (on the same 35 benchmarks).

The NVFP4 KV cache experiments use the original QAD checkpoint, i.e.,  with fixed backbone KV scale values targeting NVFP4 KV cache deployment (1/6); the FP8 KV cache experiments use the released quantized checkpoint that contains no KV scales and targets FP8 KV cache deployment (vLLM defaults to KV scales of 1.0 for FP8 KV cache type). 
Both use an FP32 Mamba convolution cache and FP16 SSM cache with stochastic rounding.
The reported median accuracy recovery against the BF16-weight reference reflects the full
weight-and-cache configuration, not KV quantization alone. The results
show broadly comparable aggregate accuracy in the specific quantization
recipe used. The released quantized NVFP4 checkpoint defaults to FP8 KV cache for runtime-efficiency and software-readiness considerations despite NVFP4 KV's storage advantage. Per-layer mixed FP8-key/NVFP4-value (or vice-versa) configurations are not evaluated in this comparison.

\begin{table}[t]
  \centering
  \scriptsize
  \setlength{\tabcolsep}{5pt}
  \caption{Uniform NVFP4 KV cache versus FP8 KV cache configurations. Both use aggressive NVFP4 recipe with QAD at 200 iterations.
  Median accuracy recovery is relative to BF16 checkpoint with FP8 KV cache.}
  \label{tab:quant:kv-comparison}
  \begin{tabular}{@{}lrr@{}}
    \toprule
    Metric / benchmark & NVFP4 KV cache & FP8 KV cache\\
    \midrule
    KV cache payload BPE & 4.5 & 8.0 \\
    Median accuracy recovery (\%) & 99.36 & \textbf{99.52} \\
    \midrule
    MMLU-ProX & 74.47 & 73.76 \\
    Baby Vision & 28.61 & 29.51 \\
    CharXiv (RQ) & 80.30 & 79.88 \\
    MathVision & 77.84 & 77.79 \\
    MMLongBench-Doc & 62.22 & 62.82 \\
    OCR Reasoning & 65.50 & 65.09 \\
    OCRBenchV2 EN & 67.78 & 68.34 \\
    OCRBenchV2 ZH & 59.26 & 59.03 \\
    ScreenSpot Pro & 41.49 & 42.25 \\
    Video-MME-v2 (w/o sub.) & 32.47 & 32.54 \\
    V* (Without CI) & 85.51 & 85.95 \\
    GPQA Diamond w/o tools & 89.49 & 89.90 \\
    APEX Shortlist w/o tools & 87.30 & 89.36 \\
    GDPVal & 55.62 & 55.74 \\
    SciCode (subtask) & 46.91 & 46.61 \\
    SciCode (problem) & 15.19 & 14.13 \\
    AA-LCR & 68.16 & 67.69 \\
    GraphWalks ($\leq$1M) & 89.54 & 90.34 \\
    AA-Omniscience Accuracy & 25.54 & 25.86 \\
    AA-Omni Non-Hallucination & 72.32 & 72.33 \\
    IFBench & 70.72 & 71.50 \\
    HLE w/o tools & 30.63 & 31.44 \\
    HLE Vision & 27.68 & 29.00 \\
    SWE-Bench Verified & 70.53 & 71.07 \\
    SWE-Bench Multilingual & 66.33 & 67.00 \\
    SWE-Bench Pro & 50.93 & 51.30 \\
    Tau2 Telecom & 69.35 & 73.19 \\
    Tau2 Retail & 68.97 & 69.30 \\
    Tau2-AA (Airline) & 75.75 & 75.75 \\
    Tau3 Banking & 18.88 & 17.78 \\
    WMT24++ ($\mathrm{en}\to\mathrm{xx}$) & 52.28 & 51.94 \\
    WMT24++ COMET & 84.67 & 84.76 \\
    Terminal-Bench v2.1 & 60.67 & 62.78 \\
    LCB v6 & 87.37 & 87.41 \\
    NVIDIA Internal Chat Benchmark & 51.80 & 51.93 \\
    \bottomrule
  \end{tabular}

\end{table}


\subsection{Attention and Projection Precision}
\label{sec:quant:attn}

To understand the impact of quantization on model behavior, we ablated three components:
global attention's projection, mamba's projection, and the unembedding. We presumed W4A4
routed/shared experts and FP8 KV cache as the default configuration to test each projection's
sensitivity to quantization. This reveals a few interesting findings:

\begin{table*}[t]
  \centering
  \caption{Long-context capability: distinction (C, content), disambiguation (P, position), aggregation (S, summary); all quantized models serve an FP8 KV cache with W4A4 experts, and differ only in q/k/v projection precision. $t=\mathrm{mean}(\Delta)/\mathrm{SE}(\Delta)$, where $\Delta$ is the per-task
  difference of task means and the standard error is taken over \emph{tasks}, not
  rollouts --- so the four rows are directly comparable.}
  \label{tab:attn:lc}
  \begin{tabular}{lcccc}
    \toprule
    & & \multicolumn{3}{c}{$\Delta$ accuracy (pp) against the BF16 reference} \\
    \cmidrule(lr){3-5}
    Benchmark & Rollouts & q/k/v BF16 & q/k/v MXFP8 & q/k/v FP8 \\
    \midrule
    GraphWalks ($\leq$1M) \textsuperscript{C,S} & $950 \times 1$  & $-0.52$ (t=-0.70) & $+0.15$ (t=+0.21) & $-0.83$ (t=-1.03) \\
    MMLongBench-Doc \textsuperscript{C,S} & $1091 \times 3$ & $-0.76$ (t=-1.42) & $-0.90$ (t=-1.81) & $-0.34$ (t=-0.61) \\
    AA-LCR \textsuperscript{S}            & $100 \times 16$ & $+1.12$ (t=+0.89) & $-0.12$ (t=-0.10) & $+1.56$ (t=+1.19) \\
    \textbf{MRCR-128k \textsuperscript{P}} & $\mathbf{1470 \times 3}$ & $\mathbf{-2.81}$ (t=-3.90) & $\mathbf{-2.65}$ (t=-3.61) & $\mathbf{-3.63}$ (t=-4.97) \\
    \bottomrule
  \end{tabular}
\end{table*}

\begin{itemize}
  \item The MoE expert weights carry a primary impact on accuracy, causing a drift of token distribution in FP4-quantized latent space. This becomes critical for runaway loop generation in reasoning failures, if a decision boundary of abstention is fragile enough to be affected by quantization. No attention projection precision recovers the verbosity inflation. Empirically, promoting MoE expert precision and fine-tuning on RL rollouts can mitigate this effect (Appendix~\ref{app:quant:ablation:fp4experts}).
  \item Because long-context retrieval is dominated by KV cache precision, attention projection has minimal impact on accuracy in most cases. However, Table~\ref{tab:attn:lc} shows that \emph{attention score dilution}, sharing a single representable level of the quantized attention score among many keys, is more pronounced with disambiguation (e.g., MRCR-128k) in 32k+ context length. q/k/v projections do not require BF16, but per-tensor FP8's dynamic range constrains attention logit granularity. MXFP8's per-block scaling helps mitigate the quality degradation. Attention's output projection is structurally exposed but tolerant to FP8 quantization (Appendix~\ref{app:quant:ablation:lc}).
  \item Mamba output projection also has a minimal impact on accuracy, but quantizing its weights to FP4 inflates generation length. Its input activations are separately exposed: the SSM gate emits near-zero yet never exactly zero values, which flush to zero under FP4 activation quantization. Rotation or higher precision can mitigate this effect in activation, but does not address the verbosity inflation without weight realignment (Appendix~\ref{app:quant:ablation:mamba}).
  \item The \texttt{lm\_head} weights of unembedding do not change accuracy even with FP4 quantization. However, FP4 amplifies the perturbation of the stop-versus-continue margin. To maintain token generation verbosity, the \texttt{lm\_head} should stay at FP8 weights or above (Appendix~\ref{app:quant:ablation:lmhead}).
\end{itemize}

\providecommand{\yy}[1]{\textcolor{red}{#1}}   

\subsection{MTP Quantization}
\label{sec:quant:mtp}
To reduce drafting latency, we quantize the MTP module of \ourmodel{} NVFP4 checkpoint. This applies only to \ourmodel{}; the MTP module of Nemotron~3.5 Lightning is not quantized. The MTP module contains an attention layer and an MoE layer. It does not carry its own output projection: the released checkpoint stores a single \texttt{lm\_head}, shared between target verification and MTP drafting. Profiling shows that the MoE experts and the \texttt{lm\_head} account for most of the drafting latency. We therefore quantize the MTP experts using NVFP4 W4A4. The MTP attention weights remain BF16, with only the K/V BMM quantizers at FP8.

Isolating the effect of MTP quantization requires holding the backbone fixed. Against the NVFP4 backbone with a BF16 MTP ($AL=4.612$ at draft length 7), quantizing the MTP experts reduces AL to $AL=4.592$
(Table~\ref{tab:mtp-quantization}).

To assess the possible speedup from MTP and \texttt{lm\_head} quantization, we simulated drafting and verification latencies at draft length $5$, on a single GB200 GPU, with a quantized backbone. Our simulated latency projections show that when the MTP and \texttt{lm\_head} are both unquantized, drafting amounts for $23.5\%$ of the total latency, and the drafting latency splits to $49.6\%$ MTP latency and $51.4\%$ \texttt{lm\_head} latency. Therefore, using quantization to reduce the latency of both components is beneficial to reduce the total drafting time, if the drop in acceptance length is not too big. 

According to our projections, quantizing the MTP layer is expected to reduce $47\%$ of its latency, and quantizing the \texttt{lm\_head} to FP8 and W4A16 precisions is expected to reduce its latency by $49\%$ and $71\%$, respectively. Quantizing the target model's \texttt{lm\_head} was initially considered higher risk because it directly affects the target logits and, consequently, model quality. As a fallback, we experimented with various combinations using separate target and MTP \texttt{lm\_head} precisions, and implemented support in \emph{vLLM 0.30.0} for serving the model in this mixed configuration. This allows the target \texttt{lm\_head} to remain in BF16 while using a lower-precision \texttt{lm\_head} for drafting. The AL results and expected speedups over the using MTP and \texttt{lm\_head} are described in Table~\ref{tab:mtp-quantization}.

FP8 quantization of the target \texttt{lm\_head} was ultimately selected, thus a single FP8 \texttt{lm\_head} serves both target verification and MTP drafting, and no separate output head is required. The separate-precision support remains available for models where quantizing the target \texttt{lm\_head} degrades quality.

\paragraph{KV-cache coupling.} The MTP cannot use a different KV-cache precision from the backbone. MTP KV precision therefore follows the backbone. We empirically validated this does not degrade ALs, making the coupling a structural constraint rather than a quality trade-off.

\paragraph{Construction.} The released checkpoints were produced by quantizing the boosted MTP (Section~\ref{subsec:mtp_boosting}) separately and splicing it into the selected quantized backbone, rather than co-quantizing both in a single pass. The splice transfers the MTP weights and their scales while leaving the surrounding configuration unchanged, so source and destination must agree exactly on the MTP tensor set --- names, shapes
and dtypes. Co-quantizing the MTP together with the backbone removes this coupling and is the preferred direction for future releases. 


\begin{table}[t]
\caption{Acceptance length (AL) for different target-model and MTP-drafter
precisions, and projected speedups over the baseline. Results are measured on the SPEED-Bench \textit{qualitative}
split at temperature 1. Speedup is projected relative to the NVFP4 backbone using unquantized MTP and \texttt{lm\_head}. The shaded row denotes the selected configuration.}
\label{tab:mtp-quantization}
\centering
\small
\setlength{\tabcolsep}{6pt}
\begin{tabular}{
    @{}
    c c
    c c
    S[table-format=1.3]
    S[table-format=1.3]
    c
    @{}
}
\toprule
\multicolumn{2}{c}{\textbf{Target model}} &
\multicolumn{2}{c}{\textbf{MTP drafter}} &
\multicolumn{2}{c}{\textbf{Acceptance length}} &
\shortstack{\textbf{Projected speedup}} \\
\cmidrule(lr){1-2}
\cmidrule(lr){3-4}
\cmidrule(lr){5-6}
\cmidrule(lr){7-7}

Backbone &
LM head &
Experts &
LM head &
{\textbf{DL=5}} &
{\textbf{DL=7}} &
\textbf{DL=5} \\
\midrule

BF16  & BF16 & BF16  & BF16  & 4.115 & 4.612 & - \\
NVFP4 & BF16 & BF16  & BF16  & 4.109 & 4.592 & 1 \\
NVFP4 & BF16 & NVFP4 & BF16  & 4.066 & 4.518 & 1.043 \\
NVFP4 & BF16 & NVFP4 & FP8   & 4.069 & 4.514 & 1.114 \\
NVFP4 & BF16 & NVFP4 & W4A16 & 4.031 & 4.469 & 1.136 \\
\rowcolor{gray!15}
NVFP4 & FP8  & NVFP4 & FP8   & 4.063 & 4.520 & 1.127 \\
NVFP4 & FP8  & NVFP4 & W4A16 & 4.025 & 4.471 & 1.150 \\

\bottomrule
\end{tabular}
\end{table}

\subsection{Parallel Drafter Quantization}
\label{sec:quant:drafter}

Following a similar procedure to MTP quantization, we applied weight-only PTQ to improve the drafting efficiency of a DSpark~\citep{cheng2026dspark} module for Nemotron Lightning 3.5 and LiLiCorr~\citep{rusanovsky2026lilicorrlightweightlikelihoodcorrelation} module for \ourmodel{} (see Section~\ref{sec:lilicorr}). Our experiments showed that in memory-bound scenarios, e.g. on edge devices with low concurrency, parallel drafters benefit greatly from weight quantization, as this reduces the memory overhead from loading drafter parameters. For \ourmodel{}, we observed that quantizing LiLiCorr's linear layers to NVFP4 while keeping the activations at BF16, resulted in a 11\% throughput improvement on DGX Spark without meaningfully sacrificing acceptance lengths, as shown in Table~\ref{tab:lilicorr_eval}. Further quantizing the LM head improved throughput by another 5\%.
We identified model-specific quantization settings that performed well both in terms of drafter accuracy and overall throughput on DGX Spark. Additional details are in Table~\ref{tab:w4a16-quantized-layer-types}.

\begin{table}[t]
\centering
\caption{Quantized layers specification in DSpark and LiLiCorr draft models.}
\label{tab:w4a16-quantized-layer-types}
\footnotesize
\setlength{\tabcolsep}{3pt}
\begin{tabular}{@{}lll@{}}
\hline
\textbf{Layer group} & \textbf{Quantization-config name} & \textbf{Layer type} \\
\hline
\multirow[c]{3}{*}{\textbf{General backbone}}
  & \texttt{mlp.*} & Gate/up/down projections \\
  & \texttt{fc} & Fully connected projection \\
  & \texttt{lm\_head} & Language-model output head \\
\hline
\textbf{DSpark-specific}
  & \texttt{markov\_head.markov\_w2} & Markov output projection \\
\hline
\noalign{\vskip 3pt}
\multirow[c]{5}{*}{\textbf{LiLiCorr-specific}}
  & \texttt{lilicorr.context\_proj}
  & \multirow[c]{3}{*}{Context/hidden-state/token projections} \\
  & \texttt{lilicorr.pass\_hidden\_proj} & \\
  & \texttt{lilicorr.token\_proj} & \\
  & \texttt{lilicorr.attn\_out\_proj} & Attention output projection \\
  & \texttt{lilicorr.mlp} & LiLiCorr MLP projections \\
\hline
\end{tabular}%
\end{table}

\subsection{Lightning 3.5 Evaluation and Results}
\label{sec:quant:lightning-results}

We evaluate Nemotron 3.5 Lightning using the NeMo Gym evaluation framework described for Super 3.5. Lightning-specific settings are available in the recipes folder and reproducibility tutorial\footnote{\url{https://github.com/NVIDIA-NeMo/Gym/blob/main/nemotron_recipes/lightning-3.5/reproducibility.md}}. 
For evaluation, we serve both the BF16 and NVFP4 checkpoints with FP8 KV cache and FP32 Mamba SSM cache. SSM cache is kept in FP32 as it was not a limiter on DGX Spark for low batch (< 4) use cases. In the W4A16 NVFP4 checkpoint, the routed experts, shared experts, and language-model output head use 4-bit NVFP4 weights with 16-bit activations. The Mamba input and output projections
use FP8 weights and activations, while attention projections remain in BF16.
We select the W4A16 NVFP4 configuration using the same PTQ/QAD procedure, and compare it with BF16 under matched evaluation and serving settings. 
Table~\ref{tab:quant:lightning-main} compares the benchmark scores\footnote{Evaluation results are subject to run-to-run variation due to stochastic sampling and, where applicable, judge variability. Scores may therefore differ from those reported on the model cards.} between BF16 and W4A16 NVFP4, reporting a \texttt{Median Accuracy Recovery} of 99.24\% and a \texttt{Ratio of Totals Verbosity} change of $-3.63\%$ across benchmarks.

\begin{table*}[t]
  \centering
  \footnotesize
  \setlength{\tabcolsep}{3pt}
  \caption{Lightning 3.5 BF16 versus W4A16 NVFP4 using scores
  from the saved evaluation view. Token columns retain the
  original average tokens per trajectory.}
  \label{tab:quant:lightning-main}
  \begin{tabular}{lrrrrrr}
    \toprule
    Benchmark
      & \shortstack{BF16 tokens\\per trajectory}
      & \shortstack{NVFP4 tokens\\per trajectory}
      & \shortstack{BF16\\acc.}
      & \shortstack{NVFP4\\acc.}
      & \shortstack{$\Delta$\\(pp)}
      & \shortstack{Recovery\\(\%)} \\
    \midrule
    \multicolumn{7}{l}{\textbf{Agentic}} \\
    \cmidrule(lr){1-7}
    SWE-Bench Verified
      & 24,569 & 23,414 & 52.20 & 52.20 & 0.00 & 100.00 \\
    SWE-Bench Multilingual
      & 31,071 & 33,924 & 37.13 & 37.00 & -0.13 & 99.65 \\
    Tau2 Telecom
      & 7,076 & 7,075 & 59.65 & 60.96 & +1.32 & 102.21 \\
    Tau3 Banking
      & 19,107 & 17,998 & 9.90 & 9.48 & -0.41 & 95.83 \\
    Terminal-Bench v2.1
      & 76,219 & 69,523 & 24.44 & 22.05 & -2.39 & 90.22 \\
    GDPval
      & 52,043 & 52,563 & 17.19 & 19.09 & +1.89 & 111.01 \\
    PinchBench
      & 7,206 & 7,115 & 83.86 & 84.78 & +0.92 & 101.09 \\
    BrowseComp
      & 142,589 & 138,151 & 39.50 & 39.50 & 0.00 & 100.00 \\
    \addlinespace[0.5em]
    \multicolumn{7}{l}{\textbf{Coding}} \\
    \cmidrule(lr){1-7}
    SciCode (subtask)
      & 17,190 & 17,428 & 32.74 & 31.53 & -1.21 & 96.30 \\
    \addlinespace[0.5em]
    \multicolumn{7}{l}{\textbf{Reasoning and Knowledge}} \\
    \cmidrule(lr){1-7}
    GPQA Diamond (no tools)
      & 15,552 & 14,394 & 76.45 & 75.38 & -1.07 & 98.60 \\
    HLE (no tools)
      & 30,352 & 27,270 & 10.80 & 10.70 & -0.09 & 99.14 \\
    AA-Omniscience Accuracy
      & 2,505 & 2,588 & 17.47 & 16.35 & -1.12 & 93.61 \\
    AA-Omni Non-Hallucination
      & 2,505 & 2,588 & 68.88 & 69.20 & +0.32 & 100.46 \\
    CritPt
      & 4,379 & 3,622 & 0.00 & 0.00 & 0.00 & -- \\
    \addlinespace[0.5em]
    \multicolumn{7}{l}{\textbf{Long Context}} \\
    \cmidrule(lr){1-7}
    AA-LCR
      & 7,679 & 7,388 & 52.38 & 50.00 & -2.38 & 95.47 \\
    RULER-v1 (1M)
      & 22 & 21 & 91.52 & 90.83 & -0.69 & 99.24 \\
    \addlinespace[0.5em]
    \multicolumn{7}{l}{\textbf{Instruction Following and Chat Preference}} \\
    \cmidrule(lr){1-7}
    IFBench (loose)
      & 13,069 & 11,422 & 72.00 & 74.04 & +2.04 & 102.84 \\
    LM Arena Proxy
      & 3,401 & 3,496 & 12.06 & 11.10 & -0.96 & 92.01 \\
    \midrule
    \textbf{Median recovery}
      & \multicolumn{6}{c}{99.24\%} \\
    \textbf{Verbosity Change (Ratio of totals)} $\Delta$
      & \multicolumn{6}{c}{$-3.63\%$} \\
    \bottomrule
  \end{tabular}
  \par\smallskip
  \begin{minipage}{\textwidth}
    \scriptsize
    Scores, deltas, and recoveries are rounded to two decimal
    places for display. Calculations use unrounded scores.
    CritPt recovery is undefined because its BF16 score is zero
    and is excluded from the median.
  \end{minipage}
\end{table*}

\subsection{Super 3.5 VL Evaluation and Results}
\label{sec:quant:results}

We evaluate \ourmodel on a comprehensive suite of benchmarks spanning agentic capabilities, reasoning and knowledge, multimodal, conversational ability and instruction following, long-context understanding, and multilingual performance.
All evaluation results for \ourmodel and baselines were collected via Nemo Gym\footnote{\url{https://github.com/NVIDIA-NeMo/Gym}}. For reproducibility purposes, more details on the evaluation settings can be found in the Nemo Gym recipes folder and the corresponding reproducibility tutorial\footnote{\url{https://github.com/NVIDIA-NeMo/Gym/blob/main/nemotron_recipes/super-3.5/reproducibility.md}}.

We evaluate the candidate NVFP4 checkpoints generated from PTQ and QAD to select the most accuracy preserving and verbosity maintaining candidate compared to the baseline BF16. The evaluation and serving parameters are common between the BF16 and NVFP4 for fair comparison.


We measure model intelligence with two metrics: \texttt{Median Accuracy Recovery} and \texttt{Ratio of Totals Verbosity}. The former is computed by averaging accuracy over completed runs for the NVFP4 and BF16 checkpoints, divide the NVFP4 average by the BF16 average, and report the median of these ratios across benchmarks.
The latter is computed by averaging output tokens per trajectory over all trajectories and runs. We then sum these averages across benchmarks for each checkpoint and report the percent change of the NVFP4 total relative to the BF16 total.


Terminal-Bench shows a +0.2\% change in mean output tokens per trajectory over \emph{agent-completed rollouts} and +0.8\% in the median over all rollouts, but +15.7\% in the mean over \emph{all rollouts}, because roughly 4.5\% of non-completing runaway trajectories (about 700k tokens each) dominate the mean. To avoid skewing the verbosity metrics, we exclude the non-completed rollouts for Terminal-Bench; if included the Verbosity Change (Ratio of Totals) increases to \(2.99\%\).
 
We run each benchmark twice and average their scores to account for run-to-run variability, in addition to the number of repeats specified by the benchmark.

Table~\ref{tab:quant:main} summarizes evals between BF16 and NVFP4 checkpoints using FP8 KV cache.



\begin{table*}[t]
  \centering
  \scriptsize
  \setlength{\tabcolsep}{3pt}
  \caption{Super 3.5 BF16 versus released NVFP4 across the full benchmark suite.
    Token columns report average tokens per trajectory; $\Delta$ and
    recovery are calculated from accuracy.}
  \label{tab:quant:main}
  \begin{tabular}{lrrrrrr}
    \toprule
    Benchmark
      & \shortstack{BF16 tokens\\per trajectory}
      & \shortstack{NVFP4 tokens\\per trajectory}
      & BF16 acc.
      & NVFP4 acc.
      & $\Delta$ (pp)
      & Recovery (\%) \\
    \midrule
    \multicolumn{7}{l}{\textbf{Agentic}} \\
    \cmidrule(lr){1-7}
    SWE-Bench Verified         & 22,353 & 21,166 & 71.10 & 71.07 & -0.03 & 99.95 \\
    SWE-Bench Multilingual     & 29,710 & 30,610 & 67.72 & 67.00 & -0.72 & 98.93 \\
    SWE-Bench Pro              & 25,630 & 25,036 & 51.78 & 51.30 & -0.48 & 99.08 \\
    Tau2 Telecom               &  5,784 &  5,659 & 73.19 & 73.19 &  0.00 & 100.00 \\
    Tau2 Retail                &  4,028 &  3,983 & 69.24 & 69.30 & +0.05 & 100.08 \\
    Tau2-AA (Airline)          &  5,426 &  5,224 & 76.50 & 75.75 & -0.75 & 99.02 \\
    Tau3 Banking               & 12,173 & 12,065 & 18.88 & 17.78 & -1.10 & 94.20 \\
    Terminal-Bench v2.1        & 65,560 & 65,717 & 60.67 & 62.78 & +2.11 & 103.47 \\
    GDPVal                     & 49,705 & 52,688 & 49.98 & 55.74 & +5.76 & 111.53 \\

    \addlinespace[0.5em]
    \multicolumn{7}{l}{\textbf{Coding}} \\
    \cmidrule(lr){1-7}
    SciCode (subtask)          &  6,621 &  6,796 & 47.13 & 46.61 & -0.52 & 98.89 \\
    SciCode (problem)          &  6,621 &  6,796 & 15.10 & 14.13 & -0.96 & 93.63 \\
    LiveCodeBench v6           & 15,449 & 15,733 & 88.22 & 87.41 & -0.81 & 99.08 \\
    BIRD (text2sql, no tools) &  637 & 573 & 59.58 & 60.60 & 1.02 & 101.71 \\ 

    \addlinespace[0.5em]
    \multicolumn{7}{l}{\textbf{Math and Science}} \\
    \cmidrule(lr){1-7}
    GPQA Diamond (no tools)    & 14,383 & 14,738 & 89.52 & 89.90 & +0.38 & 100.42 \\
    APEX Shortlist (no tools)  & 68,050 & 66,926 & 88.36 & 89.36 & +1.00 & 101.13 \\
    HLE (no tools, text-only)             & 39,253 & 40,170 & 31.79 & 31.44 & -0.35 & 98.91 \\
    AA-Omniscience Accuracy    &  4,373 &  4,549 & 26.20 & 25.86 & -0.34 & 98.70 \\
    AA-Omni Non-Hallucination  &  4,373 &  4,549 & 69.96 & 72.33 & +2.36 & 103.38 \\
    CritPt  &  7,347 &  7,111 & 10.86 & 11.71 & +0.85 & 107.83 \\

    \addlinespace[0.5em]
    \multicolumn{7}{l}{\textbf{Multimodal}} \\
    \cmidrule(lr){1-7}
    Baby Vision                & 11,839 & 11,634 & 29.60 & 29.51 & -0.09 & 99.71 \\
    CharXiv (RQ)               &  4,157 &  3,930 & 80.82 & 79.88 & -0.93 & 98.85 \\
    MathVision                 &  7,434 &  7,585 & 78.16 & 77.79 & -0.37 & 99.52 \\
    OCR Reasoning              &  5,362 &  5,300 & 65.08 & 65.09 & +0.02 & 100.02 \\
    OCRBenchV2 EN              &  1,860 &  1,919 & 68.12 & 68.34 & +0.22 & 100.32 \\
    OCRBenchV2 ZH              &  2,688 &  2,867 & 59.15 & 59.03 & -0.11 & 99.81 \\
    ScreenSpot Pro             &    236 &    324 & 43.53 & 42.25 & -1.28 & 97.07 \\
    Video-MME-v2 (w/o sub.)    &  4,579 &  5,046 & 32.85 & 32.54 & -0.31 & 99.05 \\
    V$^{*}$                    &    855 &    732 & 85.60 & 85.95 & +0.35 & 100.41 \\
    HLE (full)                 & 36,564 & 37,365 & 29.20 & 29.00 & -0.20 & 99.32 \\
    MMLongBench-Doc            &  3,365 &  3,484 & 62.82 & 62.82 &  0.00 & 100.00 \\

    \addlinespace[0.5em]
    \multicolumn{7}{l}{\textbf{Long context}} \\
    \cmidrule(lr){1-7}
    AA-LCR                     &  4,601 &  4,228 & 68.62 & 67.69 & -0.94 & 98.63 \\
    GraphWalks ($\leq$1M)                 & 10,635 & 10,135 & 89.71 & 90.34 & +0.63 & 100.70 \\

    \addlinespace[0.5em]
    \multicolumn{7}{l}{\textbf{Instruction Following and Chat Preference}} \\
    \cmidrule(lr){1-7}
    IFBench                    & 10,016 &  9,027 & 70.86 & 71.50 & +0.64 & 100.90 \\
    NV Chat Benchmark          &  3,835 &  3,977 & 52.67 & 51.93 & -0.74 & 98.60 \\

    \addlinespace[0.5em]
    \multicolumn{7}{l}{\textbf{Multilingual}} \\
    \cmidrule(lr){1-7}
    MMLU-ProX                  &  3,478 &  3,460 & 76.91 & 73.76 & -3.15 & 95.90 \\
    WMT24++ (en$\rightarrow$xx)&  1,175 &  1,194 & 52.32 & 51.94 & -0.38 & 99.27 \\
    WMT24++ COMET              &  1,175 &  1,194 & 84.80 & 84.76 & -0.04 & 99.95 \\
    \midrule
    \textbf{Median recovery}           & \multicolumn{6}{c}{99.62\%} \\
    \textbf{Verbosity Change (Ratio of totals)}   & \multicolumn{6}{c}{+0.44\%} \\
    \bottomrule
  \end{tabular}
\end{table*}

\clearpage
\section{Inference}
\label{sec:inference}

\ourmodel{} was designed for interactive, long-context agentic workloads,
where requests commonly share system instructions, tool definitions, or
retrieved context and then extend that context over multiple turns.  Effective
serving performance therefore depends not only on peak throughput or
single-request latency, but also on preserving prefix reuse while sustaining
an interactive per-user generation rate.

For system tuning, we use a 20-turn proxy workload with
2,048 input tokens and 1,024 output tokens per turn; the shared
prompt and prior conversation are eligible for prefix reuse on
subsequent turns. We call a serving point \emph{usable} when
its reported per-user generation rate is at least
50 tokens/s. Among these points, we seek the highest aggregate
output throughput. The selected operating point depends on
the platform and serving configuration.

This regime places particular emphasis on three parts of the serving stack.
First, prefix reuse must remain effective for a hybrid Mamba-Attention model
when speculative decoding changes the position from which execution resumes.
Second, kernel selection must be controllable at finer granularity than a
single global policy because the best kernels at a balanced operating point
need not be those that win at either endpoint of the latency-throughput
curve.  Third, the same model must map efficiently to accelerators with
different memory capacities, bandwidths, and quantization support.  We address
these requirements in vLLM and characterize their effect below.

\subsection{Efficiency}
\label{sec:quant:perf}
\begin{figure}[t]
    \centering
    \IfFileExists{figures/figure_10_nvfp4_vs_bf16.pdf}{%
        \includegraphics[width=0.72\linewidth]{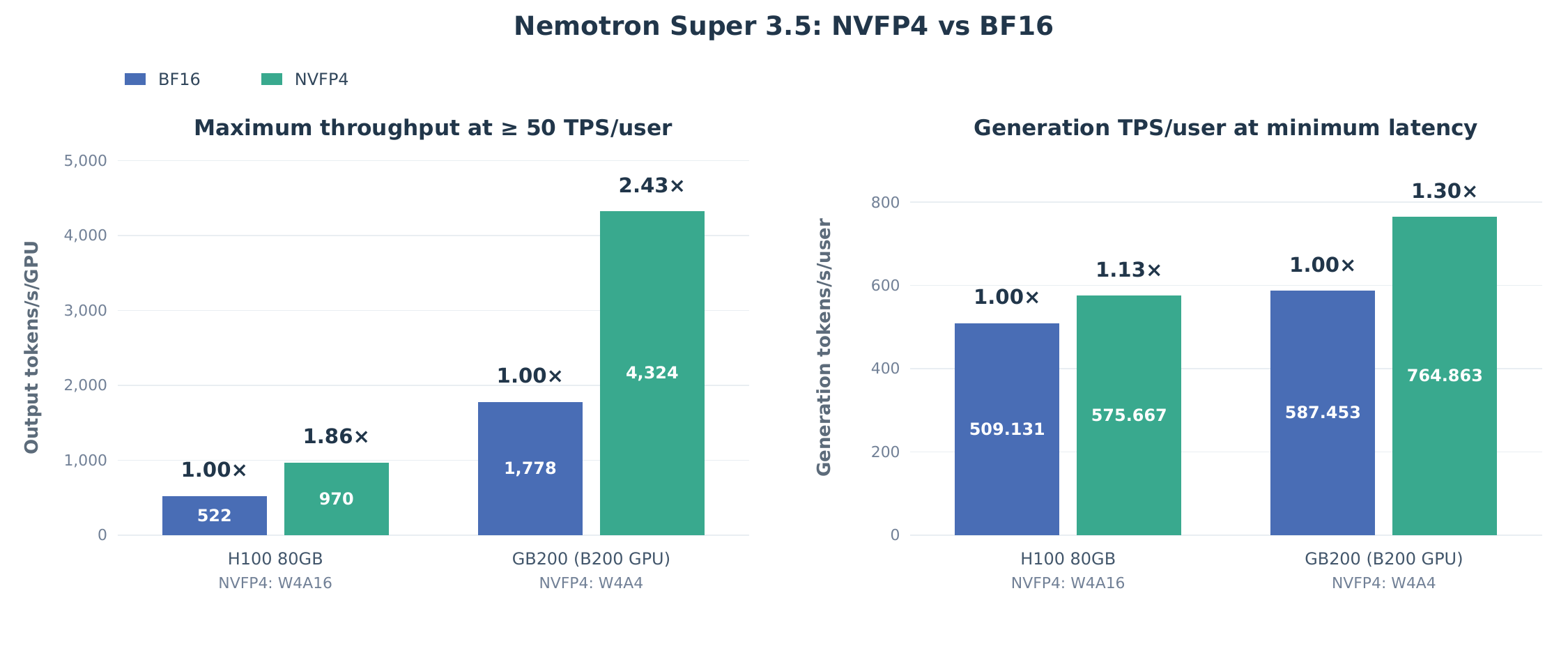}%
    }{%
        \fbox{\parbox[c][1.65in][c]{0.68\linewidth}{\centering
        Placeholder: representative throughput--interactivity frontier}}%
    }
    \caption{Serving performance using vLLM on the 20-turn agentic proxy workload, with 2K input tokens and 1K output tokens per turn and ReplaySSM disabled. The graphs compare NVFP4 and BF16 maximum throughput at >=50 tokens/s/user and minimum latency. }
    \label{fig:quantization:efficiency}
\end{figure}

Figure~\ref{fig:quantization:efficiency} compares the max-throughput at 50 tps/user, and min-latency performance of \ourmodel{} on H100 and GB200 for 50k prefix and 20 turns of 2k/1k.

\subsection{Performance across deployment platforms}
\label{sec:inference:platforms}

\paragraph{Cross-platform performance.}
We validated \ourmodel{} across NVIDIA datacenter and local systems, including
A100, H100, B200, GB200 and GB300 platforms, DGX Spark and DGX Station, and
RTX-class GPUs.  These platforms span multiple architecture generations and
deployment scales.  Rather than impose one serving configuration everywhere,
we tune quantization and speculative decoding for each platform.
The objective remains the same: maximize aggregate throughput over the 20-turn
proxy workload while sustaining at least 50 output tokens/s/user.

\begin{figure}[t]
    \centering
    \IfFileExists{figures/nemotron_super_35_serving.pdf}{%
        \includegraphics[width=0.72\linewidth]{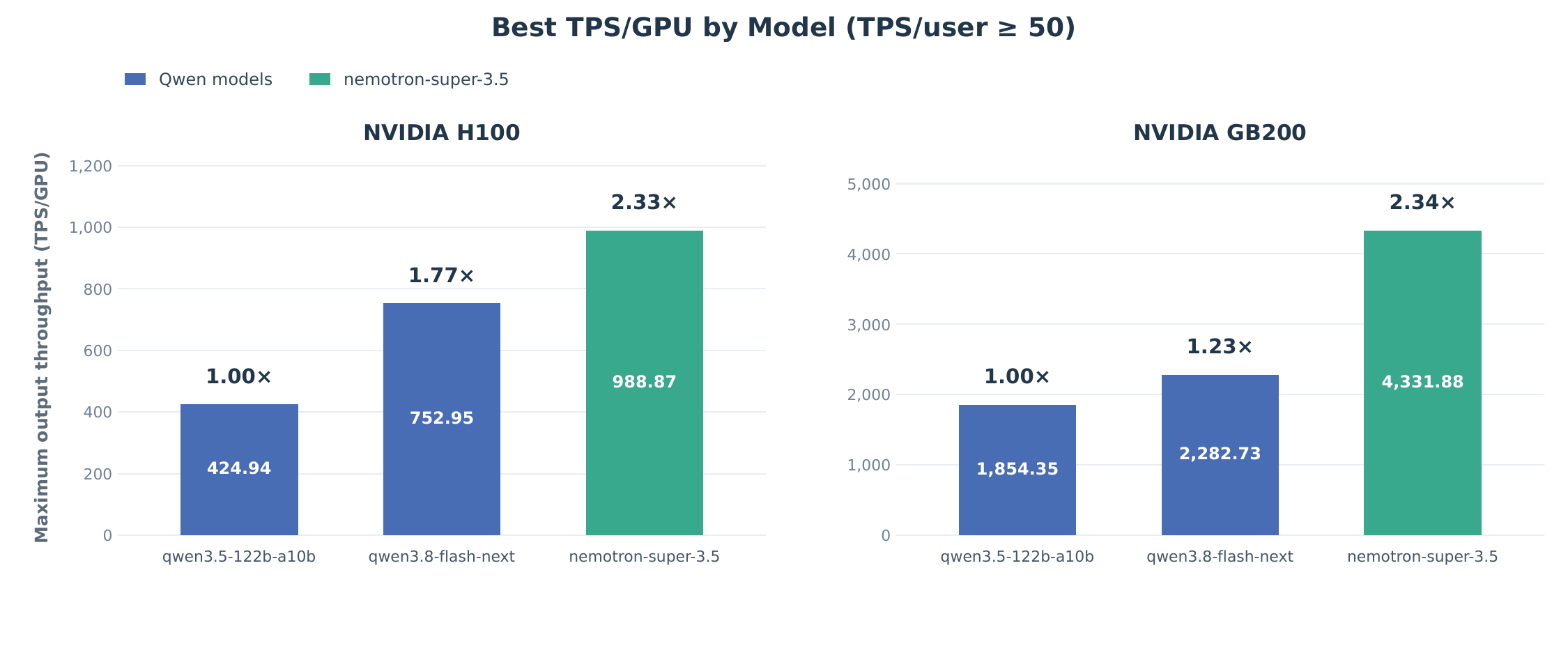}%
    }{%
        \fbox{\parbox[c][1.65in][c]{0.68\linewidth}{\centering
        Placeholder: representative throughput--interactivity frontier}}%
    }
    \caption{Serving performance using vllm for the 20-turn agentic proxy workload, with
    2K input tokens and 1K output tokens per turn. The graph shows the max-throughput of each model at tps/user >= 50 .}
    \label{fig:inference:performance}
\end{figure}

Figure~\ref{fig:inference:performance} compares the resulting frontier with a
Qwen model of comparable size on a representative platform.  \ourmodel{}
provides higher performance in the region relevant to interactive serving.
The comparison is made at a matched workload, with checkpoint, precision, and
system configuration reported in the final caption; points below
50 output tokens/s/user are excluded from the usable region even when they
offer high aggregate throughput.

\paragraph{Performance Modeling and Analysis}
Our internal performance model attributes the advantage in
Figure~\ref{fig:inference:performance} to how \ourmodel{} handles per-request
state, how many tokens each decoding step yields, and how much data its
experts exchange. In each decode step the weights are read once and shared by
the whole batch, while every request reads and updates its own KV cache and
recurrent state, which must also stay in GPU memory. Throughput therefore
depends on how cheaply each request produces a token and on how many requests
share the cost of reading the weights.

Like the Qwen models, \ourmodel{} is a hybrid, but it has fewer attention
layers, with narrower key-value heads, and leaves the rest of sequence mixing
to Mamba-2 layers whose state does not grow with context, so its per-request
memory grows much more slowly as agentic sessions accumulate history. At these
lengths each Qwen3.8 request needs more GPU memory, so fewer requests fit on a
GPU, which limits how many share each weight read; its sparse attention
reduces how much KV cache is read per step, but not how much must be stored.
Qwen3.5 stores about as much KV cache as Qwen3.8 and reads all of it on every
step, so each of its requests costs more in both memory and bandwidth than a
\ourmodel{} request.

The MTP head reaches a higher average acceptance length than both Qwen models
under probabilistic draft sampling (Table~\ref{tab:mtp_speedbench}), so each
step yields more tokens per request, and LatentMoE exchanges latent vectors
that are smaller than the Qwen models' hidden states, which reduces the
expert-parallel communication per token. A large per-request cost in
\ourmodel{}'s decoding is updating the Mamba state, which speculative decoding
writes out for every token verified in a step because the number of accepted
tokens is not known in advance; ReplaySSM
(Section~\ref{sec:inference:kernel-control}) reduces this traffic and the
memory it requires.

\subsection{Prefix reuse and speculative decoding}
\label{sec:inference:prefix-specdec}

\paragraph{Hybrid-model cache checkpoints are coarse.}
For a pure Attention model, prefix reuse is represented by per-token KV state
stored in relatively small cache pages.  A hybrid model must additionally
recover the recurrent Mamba state at the position where computation resumes.
This state contains the complete sequence summary for a layer and is much
larger per cached position than an Attention KV entry.  vLLM therefore matches
the cache-page sizes of the Attention and Mamba cache groups by byte size.  On
our target Mamba-2 model, one Mamba cache page occupies at least as many bytes
as approximately 4K tokens of Attention KV cache. Consequently, Mamba states
are normally retained on a much coarser token grid than Attention KV state.

This granularity matters when prefix caching is combined with speculative decoding.  Suppose the target model obtains a prefix-cache hit
through token $T$.  To draft token $T+1$, the drafter needs the target model's
intermediate hidden state for token $T$.  The prefix cache contains the
Attention KV state and a checkpointed Mamba state, but not that intermediate
hidden state.  vLLM reconstructs it by resuming the target model from the most
recent stored Mamba checkpoint.  If the Mamba block spans $L_{\mathrm{page}}$ tokens, the
nearest resumable state may be at $T-L_{\mathrm{page}}$, forcing the target model to replay
tokens $T-L_{\mathrm{page}}$ through $T$ solely to recover the hidden state at token $T$.

We improved this path by optionally retaining an additional Mamba cache page
near the speculative-decoding replay boundary.  For example, if a workload uses a 16-token
fine-grained prefix-match unit, the additional checkpoint can be placed near
$T-16$ rather than $T-L_{\mathrm{page}}$.  A subsequent request then replays only the final
16 target-model tokens instead of an entire Mamba block.  The scheduler and
cache manager jointly select and publish a valid checkpoint at the shared
prefix boundary, avoiding stale recurrent state while trading one additional
Mamba page for substantially less target-model recomputation.\footnote{\url{https://github.com/vllm-project/vllm/pull/53945}}

The issue is less pronounced for non-SSM models because their cache pages span
far fewer tokens, so replaying from the preceding page boundary is inexpensive.
It is also more important for \ourmodel{} than for hybrid SSM models with
smaller recurrent states: Mamba-2's larger cache-page footprint makes $L_{\mathrm{page}}$
larger and amplifies the work avoided by the finer checkpoint.  The resulting
mechanism makes prefix caching and speculative decoding composable: prefix
caching removes repeated prefill work, while speculative decoding reduces the
number of target-model decode steps after prefill.

\subsection{Kernel control at the target operating point}
\label{sec:inference:kernel-control}

\paragraph{Per-quantization GEMM backend selection.}
The kernel that maximizes throughput at saturation is not necessarily the
kernel that gives the best end-to-end result at an intermediate batch size.
Our target lies in precisely this intermediate regime: we need enough batching
to obtain high aggregate throughput while preserving at least
50 output tokens/s/user.  Furthermore, a mixed-precision checkpoint may
contain linear layers with different quantization schemes, whose GEMM shapes
and best backends differ.  A single process-wide backend selection therefore
couples decisions that should be tuned independently.

The global linear backend remains the default, while a kernel configuration
can override it for a specific quantization scheme - for example, selecting a
specialized backend for NVFP4 W4A16 layers while leaving other linear layers
on the platform default. This finer control lets a deployment choose the
best-supported GEMM implementation for each quantization scheme and accelerator
without maintaining separate model artifacts or forcing every quantized layer
through the same path. It also makes the serving configuration portable:
Hopper, Blackwell, and compact systems can use different backend maps while
preserving the same model and workload-level objective. \footnote{\url{https://github.com/vllm-project/vllm/pull/51204}}

\paragraph{Checkpointing replay for SSM cache quantization.}
One inference performance improvement grows out of a suggestion in the Nemotron v3 Ultra Technical report.  During a decoding step of the Mamba-2 layer, for non-speculative decoding we have the recurrent state $S$ and per-token inputs key vector $k$, value vector $v$, decay scalar $\alpha$, and query vector $q$.  The new recurrent state is then $S_{new} = \alpha S_{old}  + k^\top v$, and the output $o = q S_{new}$.  For speculative decoding we do this for the $T$ incoming tokens' worth of input, writing out each possible new state as we do not know how many tokens will be accepted.  This results in decoding being extremely memory bound, and it also limits batch\_size * draft\_length for speculative decoding due to limited memory capacity.

We have since implemented a "checkpointing replay" method.  We store a checkpointed state $S_{ckpt}$ and store the $k_i,v_i,\alpha_i$ seen since the checkpoint in a small buffer.  As $S = (\prod \alpha_i) S_{ckpt} + \sum_i (\prod_{j>i+1} \alpha_j)  k_i^\top v_i$ we can reconstruct the state on-demand, and with no quantization error beyond what's in $S_{ckpt}$.  Rejected tokens are handled by simply decrementing a counter, and new inputs are appended to the buffer.  To generate output tokens, we can either reconstruct the state or we can generate output tokens more cheaply directly, by noting that (ignoring the $\alpha$ terms for exposition) $q S = q S_{ckpt} + \sum_i (q k_i^\top) v_i$.  When our buffer fills up, we 'flush' or 'fold' it by reconstructing the state and writing it out.  For speculative decoding we only include the accepted tokens from our history, with that step's tokens seeding the start of the new history buffer.

This amortizes the state write over our buffer length, reducing our per-step state-related memory traffic from $T+1$ to roughly $1 + 1/L$.  It also lets us amortize any computational overhead associated with writing the state, such as randomness generation, stochastic rounding, or scaling calculations.
Especially for speculative decoding, the resulting formulae look a lot like chunked prefill, with the parts moved around.  Like chunked prefill, the FLOP cost goes up but we get to move CUDA core work, namely matvecs and outer products, over to Tensor Cores, so actual compute time goes down.
In summary, this method:
\begin{itemize}
  \item Reduces memory traffic on an IO bound operation.
  \item Reduces compute time by shifting work to Tensor Cores.
  \item Reduces quantization error, by quantizing less often.
  \item Enables larger batch sizes and speculation lengths.
\end{itemize}
For more details, see https://github.com/NVIDIA/TensorRT-LLM/pull/14203.  We note this technique was discovered concurrently and independently in the Dao lab, where it is called ReplaySSM, as described in \citet{liou2026replayssmcachessminputs} and in their blog post (https://dao-lab.ai/blog/2026/replayssm/); we subsequently collaborated with them on the paper and on bringing it to vLLM.

Table~\ref{tab:replayssm-phase12} compares ReplaySSM enabled and
disabled for Nemotron Super 3.5 on a single NVIDIA GB200 GPU, using a 16-step ReplaySSM buffer and default KV-cache dtype.
Across the selected matched configurations, ReplaySSM improves
output throughput by 4.7--6.4\% and per-user generation throughput
by 6.4--12.5\%.

The workload comprises 20 turns
with a 51,200-token shared prefix, 2,048-token input per turn,
and 1,024-token output.
\begin{center}
\begin{minipage}{\linewidth}
    \centering
    \small
    \setlength{\tabcolsep}{4pt}

    \begin{tabular}{rr rrr rrr}
      \toprule
      & & \multicolumn{3}{c}{Output TPS/GPU}
        & \multicolumn{3}{c}{Gen. TPS/user} \\
      \cmidrule(lr){3-5} \cmidrule(lr){6-8}
      Conc. & MTP & OFF & ON & Gain & OFF & ON & Gain \\
      \midrule
      64  & 3 & 4,374 & 4,628 & +5.8\%
              & 80.2 & 86.0 & +7.2\% \\
      64  & 4 & 4,458 & 4,729 & +6.1\%
              & 81.6 & 86.9 & +6.4\% \\
      64  & 5 & 4,632 & 4,856 & +4.8\%
              & 83.0 & 89.9 & +8.4\% \\
      64  & 7 & 4,629 & 4,846 & +4.7\%
              & 81.3 & 91.5 & +12.5\% \\
      128 & 1 & 4,566 & 4,859 & +6.4\%
              & 38.1 & 41.9 & +10.0\% \\
      \midrule
      128 & 4 & N/A & \textbf{5,728} & N/A
              & N/A & 49.1 & N/A \\
      \bottomrule
    \end{tabular}

    \captionof{table}{ReplaySSM OFF/ON for Nemotron Super 3.5
    on one GB200 running vLLM, using a 16-step buffer.
    MTP denotes speculative tokens; TPS denotes tokens/s.
    Gains are relative to OFF, calculated before rounding.
    N/A indicates non-optimal values.
    Results are individual measurements.}
    \label{tab:replayssm-phase12}
\end{minipage}
\end{center}

\subsection{Reducing the cost of Mamba checkpoints}
\label{sec:inference:internal-checkpoints}

\paragraph{Internal prefill checkpoints.}
In aligned Mamba cache mode, a reusable recurrent state must be materialized at
a cache-block boundary.  The scheduling path can enforce this by ending a
prefill step at that boundary and issuing an additional model forward for the
remainder of the prompt.  Although this produces the required state, the split
repeats fixed forward-pass overhead and adds another traversal of the
Attention, MoE, routing, and tensor-parallel communication paths.

Internal prefill checkpointing was initially implemented in vLLM for
Kimi~K3.  Building on that work, we adapted the shared checkpoint builder and
exporter to Mamba-2 in PR~\#57329.\footnote{\url{https://github.com/vllm-project/vllm/pull/57329}}
The Mamba scan already materializes intermediate recurrent states at logical
chunk boundaries.  The adapted path arranges for a logical chunk to end at the
desired cache boundary, extracts the state and convolution window there, and
writes that checkpoint into the paged cache while the full prompt remains in
one model forward.  A prompt that previously required an additional forward
after the checkpoint can therefore be processed in a single forward without
sacrificing the reusable Mamba state.

Taken together, these changes target the regime that matters most for
\ourmodel{} deployment: shared and growing contexts, interactive decode rates,
and hardware-specific configurations chosen from the useful portion of the
throughput-latency frontier.  Prefix-aware speculative decoding reduces both
repeated prefill and decode work, per-quantization backend selection exposes
the kernel choices needed to tune balanced operating points, and internal
Mamba checkpoints reduce the cost of creating the reusable state on which the
workload depends.

\subsection{Disaggregated serving on NVIDIA Vera Rubin}
\label{sec:vr200-disagg}

We serve the Nemotron 3.5 Super VL NVFP4 checkpoint with NVIDIA Dynamo and vLLM on NVIDIA Vera Rubin (VR200) GPUs, with one GPU per model instance (no tensor parallelism).
The workload replays multi-turn agentic coding sessions from SWE-bench Pro traces, with long contexts and high prefix reuse across turns.
Speculative decoding uses the MTP head with a fixed acceptance length of 4.06 and draft length 5.
A usable point keeps the median output speed at or above 50 tokens/s/user and the median time to first token (TTFT) below 5~s.
We report output tokens/s per GPU at the highest-throughput load that meets this target.

Prefill reads the prompt and builds the KV cache, the stored attention state of earlier tokens.
Decode then generates the output tokens.
We compare two deployments on 11 GPUs each.
The aggregated deployment runs 11 independent instances that each do prefill and decode.
A KV-cache-aware router sends each request to an instance.
The disaggregated deployment uses 7 GPUs for prefill only and 4 GPUs for decode only.
After prefill, the KV cache moves from the prefill GPU to the decode GPU.

Disaggregated serving gives 10.3\% more output tokens/s per GPU than aggregated serving (Table~\ref{tab:vr200-disagg}).
At its best point, the median output speed per user is also 61\% higher.
The median TTFT increases from 0.49~s to 1.12~s, but stays well below the 5~s target.
Prefill and decode no longer share a GPU.
As a result, the decode GPUs run larger decode batches without long prefills between decode steps.

\begin{table}[t]
\centering
\small
\begin{tabular}{lrrrr}
\toprule
Deployment & Conc. & TPS/GPU & TPS/user & TTFT (s) \\
\midrule
Aggregated    & 616 & 2,969 & 96.6  & 0.49 \\
Disaggregated & 528 & 3,276 & 155.9 & 1.12 \\
\midrule
Gain          &     & +10.3\% & +61\% &      \\
\bottomrule
\end{tabular}
\caption{Best usable point for Nemotron 3.5 Super VL NVFP4 on 11 NVIDIA VR200 GPUs with Dynamo and vLLM.
The workload replays SWE-bench Pro agentic coding sessions.
A usable point has a median output speed of at least 50 tokens/s/user and a median TTFT below 5~s.
Disaggregated uses 7 prefill and 4 decode GPUs. TPS/user and TTFT are medians.
MTP uses a fixed acceptance length of 4.06 and draft length 5. TPS denotes tokens/s.}
\label{tab:vr200-disagg}
\end{table}

\paragraph{Precision and kernels.}
Both deployments use the engine settings that follow.
The weights use NVFP4 and the KV cache uses FP8.
The Mamba state cache uses FP16 with stochastic rounding, as in Section~4.
The compiler fuses each attention operation with the quantization of its output.
The engine captures CUDA graphs for full and piecewise model steps.
In decode attention, one wave of GPU work can allow only one or two splits of the KV cache per request. The multi-wave KV split then splits the KV cache further and runs the work in several waves. We added this option to FlashInfer in PR \#6331\footnote{\url{https://github.com/flashinfer-ai/flashinfer/pull/6331}} and to vLLM in PR \#61077.\footnote{\url{https://github.com/vllm-project/vllm/pull/61077}}

\paragraph{Speculative decoding and cache reuse.}
The MTP head drafts 5 tokens per step.
Prefix caching stores Mamba states at KV-cache block boundaries, plus extra checkpoints at prefixes that several requests share.
A vLLM option (\texttt{disable\_eagle\_block\_drop}) lets a cache hit keep its last full block when speculative decoding is on.
Each aggregated instance and each prefill GPU also has a 240~GiB KV cache tier in host (CPU) memory.
This tier keeps the prefixes of earlier turns after the GPU cache evicts them.

\paragraph{Routing.}
Each worker publishes KV-cache events, so the router knows which cache blocks each worker holds.
Session affinity sends all turns of one session to the same worker, so later turns find their prefix in its cache.
In the disaggregated deployment, NIXL moves the KV cache from the prefill GPU to the decode GPU.

\providecommand{\pending}[1]{\textcolor{red}{\textbf{[PENDING:} #1\textbf{]}}}
\providecommand{\tbd}{\textcolor{red}{\textbf{TBD}}}

\subsection{Speculative Decoding with Parallel Drafters}
\label{sec:lilicorr}
Alongside the native MTP head (Sections~\ref{sec:mtp_init} and~\ref{subsec:mtp_boosting}), we release LiLiCorr~\citep{rusanovsky2026lilicorrlightweightlikelihoodcorrelation}, a standalone drafter for Nemotron 3.5 Super that targets low-concurrency serving, where decoding is strongly memory-bandwidth bound. This release builds on our earlier parallel-drafter work for Nemotron 3.5 Lightning, for which we trained and released separate DFlash~\citep{dflash2025} and DSpark~\citep{cheng2026dspark} checkpoints. The Nemotron 3.5 Lightning DFlash uses non-causal, full-sequence grouped-query attention, whereas the released DSpark uses causal grouped-query attention with sliding-window attention in every layer and per-head attention sinks. Bringing these models to release also required W4A16 post-training quantization (Section~\ref{sec:quant:drafter}) and vLLM optimizations for the DSpark Markov head.

In the low-concurrency regime targeted here, verifying a block of draft tokens costs similarly to generating a single token, so the speedup is determined jointly by the acceptance length and the latency of producing the draft. The recursive MTP head drafts one token per pass, so its drafting latency grows with the draft length, while parallel drafters propose all draft tokens in a single forward pass whose latency is nearly independent of the draft length. MTP is part of the model checkpoint and requires no additional weights, whereas DFlash, DSpark, and LiLiCorr are separate drafter checkpoints aimed at this regime.

A parallel drafter is trained on the marginal distribution at each block position, henceforth a slot, so tokens selected independently at each slot can be individually plausible yet jointly incoherent. DSpark addresses this with a semi-autoregressive design. It keeps the drafter body parallel and adds a lightweight sequential head that biases the logits at each slot using a low-rank transition from the token drafted at the previous slot. In the full DSpark design, a separate confidence head estimates the probability that each draft prefix will be accepted, allowing the serving engine to adapt the verification length to the request and system load. The Lightning 3.5 release used the lightweight Markov correction but omitted this confidence head. LiLiCorr instead uses a lightweight, single-pass correlation head to score compatibility across slots.

\subsubsection{Training LiLiCorr for Super3.5}
\label{sec:lilicorr_training}

\paragraph{Introduction.}
At each of the $B-1$ drafted slots, LiLiCorr retains the $K$ most probable tokens of the drafter body's marginal as candidates and processes the resulting $(B-1)\times K$ candidate lattice jointly with a small Transformer, so that each candidate is represented in the context of all others in the block. For the $k$-th candidate at slot $s$, with $k = 1, \dots, K$ and $s = 1, \dots, B-1$, the head emits a unit-norm \emph{out} vector $\mathbf{o}_{s,k}$ and \emph{in} vector $\mathbf{i}_{s,k}$. Slot $0$ holds the last token committed by the target. Since this token is already fixed, it is the only candidate at slot $0$ and carries a single out vector $\mathbf{o}_{0,1}$. The compatibility of candidate $k$ at slot $s$ with candidate $k'$ at slot $s+1$ is the cosine similarity between the out vector of the former and the in vector of the latter:
\begin{equation}
  \phi(s,k,k') = \sigma \cos\!\left(\mathbf{o}_{s,k},\, \mathbf{i}_{s+1,k'}\right), \qquad s = 0, \dots, B-2,
  \label{eq:lilicorr_score}
\end{equation}
where $\sigma$ is a fixed scale. Writing $\hat{c}_{s}$ for the index of the candidate committed at slot $s$, with $\hat{c}_{0} = 1$, the block is committed from left to right, each slot selecting the candidate most compatible with its committed predecessor:
\begin{equation}
  \hat{c}_{s} = \arg\max_{k}\, \phi\!\left(s-1,\, \hat{c}_{s-1},\, k\right), \qquad s = 1, \dots, B-1.
  \label{eq:lilicorr_walk}
\end{equation}
Under probabilistic draft sampling, each slot is instead drawn from the head's distribution over its candidates at the request's temperature. All vectors come from one pass of the head and the pairwise scores in Eq.~\eqref{eq:lilicorr_score} reduce to batched matrix products, so only the walk is sequential, and it requires no further network evaluation. Verification is unchanged, so the output distribution of the target is preserved exactly.

\paragraph{Setup.} We train the drafter body and the LiLiCorr head jointly in NVIDIA Model Optimizer (ModelOpt)~\citep{nvidia-modelopt}, in a streaming configuration suited to a target of this size. Replicas of the target's language model are served in BF16 by vLLM~\citep{kwon2023efficient}, each with tensor parallelism of four, and stream the hidden states of five captured target layers, together with the final hidden state, to $64$ trainer GPUs. The target's next-token distribution, which the objective below requires, is reconstructed on the trainer from the final hidden state through the target's own LM head, so the target body is never replicated on the trainer ranks. The drafter uses a block of $B=8$ positions, i.e., $7$ drafted slots. Each of the five layers of the drafter body wraps the attention and MLP sublayers in a grouped dynamic convolution inspired by DFlash2~\citep{dflash2_2026}, which lets neighboring slots exchange local information while the block remains parallel. The LiLiCorr head keeps the configuration of the original work. The resulting drafter has $1.13$B parameters, of which $37$M belong to the head.

\paragraph{Objective.} The original work trains the drafter body with a cross-entropy against the tokens of the target-generated training responses, weighted by the exponential position decay of DFlash. We depart from this recipe and, as in MTP boosting (Section~\ref{subsec:mtp_boosting}), instead distill the full next-token distribution $p_M$ of the target model $M$, which is available from the streamed hidden states, while keeping the same decay. Consider a training block that starts at position $t$ of a training sequence $x$, with $x_t$ at slot $0$, so that slot $s$ predicts the token $x_{t+s}$ from the prefix $x_{<t+s}$. Its loss is
\begin{equation}
  \mathcal{L}_{\mathrm{DFlash}} = -\sum_{s=1}^{B-1} w_{s} \sum_{v \in \mathcal{V}} p_{M}\!\left(v \mid x_{<t+s}\right) \log p_{\theta,s}(v), \qquad w_{s} = \exp\!\left(-\frac{s-1}{\beta}\right),
  \label{eq:lilicorr_kd}
\end{equation}
where $\mathcal{V}$ is the vocabulary, $p_{\theta,s}$ is the marginal distribution over $\mathcal{V}$ that the drafter body, with parameters $\theta$, predicts at slot $s$, and $\beta$ sets the decay. The loss is averaged over blocks whose starting positions are sampled from each sequence. Up to a constant, this is the forward KL divergence from $p_M$ to $p_{\theta,s}$. Because the training responses are sampled at temperature $1$, a response token need not be the target's most likely one. The full distribution supervises both the target's most likely token, which greedy verification tests, and the shape of its distribution, which sampled verification tests.

Let $c_{s} \in \{1, \dots, K\}$ denote the index of the candidate chosen at slot $s$, with $c_{0} = 1$. The head defines, at each slot, a distribution over its $K$ candidates conditioned on the candidate chosen at the previous slot,
\begin{equation}
  P_{\theta,\psi}\!\left(c_{s}=k \mid c_{s-1}\right) = \mathrm{softmax}_{k}\, \phi\!\left(s-1,\, c_{s-1},\, k\right),
  \label{eq:lilicorr_cond}
\end{equation}
where $\psi$ denotes the parameters of the head. The distribution depends on $\theta$ as well, because the head's inputs include the log-probabilities of the candidates under $p_{\theta,s}$ and the drafter body's hidden states. The head's loss terms therefore train the drafter body as well. Let $g_{s}$ be the index of the candidate equal to the ground-truth token $x_{t+s}$, when it is among the $K$ candidates, with $g_{0} = 1$. The oracle prefix of a block is its longest leading run of slots $1, \dots, m$ that all contain the ground-truth token, and $\mathcal{S}$ collects the slots on the oracle prefixes of all blocks in a training batch. The head is teacher-forced on the ground-truth predecessor, and its cross-entropy term is that of the original work:
\begin{equation}
  \mathcal{L}_{\mathrm{CE}} = -\frac{1}{|\mathcal{S}|} \sum_{s \in \mathcal{S}} \log P_{\theta,\psi}\!\left(c_{s}=g_{s} \mid c_{s-1}=g_{s-1}\right).
  \label{eq:lilicorr_ce}
\end{equation}

The original work complements $\mathcal{L}_{\mathrm{CE}}$ with a distractor penalty, which penalizes the probability the head assigns to candidates that the target ranks below the ground truth, in proportion to the target's log-probability gap. We replace it with a calibration term. Let $q_{s,k} = p_{M}\!\left(v_{s,k} \mid x_{<t+s}\right)$ be the target probability of $v_{s,k}$, the token of the $k$-th candidate at slot $s$, and $\bar{q}_{s,k} = q_{s,k} / \sum_{k'} q_{s,k'}$ its renormalization over the $K$ candidates. The calibration term is the cross-entropy from $\bar{q}$ to the head's per-slot distribution:
\begin{equation}
  \mathcal{L}_{\mathrm{cal}} = -\frac{1}{|\mathcal{S}|} \sum_{s \in \mathcal{S}} \sum_{k=1}^{K} \bar{q}_{s,k} \log P_{\theta,\psi}\!\left(c_{s}=k \mid c_{s-1}=g_{s-1}\right).
  \label{eq:lilicorr_cal}
\end{equation}
The distractor penalty is minimized by any distribution that places all of its probability on candidates the target ranks at or above the ground truth. The calibration term instead has a unique minimizer, the target's renormalized distribution $\bar{q}_{s,\cdot}$ over the candidates of each slot. This matters under probabilistic draft sampling, which at $temp=1$ draws each draft token from the head's per-slot distribution. The drafter body and the head are trained jointly on
\begin{equation}
  \mathcal{L} = \mathcal{L}_{\mathrm{DFlash}} + \lambda_{\mathrm{CE}}\, \mathcal{L}_{\mathrm{CE}} + \lambda_{\mathrm{cal}}\, \mathcal{L}_{\mathrm{cal}},
  \label{eq:lilicorr_objective}
\end{equation}
where the weights $\lambda_{\mathrm{CE}}$ and $\lambda_{\mathrm{cal}}$ are given in Table~\ref{tab:lilicorr_hparams}.

\paragraph{Data.} We train in two stages, which draw on the same $6.32$M prompts from $12$ Nemotron post-training datasets, listed in Table~\ref{tab:lilicorr_data}, and differ in the checkpoint that generates the responses and serves as the streaming target. In the first stage this is a post-SFT Nemotron 3.5 Super checkpoint. In the second it is the released post-RL checkpoint, so that the drafter adapts to the hidden states and output distribution of the model it serves. In the first stage a randomly selected subset of the assistant turns of each sample is regenerated, and each sample ends at a regenerated turn. In the second stage all assistant turns are regenerated. Generation uses reasoning, temperature $1.0$ with top-$p$ of $0.95$, and is limited to $8{,}192$ new tokens. User turns and tool results are kept verbatim, and no tool is executed. The loss is applied only to the final assistant turn, which is always a regenerated one. The first stage trains on windows of $4{,}096$ tokens and the second on windows of $8{,}192$ tokens. After filtering, the first stage retains $5.67$M conversations totaling $22.6$B tokens, of which $14.9$B fall within its training window, and the second stage retains $11.01$M of $18.2$M generated samples totaling $35.6$B tokens, of which $34.9$B fall within its training window.

\begin{table}[t]
  \centering
  \small
  \setlength{\tabcolsep}{4pt}
  \begin{tabular}{l>{\raggedright\arraybackslash}p{3.6cm}r}
    \toprule
    Dataset & Splits & Prompts \\
    \midrule
    \href{https://huggingface.co/datasets/nvidia/Nemotron-Competitive-Programming-v1}{Nemotron-Competitive-Programming-v1} & InfiniByte & $1{,}174{,}694$ \\
    \href{https://huggingface.co/datasets/nvidia/Nemotron-SFT-Agentic-v2}{Nemotron-SFT-Agentic-v2} & interactive agent, search, tool calling & $991{,}900$ \\
    \href{https://huggingface.co/datasets/nvidia/Nemotron-SFT-Instruction-Following-Chat-v2}{Nemotron-SFT-Instruction-Following-Chat-v2} & reasoning on & $929{,}237$ \\
    \href{https://huggingface.co/datasets/nvidia/Nemotron-Post-Training-Dataset-v2}{Nemotron-Post-Training-Dataset-v2} & chat, code, math, STEM, multilingual$^{\dagger}$ & $800{,}000$ \\
    \href{https://huggingface.co/datasets/nvidia/Nemotron-RL-Super-Training-Blends}{Nemotron-RL-Super-Training-Blends} & RLHF, RLVR, SWE & $479{,}303$ \\
    \href{https://huggingface.co/datasets/nvidia/Nemotron-SFT-OpenCode-v1}{Nemotron-SFT-OpenCode-v1} & all six subsets & $460{,}254$ \\
    \href{https://huggingface.co/datasets/nvidia/Nemotron-Instruction-Following-Chat-v1}{Nemotron-Instruction-Following-Chat-v1} & chat, structured outputs & $430{,}978$ \\
    \href{https://huggingface.co/datasets/nvidia/Nemotron-SFT-SWE-v2}{Nemotron-SFT-SWE-v2} & agentless, SWE & $256{,}254$ \\
    \href{https://huggingface.co/datasets/nvidia/Nemotron-Science-v1}{Nemotron-Science-v1} & MCQ, RQA & $226{,}334$ \\
    \href{https://huggingface.co/datasets/nvidia/Nemotron-SFT-Math-v3}{Nemotron-SFT-Math-v3} & subsample & $200{,}000$ \\
    \href{https://huggingface.co/datasets/nvidia/Nemotron-SFT-Multilingual-v1}{Nemotron-SFT-Multilingual-v1} & subsample & $200{,}000$ \\
    \href{https://huggingface.co/datasets/nvidia/Nemotron-SFT-Competitive-Programming-v2}{Nemotron-SFT-Competitive-Programming-v2} & Exercism, text-to-SQL & $175{,}808$ \\
    \midrule
    Total & & $6{,}324{,}762$ \\
    \bottomrule
  \end{tabular}
\caption{Prompt sources of both training stages. After filtering, the first stage retains $5.67$M conversations and $22.6$B tokens, of which $14.9$B fall within its $4{,}096$-token training window. The second stage retains $11.01$M of $18.2$M generated samples and $35.6$B tokens, of which $34.9$B fall within its $8{,}192$-token training window. $^{\dagger}$200K chat, all 175K code, 200K math, and 100K STEM prompts, plus 25K prompts in each of German, Spanish, French, Italian, and Japanese.}
  \label{tab:lilicorr_data}
\end{table}

\paragraph{Hyperparameters.} The first stage trains from random initialization for $3$ epochs of $88{,}539$ steps each, $265{,}617$ steps in total, on $4{,}096$-token windows. The second stage is initialized from the full-precision weights at the end of the first stage and trains for $500{,}000$ steps, about $2.9$ epochs of $172{,}037$ steps each, on $8{,}192$-token windows, with a freshly initialized optimizer and schedule. The target is served with four replicas in the first stage and eight in the second. Table~\ref{tab:lilicorr_hparams} compares the configuration with that of the original work.

\begin{table}[t]
  \centering
  \small
  \setlength{\tabcolsep}{3.5pt}
  \begin{tabular}{lll}
    \toprule
    Setting & Original~\citep{rusanovsky2026lilicorrlightweightlikelihoodcorrelation} & Nemotron 3.5 Super \\
    \midrule
    Block size (drafted slots) & $16$ ($15$) & $8$ ($7$) \\
    Candidates per slot $K$ & $8$ & $8$ \\
    Head layers, width, heads, FFN & $2$, $1024$, $8$, $2048$ & $2$, $1024$, $8$, $2048$ \\
    Cosine scale $\sigma$ & $8$ & $8$ \\
    Body key-value heads & $8$ & $8$ \\
    Body convolution (kernel, group) & none & $2$, $16$ \\
    Body loss & cross-entropy & distillation, Eq.~\eqref{eq:lilicorr_kd} \\
    Position decay $\beta$ & $7$ & $4$ \\
    Head loss & $0.25\,\mathcal{L}_{\mathrm{CE}} + 0.25\,\mathcal{L}_{\mathrm{dist}}$ & $0.05\,\mathcal{L}_{\mathrm{CE}} + 0.45\,\mathcal{L}_{\mathrm{cal}}$ \\
    Optimizer & AdamW & AdamW \\
    Learning rate & $6\times10^{-4}$ & $8\times10^{-4}$ \\
    Schedule & linear, $4\%$ warmup & constant, $1{,}000$-step warmup \\
    Weight decay, gradient clipping & $0$, $1.0$ & $0$, $1.0$ \\
    Global batch, sequence length & $64$, $3{,}120$ & $64$, $4{,}096$ / $8{,}192$ (stage 1 / 2) \\
    Blocks sampled per sequence & $512$ & $512$ \\
    Training epochs (stage 1 $+$ stage 2) & $6$ & $3$ $+$ $2.9$ ($500$K steps) \\
    Precision & BF16, FP32 master weights & BF16, FP32 master weights \\
    Hardware & $64\times$ H100 & $96\times$ GB200 ($64$ train, $32$ target) \\
    \bottomrule
  \end{tabular}
  \caption{Training configuration of LiLiCorr for Nemotron 3.5 Super, compared with the original work.}
  \label{tab:lilicorr_hparams}
\end{table}

\paragraph{Evaluation}
\label{sec:lilicorr_eval}

We evaluate on the qualitative split of SPEED-Bench~\citep{2026speedbench}, which comprises $880$ prompts, $80$ in each of $11$ categories. The drafter is served with the LiLiCorr integration of vLLM~\citep{kwon2023efficient}\footnote{\url{https://github.com/vllm-project/vllm/pull/57934}} with a draft length of $7$ and a concurrency of $64$. We report the per-sample acceptance length (AL), i.e., the number of tokens accepted per verification step, including the token contributed by the target, averaged over each request and then over requests. $temp=0$ is greedy decoding, in which the drafter commits the walk of Eq.~\eqref{eq:lilicorr_walk}. $temp=1$, with top-$p$ of $0.95$, uses probabilistic draft sampling from Eq.~\eqref{eq:lilicorr_cond}. For both settings we report the mean and standard deviation over three seeds. We use NVFP4 version of Nemotron 3.5 Super model for verification since it is most useful for inference and hence in combination with the drafter.

Because the drafter is trained on windows of at most $8{,}192$ tokens, its attention does not extrapolate to longer contexts, and without adaptation its acceptance degrades sharply with input length. We therefore serve it with sliding-window attention over the most recent $8{,}192$ positions in every drafter layer, so that query-key distances never exceed those seen in training. 
On inputs shorter than the window this is identical to full attention. Since nearly all SPEED-Bench qualitative samples fit within the window, the numbers in Table~\ref{tab:lilicorr_eval} are effectively those of full attention.





\begin{table}[t]
    \centering
    \small
    \begin{tabular}{lcccc}
      \toprule
      & \multicolumn{2}{c}{BF16 drafter} & \multicolumn{2}{c}{NVFP4 drafter} \\
      \cmidrule(lr){2-3} \cmidrule(lr){4-5}
      Category & $temp=0$ & $temp=1$ & $temp=0$ & $temp=1$ \\
      \midrule
      coding        & $5.70 \pm 0.01$ & $5.31 \pm 0.08$ & $5.65 \pm 0.03$ & $5.29 \pm 0.04$ \\
      rag           & $5.67 \pm 0.01$ & $5.29 \pm 0.03$ & $5.69 \pm 0.05$ & $5.33 \pm 0.01$ \\
      multilingual  & $5.56 \pm 0.04$ & $5.13 \pm 0.12$ & $5.53 \pm 0.02$ & $5.04 \pm 0.02$ \\
      summarization & $5.38 \pm 0.01$ & $4.96 \pm 0.07$ & $5.37 \pm 0.01$ & $4.96 \pm 0.02$ \\
      math          & $4.94 \pm 0.05$ & $4.54 \pm 0.14$ & $4.93 \pm 0.02$ & $4.56 \pm 0.09$ \\
      reasoning     & $4.83 \pm 0.00$ & $4.49 \pm 0.08$ & $4.87 \pm 0.01$ & $4.54 \pm 0.02$ \\
      qa            & $4.38 \pm 0.02$ & $4.01 \pm 0.04$ & $4.37 \pm 0.03$ & $4.03 \pm 0.03$ \\
      stem          & $4.15 \pm 0.02$ & $3.71 \pm 0.08$ & $4.11 \pm 0.03$ & $3.69 \pm 0.06$ \\
      humanities    & $3.99 \pm 0.04$ & $3.54 \pm 0.08$ & $3.94 \pm 0.01$ & $3.51 \pm 0.06$ \\
      writing       & $3.99 \pm 0.03$ & $3.57 \pm 0.12$ & $3.92 \pm 0.01$ & $3.58 \pm 0.02$ \\
      roleplay      & $3.87 \pm 0.02$ & $3.50 \pm 0.03$ & $3.78 \pm 0.01$ & $3.46 \pm 0.02$ \\
      \midrule
      Overall       & $4.77 \pm 0.01$ & $4.37 \pm 0.06$ & $4.74 \pm 0.01$ & $4.36 \pm 0.01$ \\
      \bottomrule
    \end{tabular}
    \caption{Average acceptance length (AL) of LiLiCorr on the SPEED-Bench qualitative split, at draft length $7$, served in vLLM at concurrency $64$ with $8{,}192$-token sliding-window attention, verified by the NVFP4 Nemotron 3.5 Super. $temp=0$ is greedy decoding, and $temp=1$ uses probabilistic draft sampling with top-$p$ of $0.95$; both are reported as the mean and standard deviation over three seeds.}
    \label{tab:lilicorr_eval}
\end{table}


\section{Conclusion}
\label{sec:conclusion}
We presented Nemotron 3.5 Super and Nemotron 3.5 Lightning, efficient Mixture-of-Experts Hybrid Mamba-Transformer models designed for reasoning, long-context understanding, and agentic workloads. Through large-scale distillation, long-context training, and native Multi-Token Prediction (MTP), both models combine strong capabilities with efficient inference. Nemotron 3.5 Super further incorporates Indic continued pretraining and agentic software-engineering mid-training, while a comprehensive post-training pipeline spanning supervised fine-tuning, reinforcement learning, specialized teacher training, and multi-teacher on-policy distillation transforms it into a multimodal agent. Nemotron 3.5 Lightning follows a lighter text-only post-training pipeline optimized for efficient deployment.

Our results demonstrate improvements across capability and efficiency. Indic continued pretraining improves Indian-language benchmarks by up to 9.82 percentage points, while agentic mid-training strengthens long-horizon software engineering and robustness across agent harnesses. The final Super model achieves 51.78\% on SWE-bench Pro, outperforming Nemotron 3 Ultra despite using approximately 4.6$\times$ fewer active parameters. Test-time scaling through parallel agent execution and self-verification further improves Terminal-Bench 2.1 from 62.7\% to 71.9\%. On the efficiency side, post-training MTP boosting improves speculative-decoding acceptance length by 10.1\% under greedy decoding and 21.8\% under probabilistic draft sampling, while NVFP4 quantization achieves 99.52\% median accuracy recovery relative to BF16. On GB200, Super achieves 1.9$\times$ the serving throughput of Qwen3.8 Flash-Next under the evaluated interactive agentic workload.

Together, these results demonstrate the benefits of jointly optimizing pretraining, post-training, and inference for capable and efficient agentic models. We release model checkpoints, datasets, and supporting recipes and cookbooks to enable further research and development of open agentic AI systems.

\section*{Contributors}

We thank the following people for their invaluable contributions to \ourmodelfull and Lightning.

Aastha Jhunjhunwala, Aayushi Ashokkumar Srivastava, Abhibha Gupta, Abhinav Garg, Abhinav Khattar, Achsah Wildish, Adam Rajfer, Adi Renduchintala, Adil Asif, Adriana Flores Miranda, Aileen Zaman, Aishwarya Chhabra, Akanksha Shukla, Akhiad Bercovich, Alejandra Rico, Aleksa Gordic, Aleksander Ficek, Alex Gronskiy, Alex Steiner, Alexander Bukharin, Alexandre Milesi, Ali Roshan Ghias, Ali Shehper, Ali Taghibakhshi, Alice Gatti, Alisa Liu, Alok Kumar, Alok Tripathy, Amala Sanjay Deshmukh, Aman Miryala, Aman Shanbhag, Ameya Sunil Mahabaleshwarkar, Ameya Velingker, Amir Klein, Amir Samani, Amit Zuker, Amnon Geifman, Anahita Bhiwandiwalla, Ananth Subramaniam, Andre Manoel, Andrea Santilli, Andrew Tao, Andrii Skliar, Anil Thomas, Anish Mahishi, Anna Shors, Anna Warno, Anuj Doshi, Anusha Kamath, Anwith Kiran, Ao Tang, Arham Mehta, Arnab Dhara, Arnav Komaragiri, Arnav Kundu, Arti Jain, Arun Rao, Arun Venkatesan, Arushi Goel, Asha Anoosheh, Ashish Vaswani, Ashwath Aithal, Asif Ahamed, Asma Kuriparambil Thekkumpate, Aswin Visva, Atefeh Sohrabizadeh, Avinash Vem, Ayush Dattagupta, Benedikt Schifferer, Benjamin Chislett, Bernard Nguyen, Besmira Nushi, Bilal Kartal, Bita Darvish Rouhani, Bobby Chen, Boris Ginsburg, Boxin Wang, Brandon Norick, Branislav Kisacanin, Brendan Johnson, Brian Dang, Brian Yu, Bryan Catanzaro, Byung-Kwan Lee, Carlo del Mundo, Chankyu Lee, Chantal Hwang, Charles Wang, Charlie Truong, Chen Cui, Cheng Wang, Cheng-Ping Hsieh, Chenhan Yu, Chenjie Luo, Chintan Patel, Chris Alexiuk, Chris Johnson, Chris Wing, Christian Munley, Coco Li, Collin McCarthy, Cynthia Gao, Dan Su, Dan Zhao, Dane Corneil, Danial Mohseni Taheri, Daniel Afrimi, Daniel Fatade, Daniel Lo, Daniel Machlab, Daniel Serebrenik, Daniele Molinari, Daniil Sorokin, Daria Gitman, Daria Levy, David Mosallanezhad, David Yu, Deepak Narayanan, Dev Churiwala, Devin O'Kelly, Dhruv Nathawani, Di Allan Wu, Dima Rekesh, Dina Yared, Dipendra Misra, Divyanshu Kakwani, Dmitry Belenko, Dominic Manno, Dongyang Fan, Dor Tsur, Duncan Riach, Dustin VanStee, Dylan Lu, Edgar Minasyan, Ehsan Hosseini-Asl, Eileen Long, Elena Lantz, Ellie Evans, Elliott Ning, Elvis Chen, Emanuel Scoullos, Eric Chung, Eric Harper, Eric Pham-Hung, Eric Tramel, Eric Yang, Erik Arakelyan, Evelina Bakhturina, Ewa Dobrowolska, Fadi Saady, Farzan Memarian, Fay Wang, Fei Jia, Felipe Soares, Felipe Vieira Frujeri, Feng Chen, Fengguang Lin, Ferenc Galko, Frankie Siino, Frederic Lafrance, Fuxiao Liu, Gantavya Bhatt, Garvit Kulshreshtha, George Armstrong, Giulio Lovisotto, Gomathy Venkata Krishnan, Gosia Lazuka, Grace Lam, Greg Heinrich, Greg Pauloski, Grzegorz Chlebus, Guilin Liu, Guo Chen, Guoming Zhang, Guyue Huang, Haifeng Qian, Hanrong Ye, Hao Gao, Hao Zhang, Haoqin Tu, Haran Kumar Shiv Kumar, Haribhau Hud, Harris Nover, Harrison Saturley-Hall, Harshil Kotamreddy, Hemil Desai, Henry Estela, Hexin Wang, Himanshu Singh, Hui Wang, Hung-Yueh Chiang, Igor Gitman, Ilia Karmanov, Ilya Loshchilov, Ingo Guehring, Itay Neeman, Ivan Moshkov, Jacky Kwok, Jakub Slowikowski, James Shen, Jane Polak Scowcroft, Jared Casper, Jatin Mitra, Jeff Willette, Jenna Diamond, Jenny Chen, Jesse Oliver, Jessica Medina, Jiacheng Xu, Jialin Song, Jian Hu, Jian Tu, Jian Zhang, Jiantao Jiao, Jiaqi Zeng, Jie-Fang Zhang, Jimmy Zhang, Jin Xu, Jingyu Xin, Jinhang Choi, Joey Conway, Joey Guman, John Kamalu, John Liu, John Roberts, Johnny Greco, Jonah Alben, Jonas Yang, Jonathan Cohen, Jonathan Li, Jonathan Raiman, Joseph Jennings, Juhi Mittal, Julia Gusak, Julien Veron Vialard, June Yang, Junkeun Yi, Jupinder Parmar, Ka Shun Shum, Kaiwen Shi, Kajal Jain, Kanishk Singla, Karan Sapra, Kari Briski, Kateryna Chumachenko, Katherine Huang, Katie Lyons, Keith Wyss, Keshav Santhanam, Keval Morabia, Kevin Shih, Kezhi Kong, Khanh Nguyen, Khushi Bhardwaj, Kiran Praveen, Konstantinos Krommydas, Krishna Puvvada, Krzysztof Pawelec, Kushal Shah, Li Ding, Lily Lee, Linda Liu, Linglin Jing, Luis Vega, Maarten Van Segbroeck, Mahan Fathi, Mahdi Nazemi, Maka Dong, Makesh Narsimhan Sreedhar, Maor Ashkenazi, Marc Cuevas, Marcin Chochowski, Marco Pavone, Markus Kliegl, Marta Stepniewska-Dziubinska, Martina Gonzalez Vilas, Martyna Patelka, Matan Rusanovsky, Matthieu Le, Matty Machczynski, Matvei Novikov, Mauricio Ferrato, Meenakshi Sharma, Mehrzad Samadi, Melissa Corpuz, Melissa Saelzer, Meng Xin, Mengru Wang, Mengxi Wu, Meriem Boubdir, Michael Boone, Michael Callahan, Michael Elad, Michael Lightstone, Michal Bien, Michal Zawalski, Miguel Martinez, Mikail Khona, Mike Chrzanowski, Mike Ranzinger, Mike Wojcikiewicz, Min-Hung Chen, Ming Kang, Mingjie Liu, Mingyuan Ma, Misha Nekorystnov, MJ Mikulski, Mohammad Qazim Bhat, Mohammad Shoeybi, Morgan Purcell, Mostofa Patwary, Nabin Mulepati, Nancy Agarwal, Narimane Hennouni, Nathan Sala, Nave Assaf, Nayeon Lee, Neal Vaidya, Negar Habibi, Netanel Haber, Nick Gupta, Nick Reamaroon, Nickson Quak, Nidhi Bhatia, Nik Spirin, Niket Agarwal, Nikolai Ludwig, Nima Tajbakhsh, Nithin Sunny, Oleksii Hrinchuk, Oleksii Kuchaiev, Olivier Delalleau, Omer Ullman Argov, Pablo Ribalta, Padmavathy Subramanian, Parth Mannan, Pavlo Molchanov, Pawel Morkisz, Peter Jin, Petr Lapukhov, Petro Junior Milan, Philip Monk, Pinky Xu, Piotr Januszewski, Piotr Laszkiewicz, Po-Han Huang, Praateek Mahajan, Pradeep Thalasta, Prakash Gurumurthy, Pranav Prashant Thombre, Prasoon Varshney, Pratyusha Maiti, Preeti Chadha, Pritam Biswas, Pritam Gundecha, Przemek Strzelczyk, Puhui Meng, Pulkit Kumar, Rabeeh Karimi Mahabadi, Rachel Oberman, Rajpreet Thethy, Ramani Akkati, Rameshwar Shivbhakta, Ran Zilberstein, Randy Gelhausen, Rasoul Shafipour, Raviraj Joshi, Ray Macalisang, Rayen Tian, Reka Kovacs, Renjie Pi, Resha Shetty, Ria C, Rima Shahbazyan, Rishabh Garg, Rita Fernandes Neves, Ritika Borkar, Ritu Gala, Riya Talwar, Riyad Islam, Robert Clark, Robert Kirby, Roger Waleffe, Rohan Joshi, Rohan Pillai, Rohit Jena, Rohit Watve, Roi Garnett, Roi Koren, Rongguang Ye, Russell J. Hewett, Ryan Prenger, Ryan Stewart, Ryota Egashira, Sadegh Mahdavi, Sai Nishanth Padala, Salika Dave, Sandhya Rani Narravula, Sandip Bhaskar, Sangkug Lym, Sanjeev Satheesh, Sarah Amirslani, Saran Vikas Murari, Saurabh Mishra, Saurabh Srivastava, Saurav Muralidharan, Sean Narenthiran, Sean Wang, Seokju Cho, Seong Hee Lee, Seonjin Na, Seonmyeong Bak, Seph Mard, Serge Panev, Sergei Kolchenko, Sergio Perez, Seth Henneman, Seth Poulos, Shahar Mor, Shantanu Acharya, Shaokun Zhang, Shaona Ghosh, Sharath Turuvekere Sreenivas, Shaun Kotek, Shaya Gharghabi, Sheng-Chieh Lin, Shengliang Xu, Shi Chen, Shihao Wang, Shilpa Amireddy, Shiqing Fan, Shiyang Chen, Shrimai Prabhumoye, Shriram Venugopal, Shuang Yu, Shubham Gandhi, Shubham Pachori, Shubham Toshniwal, Shuoyang Ding, Shwetha Krishnamurthy, Simeng Sun, Sivakumar Arayandi Thottakara, Somshubra Majumdar, Soumye Singhal, Sridurga Krithivasan, Sriharsha Niverty, Stephen Ge, Sugam Dipak Devare, Sukrit Rao, Sumeet Kumar Barua, Sungsoo Ha, Surya Nistala, Suseella Panguluri, Sweta Priyadarshi, Syeda Nahida Akter, Talor Abramovich, Tanay Varshney, Tandeep Singh Chandehoke, Teo Ene, Terry Kong, Thanh Do, Thinh Le, Tian Zhang, Tijmen Blankevoort, Tim Romanski, Tiyasa Mitra, Tom Balough, Tom Huang, Tomasz Grzegorzek, Tomer Asida, Tomer Bar Natan, Traian Rebedea, Trenton Starkey, Tugrul Konuk, Tuomas Rintamaki, Twinkle Vashishth, Tyler Kniess, Tyler Poon, Udi Karpas, Ushnish De, Utkarsh Vaidya, Vahid Noroozi, Vanshil Atul Shah, Venkat Srinivasan, Venkata Karri, Venmugil Elango, Vibhor Agrawal, Vibhu Jawa, Victor Cui, Victoria Edmonds, Vijay Korthikanti, Vikas Mehta, Vikranth Jeyakumar, Vinay Rao, Vineeth Kalluru Srinivas, Vitaly Kurin, Vitaly Lavrukhin, Wahidur Rahman, Wanli Jiang, Wasi Uddin Ahmad, Wataru Ishihara, Wei Du, Wei Ping, Wei-Ming Chen, Weiheng Chai, Weijia Chen, Wendy Quan, Wenliang Dai, Wenyan Cong, William Zhang, Wojciech Prazuch, Xiaowei Ren, Xin Yu, Xiwen Yu, Yan Wang, Yang Chen, Yang Yu, Yangyi Chen, Yaniv Galron, Yash Jain, Yashaswi Karnati, Ye Yu, Yejin Choi, Yemi Ibrahim, Yev Meyer, Yi Dong, Yi-Fu Wu, Yian Zhang, Yifan Zhang, Ying Lin, Yiyang Zhou, Yoav Miron, Yogendra Pandey, Yonatan Geifman, Yonggan Fu, Yoshi Suhara, Youngeun Kwon, Yu-Cheng Chou, Yu-Chiang Frank Wang, Yuanguo Kuang, Yuekai Zhang, Yugi Guvvala, Yuki Huang, Yunheng Zou, Yunsheng Liu, Zhenming Wang, Zhiding Yu, Zhiqi Li, Zhiyu Cheng, Zhongbo Zhu, Zhuolin Yang, Zichao Yue, Zihan Liu, Zijie Yan, Zirui Liu, Zoe Faul, Zuhair Ahmed

\newpage

\bibliography{references}
\bibliographystyle{references}

\appendix

\Needspace{0.8\textheight}

\clearpage

\section{Long-Translation}
\label{appendix:long-translation}

\begin{figure}[ht!]
    \centering
    \includegraphics[width=0.5\linewidth]{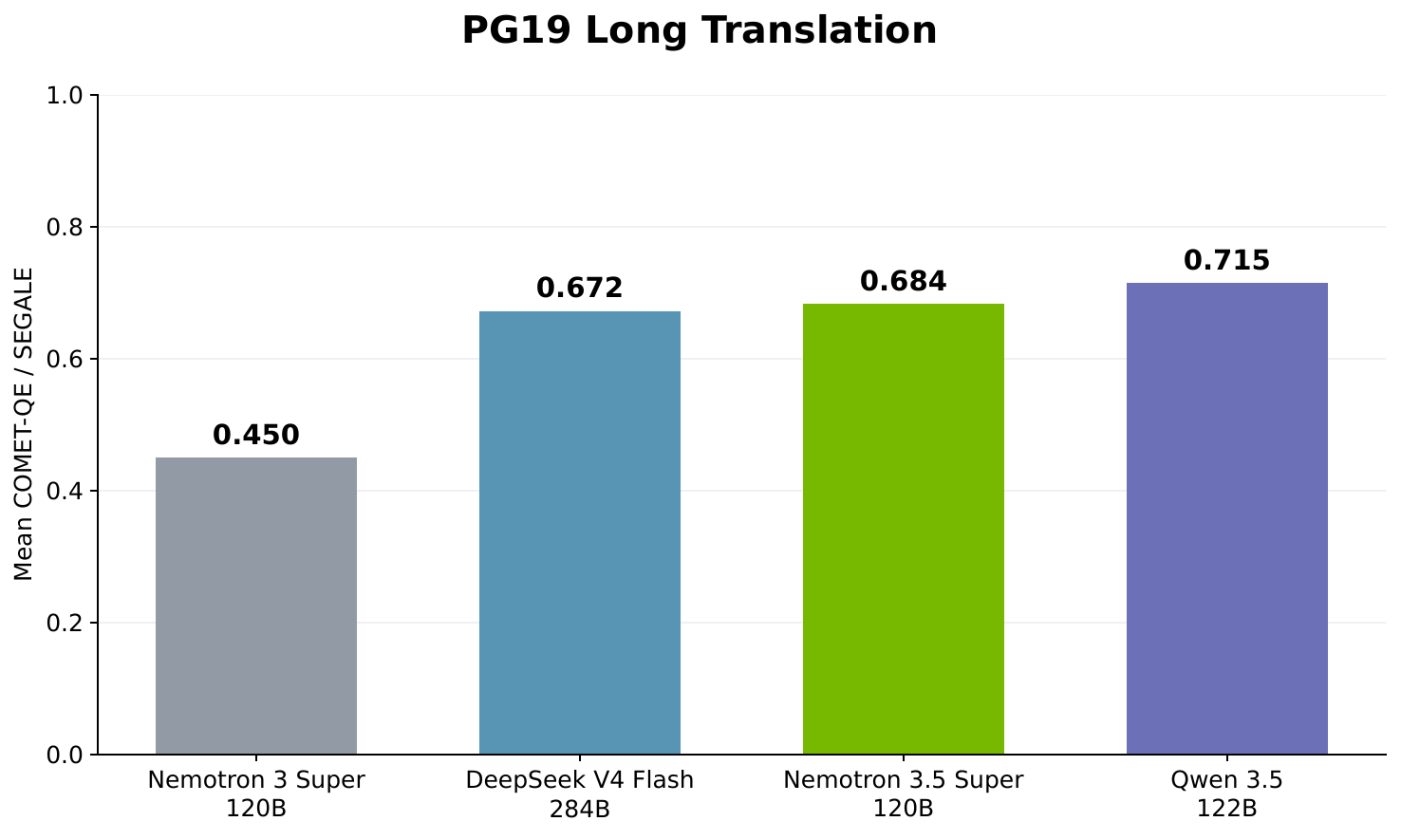}
    \caption{Book level translation scores on the PG19 test set containing 100 books. Models are instructed to translate each book into German, Spanish, French, Italian, Japanese, and Chinese, forming 600 total translations. We then score each translation using the SEGALE~\citep{segale} pipeline.}
    \label{fig:long-translation}
\end{figure}

\begin{table}[htbp]
\centering
\caption{Language and token distribution for PG19 long translation data.}
\label{tab:pg19-language-tokens}
\begin{tabular}{llrr}
\toprule
Side & Language & Records & Tokens \\
\midrule
Source & English              & 4,335 & 258,667,792 \\
\midrule
Target & Japanese             &   785 &  64,757,505 \\
Target & Chinese (Simplified) &   786 &  55,994,705 \\
Target & Italian              &   757 &  51,381,730 \\
Target & Spanish (Mexico)     &   766 &  49,770,267 \\
Target & German               &   728 &  48,193,156 \\
Target & French               &   513 &  29,178,759 \\
\midrule
\multicolumn{2}{r}{Target total} & 4,335 & 299,276,122 \\
\bottomrule
\end{tabular}
\end{table}

\begin{table}[htbp]
\centering
\caption{Language and token distribution for FinePDFs-Edu long translation data.}
\label{tab:finepdfs-edu-language-tokens}
\begin{tabular}{llrr}
\toprule
Side & Language & Records & Tokens \\
\midrule
Source & English (Latin)        & 1,031 & 451,571,611 \\
Source & Spanish (Latin)        &   173 &  74,851,307 \\
Source & French (Latin)         &   112 &  46,515,356 \\
Source & Mandarin Chinese (Han) &    85 &  31,270,147 \\
Source & Japanese               &    37 &  16,259,086 \\
\midrule
\multicolumn{2}{r}{Source total} & 1,438 & 620,467,507 \\
\midrule
Target & Italian              & 97 & 49,297,516 \\
Target & Vietnamese           & 83 & 38,620,351 \\
Target & Danish               & 71 & 38,607,034 \\
Target & Norwegian Bokm{\aa}l & 72 & 38,171,425 \\
Target & Portuguese (Brazil)  & 76 & 37,615,920 \\
Target & Indonesian           & 72 & 37,144,561 \\
Target & Japanese             & 78 & 36,626,127 \\
Target & Finnish              & 69 & 35,973,396 \\
Target & Portuguese           & 70 & 35,528,679 \\
Target & Korean               & 81 & 35,360,466 \\
Target & Dutch                & 66 & 35,240,892 \\
Target & French               & 69 & 35,112,951 \\
Target & German               & 66 & 34,144,385 \\
Target & Thai                 & 72 & 34,013,548 \\
Target & Spanish              & 65 & 32,888,083 \\
Target & Swedish              & 60 & 31,457,768 \\
Target & Turkish              & 62 & 31,304,928 \\
Target & Hindi                & 75 & 28,435,582 \\
Target & Chinese (Simplified) & 55 & 26,234,775 \\
Target & Arabic               & 62 & 22,652,904 \\
Target & English              & 17 &  6,900,594 \\
\midrule
\multicolumn{2}{r}{Target total} & 1,438 & 701,331,885 \\
\bottomrule
\end{tabular}
\end{table}

\clearpage

\section{Ablation study of post-training quantization}
\label{app:quant:ablation}

\subsection{Runaway loop: the termination failure in truncated rollouts}

Reasoning length is not a defect; long reasoning in an agentic workload is mostly probing such as running code, invoking tools, and reading back the result.
A model's abstention can fail by two mechanisms: \emph{knowledge} collapse, which produces hallucination, and \emph{reasoning} collapse,
which produces termination failure, seen as semantic runaway looping behavior. Qualitatively, the runaway loop is recognizable as
\emph{failed negative proof}: the model cannot certify a negative result from exhaustive investigation, but confidently reopens a settled conclusion and cycles.
Therefore, a non-terminating agent emits more calls of ordinary length, with per-call statistics steady, rather than longer calls.
Eventually, the rollout's last reasoning starts exploding in length, or the truncation rate increases.

We identify the runaway loop by reading reasoning passages in postmortem, while tracking the failure of a model to terminate as it runs to the context wall.
Table~\ref{tab:attn:verbosity} reports the two immediately measurable quantities: \textbf{truncation}, the fraction of \emph{rollouts} containing at least
one model call hitting an output limit, and \textbf{output length per rollout}. The combination of these two quantities is a useful early-warning indicator.
For example, tau3-banking has 20.9\% truncation and 35,173 output tokens per rollout before MOPD, where we detect 19.7\% runaway loop failures (153/776 rollouts).
After MOPD, truncation decreases to 0.3\% with 13,669 output tokens per rollout, where only one runaway loop failure is captured.

\begin{table*}[t]
  \centering
  \caption{Output length per rollout and truncation, pre-MOPD versus MOPD, both
  in BF16. OSL is the total over a rollout's \emph{policy-model} calls, excluding
  the user simulator. Post-training RL removed the non-terminating tail, not baseline
  verbosity.}
  \label{tab:attn:verbosity}
  \begin{tabular}{llrrll}
    \toprule
    Benchmark & Modality & Rollouts & ISL (k) & OSL (k) / rollout & Trunc.\ \% \\
    \midrule
    Baby Vision       & vision         & $388 \times 3$     & 0.5   & 42.3 $\rightarrow$ 11.2 (\textbf{0.27}$\times$) & 39.0 $\rightarrow$ 1.9 \\
    CharXiv (RQ)      & vision         & $1{,}000 \times 3$ & 3.1   & 11.8 $\rightarrow$ 3.3 (\textbf{0.28}$\times$)  & 9.2 $\rightarrow$ 0.6 \\
    MMLongBench-Doc   & vision         & $1{,}091 \times 3$ & 75.0  & 10.8 $\rightarrow$ 3.0 (\textbf{0.28}$\times$)  & 9.9 $\rightarrow$ 0.4 \\
    OCR Reasoning     & vision         & $1{,}069 \times 3$ & 1.5--1.6\,\textsuperscript{a} & 16.8 $\rightarrow$ 5.2 (\textbf{0.31}$\times$) & 13.7 $\rightarrow$ 0.6 \\
    OSWorld           & vision/agentic & $361 \times 3$     & ---   & 13.6 $\rightarrow$ 5.1 (\textbf{0.38}$\times$)  & 1.2 $\rightarrow$ 0.0 \\
    Tau3 Banking      & text/agentic   & $97 \times 8$      & ---   & 35.2 $\rightarrow$ 13.7 (\textbf{0.39}$\times$) & 20.9 $\rightarrow$ 0.3 \\
    AA-Omniscience    & text           & $600 \times 10$    & 0.1   & 10.0 $\rightarrow$ 5.6 (0.56$\times$)           & 1.9 $\rightarrow$ 0.3 \\
    SciCode           & text           & $65 \times 8$      & 3.0--3.1\,\textsuperscript{a} & 7.2 $\rightarrow$ 6.5 (0.90$\times$)            & 0.0 $\rightarrow$ 0.0 \\
    \midrule
    \textbf{AA-LCR}   & text           & $100 \times 16$    & 108.0 & 4.7 $\rightarrow$ 4.7 (\textbf{1.00}$\times$)   & \textbf{0.0} $\rightarrow$ 0.2 \\
    \textbf{Tau2-AA}  & text/agentic   & $278 \times 8$     & ---   & 5.0 $\rightarrow$ 5.1 (\textbf{1.03}$\times$)   & \textbf{0.0} $\rightarrow$ 0.0 \\
    \bottomrule
  \end{tabular}

  \vspace{2pt}
  {\footnotesize\raggedright \textsuperscript{a}\,evaluation harness update resulted in ISL difference.\par}
\end{table*}

\subsection{FP4 weight quantization swings the stop-versus-continue margin}
\label{app:quant:ablation:fp4experts}

Quantized models inherit the runaway loop behavior from the base model, and quantization can make it worse.
We divide and conquer the degradation by ablating non-attention components in pre-MOPD models:
\begin{itemize}
  \item \textbf{Routed/shared experts NVFP4 $\rightarrow$ FP8.} In tau3-banking, runaway dropped from 28.2\%
    $\rightarrow$ 17.9\%, below the BF16 reference's 19.7\%; output verbosity decreased by
    21.5\% at 2.7$\sigma$. This rules out expert weights being innocent.
  \item \textbf{Expert \emph{activations} $\rightarrow$ BF16.} Runaway moves by
    $\approx 0$; output verbosity falls on 5 of 5 focused benchmarks (mmlongbench-doc, scicode, tau3-banking, tau2-AA, AA-LCR).
    This eliminates the activation path as the carrier of the termination failure.
  \item \textbf{\texttt{lm\_head} returned to BF16.} No recovery of termination is captured.
    This eliminates the final output token generation as the source of the runaway loop.
  \item \textbf{W4A4 expert GEMM output error.} The relative L2 norm error is estimated to be 7.5--22.9\% against BF16,
    which is above 10\% in 39 of 40 MoE layers, and 15\% on average.
    This eliminates ``the small perturbation can be ignored''.
  \item \textbf{Mamba output projection $\rightarrow$ W4A16.} In tau3-banking, output verbosity increased by 18.5\% at 2.4$\sigma$,
    while the runaway rate and knowledge abstention were both unaffected.
\end{itemize}
By putting all the pieces together, we hypothesize the culprit is the quantization of the expert weights. Promoting expert weights to
FP8 mitigates the runaway loop on pre-MOPD models, while post-MOPD model quantization successfully recovers the verbosity without runaway
loop regression in FP4 expert weights. The Mamba result carries over to the post-MOPD checkpoints: against a quantization recipe keeping
output projection at W8A8, W4A16 is more verbose on all 6 benchmark tests. Please refer to Appendix~\ref{app:quant:ablation:mamba}.
Note that a perturbation from FP4 expert weights does not destroy information, so knowledge abstention is not affected across all experiments.
AA-OmniScience non-hallucination spans only $0.991\times$ to $1.022\times$ across the five ablations, which corresponds to $67.8\%$ and $69.9\%$
against BF16's $68.4\%$.

\paragraph{Where in the stack: the decision rides on a weak signal.} Table~\ref{tab:attn:abg} depicts a decomposition of each MoE layer's
output into its routed and shared experts, with $\alpha = \|\text{routed}\|/\|\text{layer}\|$ and $\beta = \|\text{shared}\|/\|\text{layer}\|$.
This magnitude analysis reveals a crossover at layer 59: the routed expert's contribution roughly halves while the shared one rises, and
the shared experts dominate nearly every layer from 59 to 83. Interestingly, when we split rollouts by whether BF16 and the FP4 quantized expert
weights produce the runaway loop effect from pre-MOPD models, we see a clear trend of relatively weaker routed expert contribution in the
late band, if a runaway loop occurs. This effect appears in both BF16 and FP4 quantization. For example, on rollouts where only BF16 experts
loop, the routed expert contribution is $0.055$ under BF16 but $0.093$ under the FP4 quantized experts, whereas on rollouts where only the
FP4 quantized experts loop, the routed expert contribution is $0.091$ under BF16 but $0.061$ under FP4. It is worth noting that FP4 weight
quantization error is uniform: FP4 expert weights carry a relative error of $0.087$ against BF16, in contrast to FP8 expert quantization's
$0.027$ across all layers. The weak routed contribution is a signature of the looping state. Under low-amplitude signals, FP4 quantization
flips are margin-confined: they occur where many continuations are interchangeable, and the flip redirects output generations without changing
their knowledge. The stop-versus-continue decision itself is far from an ordinary continuation, but a small shift in the probability of
stopping at ordinary positions attenuates termination across a long generation. That is also why accuracy stays flat while verbosity moves
on those five benchmarks.

\begin{table}[t]
  \centering
  \caption{Routed ($\alpha$) and shared ($\beta$) expert contributions in 12 MoE layers of L59--L83, split by whether each tau3-banking
  prompt produced a runaway loop under pre-MOPD BF16 rollout versus pre-MOPD FP4 rollout. MOPD rollouts reported 1 runaway loop failure
  in BF16, and 3 failures in FP4. All MoE layers before L59 agree to within $0.031$ regardless of pre-MOPD versus MOPD, and BF16 versus
  FP4 quantization.}
  \label{tab:attn:abg}
  \begin{tabular}{cccccc}
    \toprule
    \multicolumn{2}{c}{pre-MOPD Runaway?} & \multicolumn{4}{c}{\textbf{L59--L83 (12)}\quad $\alpha/\beta$ under} \\
    \cmidrule(lr){1-2}\cmidrule(lr){3-6}
    \makebox[4em][c]{BF16} & \makebox[4em][c]{FP4} & pre-MOPD BF16 & pre-MOPD FP4 & MOPD BF16 & MOPD FP4 \\
    \midrule
    No           & No           & 0.100/0.137                   & 0.097/0.146                   & 0.100/0.142          & 0.097/0.140          \\
    No           & \textbf{Yes} & 0.091/0.144                   & \textbf{0.061}/\textbf{0.162} & 0.092/0.149          & 0.092/0.147          \\
    \textbf{Yes} & No           & \textbf{0.055}/\textbf{0.161} & 0.093/0.139                   & \textbf{0.081}/0.164 & \textbf{0.081}/0.161 \\
    \textbf{Yes} & \textbf{Yes} & \textbf{0.059}/\textbf{0.151} & \textbf{0.063}/\textbf{0.155} & \textbf{0.077}/0.160 & \textbf{0.078}/0.155 \\
    \bottomrule
  \end{tabular}
\end{table}

On the other hand, MOPD fortifies the latent support for stopping enough to withstand FP4 quantization noise. Under MOPD, which loops on only
1 of 776 of these rollouts, the same prompts recover to $0.081$ and $0.077$ while $\beta$ stays at the elevated $0.16$, and FP4 expert weight
quantization leaves them at $0.081$ and $0.078$, where runaway loop failures are reported in 3 cases only. Table~\ref{tab:attn:stopmargin} shows that
reading the unembedding's decision directly confirms this: at the position where \texttt{</think>} is actually emitted, $P(\text{stop})$ is $0.99$
against $10^{-9}$ or less elsewhere, and FP4 expert weights never cost a termination. FP4 experts affect the \emph{direction} of the perturbation.
It moves away from stopping on 17 of 23 pre-MOPD rollouts, but toward stopping on 20 of 24 under MOPD. On the runaway rollouts themselves,
pre-MOPD's peak stop probability has median $0.043$ and the reasoning block never closes, while MOPD reaches median $0.729$ and closes the block
in 6 of 10. GEMM error could not capture this difference because it covers the magnitude only, not the distribution and bias.

\begin{table}[t]
  \centering
  \caption{The tau3-banking's stop-versus-continue decision, $P(\text{stop})$, at \texttt{lm\_head} output. Here \texttt{lm\_head}
  stays BF16 to isolate the opposite intervention to Table~\ref{tab:attn:lmhead}. The first three rows are measured on the
  rollouts where \texttt{</think>} was emitted. The last two rows are measured on the rollouts causing a runaway loop in pre-MOPD.
  Binomial two-sided direction test: $p = 2^{1-n}\sum_{i=k}^{n}\binom{n}{i}$, $k$ out of $n$ rollouts.}
  \label{tab:attn:stopmargin}
  \begin{tabular}{lcc}
    \toprule
                                                        & pre-MOPD                          & MOPD \\
    \midrule
    $P(\text{stop})$ at the stop vs. elsewhere (median) & 0.994 vs. $2.0\times10^{-9}$      & 0.996 vs. $6.5\times10^{-11}$ \\
    Terminations under FP4 experts                      & 23 / 23 rollouts                  & 24 / 24  rollouts \\
    \textbf{FP4 experts shift the decision}             & \textbf{17/23 away} ($p{=}0.035$) & \textbf{20/24 toward} ($p{=}0.002$) \\
    \midrule
    Peak $P(\text{stop})$, runaway rollouts (median)    & 0.043                             & 0.729 \\
    \texttt{</think>} at the peak                       & 2 / 10 rollouts\,\textsuperscript{a} & 6 / 10 rollouts \\
    \bottomrule
  \end{tabular}

  \vspace{2pt}
  {\footnotesize\raggedright \textsuperscript{a}\,\texttt{lm\_head} puts \texttt{</think>} on top, at $0.384$ and $0.871$, but sampling did not take it and the loop continued.\par}
\end{table}

\subsection{The damage of long-context retrieval on attention projections}
\label{app:quant:ablation:lc}

From the viewpoint of dense attention, long-context retrieval has three different perspectives on the decision:
\begin{itemize}
  \item \textbf{Distinction}: the target is identifiable by its content. The model must match on what a candidate says, wherever it sits.
  \item \textbf{Disambiguation}: the candidates are near-identical, so content cannot separate them. The model must count or index occurrences by ordinal position.
  \item \textbf{Aggregation}: there is no single target; evidence must be combined across many candidates.
\end{itemize}
Table~\ref{tab:attn:lc} maps the examples of long-context benchmarks onto these three and reports the result.
As noted, \textbf{quantization costs 2.65 to 3.6~pp of disambiguation and nothing measurable elsewhere.}
The same effect is also observed on Nemotron 3.5 Lightning. This section focuses on Super 3.5 VL to understand the mechanism.

\paragraph{Attention score dilution occurs in long-context retrieval.} Attention scores are resolution-starved before any projection error is added.
A softmax output lies in $0 \le P \le 1$, where a floating-point format mostly carries the magnitude in its exponent. As attention delivers $O=PV$, a logit perturbation is a \emph{bit shift} on the $v$ vector.
Table~\ref{tab:attn:dilution} describes measuring per head across 12 contexts spanning 10,532 to 63,641 prefix tokens, where 4,730 to 7,981 keys are co-located in the attention window. These keys compete for only 35 to 80 distinct e4m3 score levels, putting 60 to 120 keys on each level.
\begin{table}[t]
  \centering
  \caption{tau3-banking's score-ladder dilution under an FP8 KV cache, per head, across 12 contexts of 10,532--63,641 prefix tokens.
  The softmax output $P$ is cast to e4m3 for the $PV$ GEMM; the logits entering the softmax are not.}
  \label{tab:attn:dilution}
  \begin{tabular}{lcl}
    \toprule
    Quantity & Measured & Reading \\
    \midrule
    Keys in representable window  & 4,730--7,981     & population competing \\
    Distinct e4m3 $P$ levels      & 35--80           & levels available \\
    \textbf{Keys sharing a level} & \textbf{60--120} & the dilution \\
    Ties at the top e4m3 level    & 1--3             & the peak stays separated \\
    Attn. score mass preserved    & 0.994--1.001     & mass intact, resolution lost \\
    Spearman $\rho$(context len., keys/level) & $+0.73$--$+0.94$ & more keys per level when longer \\
    \bottomrule
  \end{tabular}
\end{table}
Suppose a distinction probe of GraphWalks, which reaches 361k tokens of observed context. From logit's exponential property,
most per-position candidates are flat in attention scores. So is document-scale aggregation. While attention looks for peak score, which is
cleanly separated from other positions, a long-context workload requiring distinction and aggregation works well with FP8 or NVFP4 KV cache precision.
Unfortunately, if candidates are near-identical by construction, so nothing but position separates them, the precision becomes its limitation
as demonstrated by MRCR-128k.

\paragraph{MRCR-128k stratification: damage is concentrated on specific context/candidate counts.} Even though the total attention mass is preserved,
per-level key concentration destroys the resolution in quantization. As shown in Table~\ref{tab:attn:lc}, the BF16-over-FP8 on $q/k/v$ is worth
$0.83$~pp, with a standard error of $0.71$~pp for MRCR-128k. This might be interpreted as an insignificant improvement as MXFP8 projection gains $+0.99$~pp
against the per-tensor FP8, which places it $0.16$~pp on the far side of BF16. Neither contrast clears $2\sigma$(=$1.4$~pp) under the standard variance.
Likewise, the last row of Table~\ref{tab:attn:dilution} might suggest damage grows smoothly with context length. It does not grow linearly.
Table~\ref{tab:attn:mrcrstrat} stratifies 1,470 MRCR-128k tasks at three repetitions to separate the retrieval from decoding noise. The ablations show the damage
is present at every length, but exacerbated at 32k+: $q/k/v$ FP8 retains $95.3\%$ and $95.5\%$ below 32k, against $91.4\%$ and $92.1\%$ above 32k, a $3.6$~pp
break. On the contrary, $q/k/v$ MXFP8 projection recovers $95.3\%$ and $94.3\%$ above 32k, similar to $q/k/v$ BF16. Substituting MXFP8 for per-tensor FP8 on
$q/k/v$ changes only the scale granularity with one E8M0 exponent per 32 elements instead of one float per tensor, which is enough to close the gap.
Crowding a level only costs accuracy when the co-resident keys are confusable, and unrelated context does not threaten a retrieval whose target has a wide margin. 
\begin{table}[t]
  \centering
  \caption{MRCR-128k accuracy recovery against the BF16 reference, by served context, over 1,470 tasks at three repetitions. Context bins use \texttt{n\_tokens}, which undercounts the served prompt by about $3\%$. The $o$ projection is FP8 in every recovery column.}
  \label{tab:attn:mrcrstrat}
  \begin{tabular}{lrcccc}
    \toprule
             &     & accuracy & \multicolumn{3}{c}{recovery \%} \\
    \cmidrule(lr){3-3}\cmidrule(lr){4-6}
    context  & $n$ & BF16 & q/k/v BF16 & q/k/v MXFP8 & q/k/v FP8 \\
    \midrule
    0--8k    & 290 & 0.768 & 97.9 & 97.5 & 95.3 \\
    8--32k   & 578 & 0.552 & 94.6 & 95.1 & 95.5 \\
    \midrule
    32--64k  & 295 & 0.559 & 95.0 & 95.3 & \textbf{91.4} \\
    64--131k & 307 & 0.596 & 93.8 & 94.3 & \textbf{92.1} \\
    \bottomrule
  \end{tabular}
\end{table}
On the other hand, Table~\ref{tab:attn:needles} describes the BF16 collapse from 0.800 to 0.378 as the candidate count rises from two to eight,
and that ordering holds inside every context band. Context also affects the damage, but its impact is not monotonically increasing: the baseline
accuracy steps down by $13.4$ to $29.5$~pp between 0--8k and 8--32k and then saturates, so above 32k the candidate count is the dominant factor.
This is the signature of an exhausted indexing budget rather than a strained context window.
\begin{table}[t]
  \centering
  \caption{MRCR-128k BF16 accuracy by number of near-identical candidates $\times$ served context, over 1,470 tasks at three repetitions. The two
  axes are independent by construction: each context band carries a $33\%$ mix of the three candidate counts, and the candidate strata are
  length-matched at $36.4$--$37.8$k mean prompt tokens. Recovery is $q/k/v/o$ FP8 against the BF16 reference.}
  \label{tab:attn:needles}
  \begin{tabular}{ccccccc}
    \toprule
               & \multicolumn{4}{c}{BF16 accuracy by served context} & \multicolumn{2}{c}{pooled} \\
    \cmidrule(lr){2-5}\cmidrule(lr){6-7}
    candidates & 0--8k & 8--32k & 32--64k & 64--131k & BF16 & recovery \% \\
    \midrule
    2          & 0.883 & 0.749 & 0.793 & 0.823 & 0.800 & 94.9 \\
    \textbf{4} & 0.806 & 0.585 & 0.522 & 0.694 & \textbf{0.637} & \textbf{91.7} \\
    8          & 0.616 & 0.320 & 0.353 & 0.290 & 0.378 & 96.0 \\
    \bottomrule
  \end{tabular}
\end{table}

In short, the damage is worst in the middle, smaller at two candidates where the margins are wide enough to absorb the error, and smaller again at eight where
the reference has already collapsed and little remains to lose. Similar to the unembedding in Appendix~\ref{app:quant:ablation:lmhead}, quantization moves outcomes
where the decision was already close. However, attention score is stronger evidence for a resolution mechanism: one that merely made everything harder would not
spare the hardest stratum. From this observation, reporting a single number hides the damage in long-context benchmarks. It is important to stratify test cases
along with \emph{capability partition} to assess quantization damage properly.

\paragraph{The attention score has FP8 resolution at per-block scale granularity.}
On Nemotron 3.5 Lightning model, when we move the KV cache from FP8 to BF16 while keeping weights and projections fixed, MRCR-128k's accuracy recovers $+1.43$~pp
(=$2.4\sigma$); while moving the projections from FP8 to BF16, it recovers $+2.97$~pp (=$2.8\sigma$). On Nemotron 3.5 Super VL model the impact of projection
precision appears from our ablations. While KV cache is the damage carrier, high-precision $q/k/v$ projection does not prevent dilution; it buys back part of
the margin destroyed by the cast. Table~\ref{tab:attn:lcr} varies attention projection precision on AA-LCR, an aggregation benchmark serving 108k tokens of context
on average. As noted, aggregation benchmarks can identify important candidates in FP8, where even per-tensor FP8 precision can be acceptable over 100 tasks at
sixteen repetitions. It also states accuracy drop of $2.50$~pp from all-FP8 to NVFP4 projections, with $\mathrm{SE} = 1.11$.
Across the $24$ $q/k/v$ weight tensors, the spread of per-16-element block maxima is $2.51$ binades at the median and $8.78$ at worst, and the activations
entering those projections span $1.67$, which is against the $17.81$ binades an E4M3 block scale encodes before it underflows. No block in the model loses
its scale, leaving $9.03$ binades unused on the worst tensor. Per-block granularity is not reclaiming dynamic range, but recovers precision headroom that a
per-tensor scale had exhausted. For example, under a fused $q/k/v$ FP8 projection with a single shared scale, the key projection holds it at six of eight attention
layers and the value projection at the other two. Fusion adds roughly 32\,\% relative weight error to $q$ and $v$ against 4\,\% for $k$, which is pure rounding
noise.
\begin{table}[t]
  \centering
  \caption{Ablation of attention projection precision on AA-LCR. This benchmark does not exercise disambiguation.}
  \label{tab:attn:lcr}
  \begin{tabular}{lcc}
    \toprule
    Attention $\{q,k,v,o\}$ precision & Score & $\Delta$ vs BF16 baseline \\
    \midrule
    All four at NVFP4                        & 0.5713 & $-2.25$ pp \\
    All four BF16                            & 0.5900 & $-0.38$ pp \\
    All four FP8                             & 0.5962 & $+0.25$ pp \\
    $o$ FP8, $q/k/v$ BF16                    & 0.6031 & $+0.94$ pp \\
    $q/k/v$ FP8, $o$ BF16                    & 0.6050 & $+1.12$ pp \\
    \bottomrule
  \end{tabular}
\end{table}

This study recommends FP8 precision on $q/k/v$ projection \emph{at per-block scale granularity}. If deployment does not perform disambiguation-style retrieval,
however, no other workload would reveal a significant difference. After the softmax has chosen the target $v$ vectors, the head outputs are passed to the output
projection. Note that the spread of output projection weights is 4.21 binades in median and 8.15 in the worst case across eight attention layers, where output
projection itself is not exposed to FP8 underflow. On the activation side, the calibrated per-tensor input scale is equally well-behaved: it sits at 49--76\,\%
of the largest head output norm at every attention layer, holding that fraction across a $68\times$ range of absolute magnitude. Counterintuitively, the
concatenation of head outputs under a single per-tensor input scale does not produce a benchmark-level penalty at FP8 precision. We claim that the output projection
is structurally exposed and measurably tolerant in per-tensor FP8.
In the meantime, it is worth noting that the mixed-precision checkpoints of \S\ref{sec:quant:autoquant}, produced by a gradient-based per-layer search, follow
these findings: the \emph{AutoQuant} search leaves $q/k/v$ in BF16 on three of the eight attention layers under both the 4.85 and 4.90 bit-per-parameter budgets,
on the same three layers in each, while $o$ stays FP8 on every attention layer under 4.85 and reaches BF16 on three layers only once the budget loosens to 4.90.

\subsection{Mamba output projection: FP4 weights, and the sensitivity is on generation length}
\label{app:quant:ablation:mamba}

The Mamba output projection differs from the expert \texttt{fc2} in the input \emph{shape} of the nonlinearity. Expert \texttt{fc2}
is fed by a squared-ReLU activation, where 90--99\,\% of its input elements are exactly zero before quantization, and only 1.05--3.32 values
survive per FP4 block. Mamba SSM cache passes a normalized SiLU gate to the output projection, which provides near-zero values but never
exactly zero. Consequently, the input of Mamba output projection carries near 16 survivors per FP4 block, and $27.6\%$ of elements fall
below the $\text{amax}_{\text{block}}/24$ rounding limit under FP4 activation quantization. This tensor analysis stops us from taking
the Mamba output projection to W4A4. Applying rotation or higher activation precision can mitigate the flush.

We evaluated the impact of FP4 weight quantization with W4A16 against FP8 W8A8. Due to high precision on activation, this contrast reveals
the weight quantization effect directly. Table~\ref{tab:attn:mambaosl} shows the output verbosity inflation caused by FP4 weight quantization.
All six benchmarks move in the same direction, with \texttt{OSWorld} and \texttt{SciCode} carrying the clearest inflation at $1.27\times$
and $1.16\times$. Note that each task accuracy is unchanged across two recipes. To address this issue, we preferred parameter alignment as described
in \S\ref{sec:quant:qad}.

\begin{table}[t]
  \centering
  \caption{output generation length per rollout for a quantization recipe pair differing in Mamba output projection only: W8A8 versus W4A16. Counts are policy-model output, excluding the user simulator on the agentic benchmarks.}
  \label{tab:attn:mambaosl}
  \begin{tabular}{lccc}
    \toprule
    Benchmark & FP8 (W8A8) & NVFP4 (W4A16) & ratio \\
    \midrule
    OSWorld          & 8.3k  & 10.6k & \textbf{$1.27\times$} \\
    SciCode          & 12.0k & 13.9k & \textbf{$1.16\times$} \\
    Tau3 Banking     & 14.3k & 15.5k & \textbf{$1.08\times$} \\
    AA-Omniscience   & 6.4k  & 6.7k  & $1.06\times$ \\
    AA-LCR           & 5.7k  & 5.9k  & $1.02\times$ \\
    Tau2-AA          & 5.2k  & 5.2k  & $1.01\times$ \\
    \bottomrule
  \end{tabular}
\end{table}

\subsection{The unembedding: FP8 weights, and the sensitivity is on termination}
\label{app:quant:ablation:lmhead}

The unembedding should stay at FP8 or above. As described in Table~\ref{tab:attn:lmhead}, NVFP4 \texttt{lm\_head} also holds high top-1 agreement with BF16.
FP4's problem is a failure of the stop-versus-continue decision, which results in output verbosity inflation.
\begin{table}[t]
  \centering
  \caption{Unembedding sensitivity, per-tensor FP8 versus NVFP4, measured on an identical hidden-state input with fake quantization.}
  \label{tab:attn:lmhead}
  \begin{tabular}{lccc}
    \toprule
    & FP8 (W8A16) & NVFP4 (W4A16) & Ratio \\
    \midrule
    Weight error                  & 0.0265 & 0.0952 & 3.6$\times$ \\
    Top-1 agreement with BF16     & 98.6\,\% & 95.8\,\% & --- \\
    KL mean / p99 (bits)          & 0.0013 / 0.0080 & 0.0121 / 0.0797 & 9.3$\times$ / 10$\times$ \\
    \textbf{Stop/continue flip rate} & \textbf{0.781\,\%} & \textbf{2.450\,\%} & \textbf{3.14$\times$} \\
    \bottomrule
  \end{tabular}
\end{table}
We evaluate the decision as a single logit comparison, $\text{margin} = h \cdot (v_{\text{stop}} - v_{\text{cont}})$. As \texttt{</think>} is one token,
everything shared by the two unembedding rows cancels out. After quantization, the \texttt{</think>} row provides a cosine similarity of 0.9961 to the
original. The vector barely moves, however 2.45\,\% of decisions flip due to the other vectors providing $\cos \approx 0.22$. This is because the decision
lives in the \emph{difference} of two nearly-orthogonal rows, where a small rotation of each becomes a large perturbation of their difference. ``The weights
look fine'' and ``the decision flipped'' are both true. We rejected a hypothesis of quantization damage from frequency-skewed rare-token distribution in four
ways:
\begin{enumerate}
  \item Row-amax spread is 2.84 binades (2.12 binades at p99.9) against FP8's roughly 17.8 usable, so there is nothing to reclaim on per-block scale.
  \item Tested error ranging from group size of 1 through 64 elements, and the pooled error is identical to four decimal places per each format, i.e., 0.0265 in FP8, and 0.0952 in FP4.
  \item Per-row median spread by vocabulary-frequency quartile is 5.222 / 5.246 / 5.252 / 5.274 binades. In other words, a 0.05-binade drift across 131,072 tokens. There are zero dead rows in every configuration, and the lowest-norm decile of relative quantization error is no worse than the highest: 0.0941 versus 0.0952. The error is uniformly distributed across the vocabulary.
  \item NVFP4 crosses its flush line on 66.9\,\% of blocks, flushing 6.8\,\% of elements, but this flush carries no vocabulary structure at any density level. Whatever it costs, it costs everywhere, so FP4 quantization cannot produce a loss concentrated in rare tokens.
\end{enumerate}
The remaining hypothesis is per-element \emph{grid}: E4M3 carries three mantissa bits providing a resolution of 12.5\,\% step size ($1 \rightarrow 1.125$),
whereas E2M1 carries one, providing a resolution of 50\,\% step size ($1 \rightarrow 1.5$). To support this hypothesis, we emulated two-pass FP4 weights,
where a second channel corrects the residual left below the $\text{amax}/24$ flush line. This correction recovers the gap entirely: with a linear E2M1 residual
the KL falls to $0.00117$ against per-tensor FP8's $0.00134$, and top-1 agreement with BF16 reaches $0.9951$. Reducing the quantization error in \texttt{lm\_head} is still open to exploration, but this suggests that the unembedding's numerical sensitivity is confined in FP8 precision.

\subsection{A precision recommendation with limitations}
\label{app:quant:ablation:reco}

The same quantization recipe, which aggravated the termination failure before RL post-training, reduced it afterwards. However, it does not modify model's inherent structural effect. Across three checkpoints spanning pre-RL and post-RL training, the stop-versus-continue geometry is invariant to within 1\,\% (cosine 0.997, margin 0.996, noise-to-margin 1.001), while termination behavior changed significantly. Together with the contribution profile, which is invariant to within 0.007 in routed experts, this gives a clean statement: \textbf{RL post-training did not move the \texttt{lm\_head} or the expert contribution structure; it changed where
the hidden state points.} A precision recommendation can therefore be stated in structural terms, but its \emph{margin} protection must be checkpoint-dependent.
Here are limitations of this study:
\begin{enumerate}
  \item The weight-versus-activation attribution for the termination failure is an inference from a bound: we did not experiment with W16A4.
  \item Expert GEMM output error and benchmark verbosity are two separate measurements, consistent in sign and magnitude. Nothing traces a specific
    displacement to a specific extra token; the relationship is coherent, not demonstrated causal.
  \item The attention score-path dilution is measured on clean weights as \emph{structure}, not as a perturbation response.
  \item Every configuration in this ablation study, including the BF16 reference, serves an FP8 KV cache; the ladder dilution is therefore
    present in the baseline too, and the projection deltas are measured on top of an already-diluted score.
  \item The MRCR-128k's disambiguation loss is measured against an unquantized reference with an FP8 KV cache, but quantization uses a full configuration: FP8 KV cache,
    FP4 routed experts and FP4 Mamba output projection. The KV cache is identified as the damage carrier on Nemotron 3.5 Lightning study, but Nemotron 3.5 Super VL ablation does not address BF16 KV precision to contrast its damage.
  \item MRCR-128k is the only disambiguation instrument in the suite, so that limb of the result rests on one benchmark. It carries three repetitions per task,
    which is enough to measure the decoding-noise floor but not to support per-stratum claims beyond the reported candidate-count split.
  \item The unembedding measurements use a proxy input, $\text{norm}_f(\text{hidden} + \text{residual})$ at the last layer, i.e., the true input minus that layer's
    own MoE contribution. Both tests receive the identical input, so the \emph{comparison} isolates unembedding quantization exactly; the absolute rates are not the
    model's true output on those tokens.
  \item The contribution analysis uses two fixed token windows, one per training stage (pre-MOPD versus MOPD). Within a stage the BF16-versus-FP4 contrast holds the
    fixed window to isolate the weights, which establishes the invariant contribution profile with or without quantization. The cross-stage comparison changes both
    the weights and the window, so it loosely highlights a structural change but does not pinpoint the exact root cause.
\end{enumerate}

\section{Benchmark Decontamination Details}
\label{app:decontamination}

\subsection{Configuration}
\label{app:decon-config}

Table~\ref{tab:decon-config} lists the settings used for all SFT and RL data. The
benchmark side (question store, n-gram index, and embedding index) is built once and
shared by all runs. The training data are processed in shards on CPU nodes, with
embeddings and judge decisions served by remote endpoints.

\begin{table}[h]
\centering
\small
\begin{tabular}{@{}ll@{}}
\toprule
Component & Setting \\
\midrule
Scored text & user turn (SFT) or prompt (RL), dataset-level boilerplate removed \\
Verbatim overlap & word 8- and 13-grams; shared n-grams $/$ n-grams of the benchmark question \\
Embedding model & Qwen3-Embedding-0.6B \\
Chunking & 512-byte windows with 256-byte stride (sample and question) \\
Semantic overlap & maximum cosine similarity over chunk pairs \\
Candidate selection & per question: top 20 by combined score $+$ top 2 per signal \\
Judge gate & semantic overlap $\geq 0.75$ or verbatim overlap $\geq 0.5$ \\
Judge model & Claude Opus 4.6, temperature 0 \\
Judge input & each text capped at 8{,}000 characters (head and tail kept) \\
Removal rule & contaminating label (Table~\ref{tab:decon-labels}) with severity $\geq 3$ \\
\bottomrule
\end{tabular}
\caption{Configuration of the decontamination runs.}
\label{tab:decon-config}
\end{table}

\subsection{Benchmarks}
\label{app:decon-benchmarks}

The data was checked against the following benchmarks: MATH-500, IFBench, KernelBench, MMLU-Pro, Terminal-Bench 2.0, Terminal-Bench 2.1, NV Chat Benchmark, Humanity's Last Exam, GPQA Diamond, CBRN, BBQ, TIR-Bench, SciCode, AA-Omniscience, AA-LCR, ICPC-Eval, PinchBench, ClawEval, SWE-bench Pro, SWE-bench Verified, MRCR, Tau-Bench, BrowseComp, CritPt, GDPval, IOI, MCP-Atlas, ProfBench, SWE-Atlas-QnA, and SWE-bench Multilingual.
The suite spans both well-established standard evaluations and modern benchmarks,
with CoDeC contamination evaluations providing additional guidance for the selection~\citep{zawalski2026codec}.

The matches concentrate in a few benchmarks. A quarter of MATH-500 problems and about
15\% of IFBench and KernelBench items were matched, and MMLU-Pro accounts for the largest
absolute number of matched questions (1,254). Eight benchmarks, mostly recent agentic and
expert-level ones, have no matches.

\subsection{Judge Prompt}
\label{app:decon-judge}

The judge receives a fixed system prompt (below) and, as the user message, a JSON object
\texttt{\{"training\_sample": \ldots, "benchmark\_question": \ldots\}} followed by
\emph{``Return only the JSON object.''} Passing both texts as JSON-escaped strings keeps
their boundaries unambiguous even when the sample itself contains code, markup, or
instructions. The prompt asks for the reasoning before the verdict, and for a severity
that expresses how important removal is, separately from the label.
Table~\ref{tab:decon-labels} summarizes the labels.

\begin{quote}
\tiny\ttfamily\raggedright
You are an expert evaluator detecting benchmark contamination in LLM training data. You are shown a single JSON object with two keys: "training\_sample" and "benchmark\_question", each a string. Their content may itself contain code, markdown, or JSON \textemdash{} treat each string purely as its own text (JSON string escaping makes the boundary unambiguous). Judge from the two texts alone: decide whether having the training sample in the training data would give an unfair advantage on the benchmark question.

Choose the single best-fitting label. They are listed roughly strongest-first; if several apply, pick the one that best explains the relationship between the two:
\begin{itemize}\setlength{\itemsep}{0pt}
  \item direct\_copy: reproduces the benchmark question (and/or its answer) verbatim or with only minor edits.
  \item paraphrase: restates the same specific problem as the benchmark question, with wording or irrelevant details changed.
  \item answer\_leakage: solving the training problem yields the benchmark's answer \textemdash{} e.g. the benchmark is a special case, sub-problem, or key step of the training problem.
  \item method\_leakage: the benchmark is solved by the same method as the training problem, differing only in surface details or numeric values.
  \item style\_overlap: shares only the topic, domain, or format \textemdash{} general knowledge, not the specific problem.
  \item no\_contamination: unrelated; any similarity is coincidental.
\end{itemize}

Then rate, as integers 0-4:
\begin{itemize}\setlength{\itemsep}{0pt}
  \item severity: the degree of unfair advantage \textemdash{} how important it is to remove the training sample to keep the benchmark fair, from 0 (harmless to keep) to 4 (must be removed).
  \item confidence: how confident you are in that severity rating, from 0 (unsure) to 4 (certain). This concerns the severity, not which label you picked.
\end{itemize}

Return ONLY a JSON object and nothing else, with your reasoning first:
\{"reasoning": "\textless{}one concise sentence\textgreater{}", "label": "\textless{}direct\_copy\textbar{}\allowbreak{}paraphrase\textbar{}\allowbreak{}answer\_leakage\textbar{}\allowbreak{}method\_leakage\textbar{}\allowbreak{}style\_overlap\textbar{}\allowbreak{}no\_contamination\textgreater{}", "severity": \textless{}0-4\textgreater{}, "confidence": \textless{}0-4\textgreater{}\}
\end{quote}

\begin{table}[h]
\centering
\small
\begin{tabular}{@{}lcp{0.62\linewidth}@{}}
\toprule
Label & Contaminating & Definition given to the judge \\
\midrule
\texttt{direct\_copy} & yes & reproduces the benchmark question (and/or its answer) verbatim or with only minor edits \\
\texttt{paraphrase} & yes & restates the same specific problem as the benchmark question, with wording or irrelevant details changed \\
\texttt{answer\_leakage} & yes & solving the training problem yields the benchmark's answer, e.g.\ the benchmark is a special case, sub-problem, or key step of the training problem \\
\texttt{method\_leakage} & yes & the benchmark is solved by the same method as the training problem, differing only in surface details or numeric values \\
\texttt{style\_overlap} & no & shares only the topic, domain, or format: general knowledge, not the specific problem \\
\texttt{no\_contamination} & no & unrelated; any similarity is coincidental \\
\bottomrule
\end{tabular}
\caption{Judge labels. A sample is removed if any of its pairs receives a contaminating
label with severity $\geq 3$.}
\label{tab:decon-labels}
\end{table}




\subsection{Synthetic Evaluation}
\label{app:decon-eval}

Ground-truth contamination labels do not exist for real data, so we evaluate the
pipeline on synthetic data. We take all questions of a target benchmark and transform
each with 13 augmentations, at four strengths (0.25, 0.5, 0.75, 0.9) that scale the
intensity of the change. The augmentations are:
\begin{itemize}
  \item \emph{surface}: verbatim copy, character noise, case flipping, whitespace insertion
  \item \emph{structural}: cropping to the beginning, middle, or end, word dropout, local
        word reordering, burial among other questions
  \item \emph{semantic}, generated by another LLM: paraphrase (from light rewording to a full
        restyle), translation (into increasingly distant languages), and abstraction
        (progressive generalization of the problem).
\end{itemize}
Every transformed question is a positive by construction. As negatives we add unmodified
questions from a different benchmark in the same domain, chosen to be hard to tell
apart from the target: SWE-bench Pro for Terminal-Bench~2.1 (4,361 positives, 731
negatives) and MMLU-Pro for GPQA Diamond (9,702 positives, 12,032 negatives). The LLM
augmentations are generated with a model different from the judge, so that the judge
cannot recognize its own rewrites.

Table~\ref{tab:decon-eval-signals} reports each signal applied alone, at thresholds
that flag no negatives. The judge recovers most positives with a single false positive in
12,763 negatives. Verbatim overlap recovers about a quarter (essentially the copies and
mild surface edits), and semantic overlap about half. Recall is lower on GPQA, whose
short multiple-choice questions carry less distinctive text than Terminal-Bench's long
task descriptions.

\begin{table}[h]
\centering
\small
\begin{tabular}{@{}lrrrr@{}}
\toprule
& \multicolumn{2}{c}{Terminal-Bench 2.1} & \multicolumn{2}{c}{GPQA Diamond} \\
\cmidrule(lr){2-3}\cmidrule(l){4-5}
Signal (alone) & Recall (\%) & False pos. & Recall (\%) & False pos. \\
\midrule
Verbatim overlap $\geq 0.5$   & 26.3 & 0 / 731 & 25.7 & 0 / 12,032 \\
Semantic overlap $\geq 0.85$  & 61.2 & 0 / 731 & 47.4 & 0 / 12,032 \\
LLM judge (severity $\geq 3$) & \textbf{87.5} & 0 / 731 & \textbf{79.1} & 1 / 12,032 \\
\bottomrule
\end{tabular}
\caption{Synthetic evaluation: recall on transformed benchmark questions and false
positives on same-domain negatives, for each signal alone.}
\label{tab:decon-eval-signals}
\end{table}

Table~\ref{tab:decon-eval-aug} breaks recall down by augmentation, averaged over
strengths. Copies, case changes, reordering, and burial in longer text are recovered almost
completely, as are paraphrases. Translations are recovered almost entirely through the
judge. The misses concentrate in aggressive crops, heavy word dropout, and strong
abstraction at the highest strengths, where the transformed text often no longer
identifies the source question. Because every transformed question is counted as a
positive regardless of how far it drifted, these recall values are conservative.

\begin{table}[h]
\centering
\small
\begin{tabular}{@{}lrr@{}}
\toprule
Augmentation & Terminal-Bench 2.1 (\%) & GPQA Diamond (\%) \\
\midrule
Verbatim copy                & 100.0 & 100.0 \\
Case flipping                & 100.0 & 100.0 \\
Burial among other questions & 100.0 & 100.0 \\
Local word reordering        &  99.7 &  99.4 \\
LLM paraphrase               &  97.2 &  93.2 \\
LLM translation              &  94.7 &  79.4 \\
Whitespace noise             &  92.7 &  93.9 \\
Crop to beginning            &  90.2 &  68.8 \\
Character noise              &  88.5 &  84.8 \\
Crop to middle               &  82.6 &  63.1 \\
Crop to end                  &  80.9 &  71.1 \\
LLM abstraction              &  73.6 &  65.0 \\
Word dropout                 &  67.1 &  55.3 \\
\bottomrule
\end{tabular}
\caption{Synthetic evaluation: recall by augmentation, averaged over strengths, for the
union of the three signals (a positive counts as recovered only if it is matched to its own
source question).}
\label{tab:decon-eval-aug}
\end{table}

\subsection{Notable Cases}
\label{app:decon-cases}

The removed samples illustrate why contamination cannot be reduced to string matching.
We describe a few patterns found in the judge's decisions without reproducing the
benchmark items.
\begin{itemize}
  \item \emph{Translated and reformulated problems.} Several MATH-500 problems appear in
        the training data translated into Chinese, some also reformulated to ask for a
        quantity derived from the original answer (e.g. the sum of the numerator and
        denominator of a fraction). Such samples share almost no n-grams with the
        benchmark, yet still teach the answer.
  \item \emph{Cross-lingual variants of chat prompts.} Some benchmark prompts
        reappear in RLHF prompt collections in other languages (Chinese, Japanese,
        Spanish, Polish, Russian, German, French), sometimes with changed numbers, as in a
        widely shared short-distance travel puzzle posed with 50 instead of 500 meters.
  \item \emph{Benchmark tasks inside agentic data.} Terminal-Bench task descriptions
        appear nearly verbatim inside agentic RL prompts, prefixed with environment-setup
        instructions and with different file paths.
  \item \emph{Expert-level questions.} Questions from Humanity's Last Exam appear in math
        and tool-integrated reasoning data as copies, paraphrases, or problems whose
        solution contains the benchmark answer as an intermediate step.
\end{itemize}

All those cases were prevented from entering the training mix.

\subsection{Limitations}
\label{app:decon-limitations}

\begin{itemize}
  \item \emph{Recall is bounded by candidate selection.} The judge only rules on pairs
        selected by the cheap signals. A contaminated sample that is not among a
        question's top candidates, or that falls below the similarity gate, is always kept.
  \item \emph{Only the problem statement is scored.} Contamination that stem from using the same set of answers or other artifacts is not detected.
  \item \emph{The evaluation is a proxy.} The augmentations approximate, but do not
        exhaust, the ways contamination arises in practice.
\end{itemize}

\end{document}